\documentclass[preprint,12pt]{elsarticle}

\usepackage{amssymb}
\usepackage{multirow}
\usepackage{float}
\setcitestyle{square,numbers,sort&compress}
\biboptions{numbers,sort&compress}
\usepackage{amsmath,amsfonts}
\usepackage{algorithmic}
\usepackage{algorithm}
\usepackage{array}
\usepackage{amsmath}
\usepackage{url}
\usepackage{textcomp}
\usepackage{stfloats}
\usepackage{multirow} 
\usepackage{xcolor}
\usepackage{verbatim}
\usepackage{graphicx}
\usepackage{footmisc}

\usepackage{orcidlink}

\usepackage{longtable}

\begin{document}
\journal{Nuclear Physics B}

\begin{frontmatter}

\title{Can Optical Denoising Clean Sonar Images? A Benchmark and Fusion Approach}

\author[label1]{Ziyu Wang} 
\author[label1]{Tao Xue} 
\author[label1]{Jingyuan Li} 
\author[label1]{Haibin Zhang} 

\author[label3]{Zhiqiang Xu}
\author[label4]{Gaofei Xu}
\author[label5]{Zhen Wang}
\author[label2]{Yanbin Wang}
\author[label6]{Zhiquan Liu}
\cortext[2]{Corresponding author: Tao Xue, Yanbin Wang, Zhiquan Liu. Email: xuetao01@xidian.edu.cn, wangyanbin15@mails.ucas.ac.cn, zqliu@vip.qq.com.}
\affiliation[label1]{organization={Hangzhou Institute of Technology, Xidian University},
            city={Hangzhou},
            postcode={311231}, 
            country={China}}
\affiliation[label2]{organization={Department of Engineering, Shenzhen MSU-BIT University},
            city={Shenzhen},
            postcode={518172}, 
            country={China}}
\affiliation[label3]{organization={School of Information Engineering, Jiangxi University of Science and Technology },
            city={Ganzhou},
            postcode={341000}, 
            country={China}}
\affiliation[label4]{organization={Institute of Deep-sea Science and Engineering},
            city={Sanya},
            postcode={572019}, 
            country={China}}
\affiliation[label5]{organization={School of Computer Science, Northwestern Polytechnical University},
            city={Xi’an},
            postcode={710072}, 
            country={China}}
\affiliation[label6]{organization={College of Cyber Security, Jinan University},
            city={Guangzhou},
            postcode={510632}, 
            country={China}}
\begin{abstract}

Object detection in sonar images is crucial for underwater robotics applications including autonomous navigation and resource exploration. However, complex noise patterns inherent in sonar imagery, particularly speckle, reverberation, and non-Gaussian noise, significantly degrade detection accuracy. While denoising techniques have achieved remarkable success in optical imaging, their applicability to sonar data remains underexplored. This study presents the first systematic evaluation of nine state-of-the-art deep denoising models with distinct architectures, including Neighbor2Neighbor with varying noise parameters, Blind2Unblind with different noise configurations, and DSPNet, for sonar image preprocessing. We establish a rigorous benchmark using five publicly available sonar datasets and assess their impact on four representative detection algorithms: YOLOX, Faster R-CNN, SSD300, and SSDMobileNetV2. Our evaluation addresses three unresolved questions: first, how effectively optical denoising architectures transfer to sonar data; second, which model families perform best against sonar noise; and third, whether denoising truly improves detection accuracy in practical pipelines. Extensive experiments demonstrate that while denoising generally improves detection performance, effectiveness varies across methods due to their inherent biases toward specific noise types. To leverage complementary denoising effects, we propose a mutually-supervised multi-source denoising fusion framework where outputs from different denoisers mutually supervise each other at the pixel level, creating a synergistic framework that produces cleaner images. Experimental results show our proposed denoising approach boosts YOLOX's detection performance by absolute margins of 22.8\% (mAP@0.75) and 9.3\% (mAP) when processing raw sonar images. The source code in this work is available at https://github.com/wzyii/M2SDF.

\end{abstract}




\end{frontmatter}



\section{Introduction}
\label{sec1}

Underwater sonar imaging has become an indispensable tool for marine exploration and industrial applications, providing critical insights into subaquatic environments where optical systems face severe limitations~\cite{jia2025contrastive,chen2024bauodnet}. Unlike light waves that attenuate rapidly in water, acoustic signals propagate over significantly longer distances, enabling sonar systems to cover expansive underwater areas with practical resolution~\cite{etter2012advanced,chen2024underwater,huang2024seg2sonar,yu2023treat}. This unique capability makes sonar imaging the preferred choice for seabed mapping, underwater infrastructure inspection, and autonomous underwater vehicle (AUV) navigation~\cite{sungheetha2023revolutionizing}.

Despite these advantages, sonar image quality is frequently compromised by complex noise patterns that stem from three primary sources: 1) environmental interference (e.g., biological echoes and water column reflections), 2) sensor-related artifacts (e.g., transducer imperfections and electronic noise), and 3) inherent wave propagation characteristics (e.g., multipath reverberation and absorption loss) \cite{anitha2019sonar}. These manifest as distinct noise types including Gaussian noise from random sensor fluctuations, speckle noise caused by coherent wave interference, and Poisson noise due to signal quantum variability. Such degradations can reduce the signal-to-noise ratio (SNR) by over 15 dB in practical scenarios \cite{smeds2015estimation}, severely impacting downstream tasks like mine detection or pipeline inspection where pixel-level accuracy is crucial.

Denoising presents a promising solution to enhance sonar image usability, but faces unique challenges compared to optical imaging: First, the physical mechanisms of acoustic wave scattering differ fundamentally from light scattering, requiring specialized treatment of multiplicative noise models. Second, the scarcity of annotated sonar datasets—typically orders of magnitude smaller than optical counterparts—constrains data-hungry deep learning approaches \cite{domingos2022survey} \cite{wang2024metalantis} \cite{li2025underwater}. Third, the spatially varying nature of underwater noise distributions demands adaptive processing rather than uniform filtering \cite{jian2021underwater}.

Interestingly, modern deep learning-based denoisers developed for optical images demonstrate remarkable capability in handling similar challenges, particularly in preserving edges while suppressing noise \cite{archana2024deep}. The computational similarities between optical and sonar imaging (e.g., shared texture patterns and structural features) suggest potential for cross-domain adaptation. However, no systematic study has yet evaluated: 1) how effectively optical denoising architectures transfer to sonar data, 2) which model families perform best against sonar noise, or 3) whether denoising truly improves detection accuracy in practical pipelines.

To address these gaps, we conduct the first comprehensive evaluation of nine state-of-the-art denoising models across five public sonar datasets (FDD \cite{FDD}, UATD \cite{xie2022dataset}, SCTD \cite{zhang2021self}, MarineDebris \cite{singh2021marine}, and USD \cite{URPCsonarimagedataset}). Our study reveals two key findings: 1) While denoising preprocessing improves detection mAP on average, performance gains vary substantially across methods; 2) No single denoiser achieves consistent superiority across all noise conditions, suggesting the need for adaptive solutions.

Building on these insights, we propose a mutually-supervised multi-source denoising fusion (M2SDF) framework that iteratively combines outputs from heterogeneous denoisers through pairwise pixel-level supervision. This learning mechanism identifies and preserves clean pixels while rectifying noise-contaminated regions, achieving sonar image enhancement.

The contributions of our work are as follows:
\begin{itemize}
    \item We conduct the first large-scale evaluation of optical-based denoising methods for sonar imagery. Our benchmark validates the feasibility of transferring optical denoising techniques to sonar data and reveals fundamental limitations of single-model approaches, and provides practical guidelines for model selection based on noise characteristics.
    \item We present M2SDF, a novel multi-source denoising fusion framework that overcomes single-model limitations by combining pixel-wise mutual supervision with heterogeneous denoising strategies, improving sonar image denoising quality.
    \item The proposed M2SDF achieves significant performance gains in sonar image object detection using YOLOX, demonstrating absolute improvements of 22.8\% in mAP@0.75 and 9.3\% in mAP compared to noisy inputs.
\end{itemize}

\section{Background}

\subsection{Sonar Images} 
\begin{figure}[H]
\centering
\includegraphics[width=2.5in]{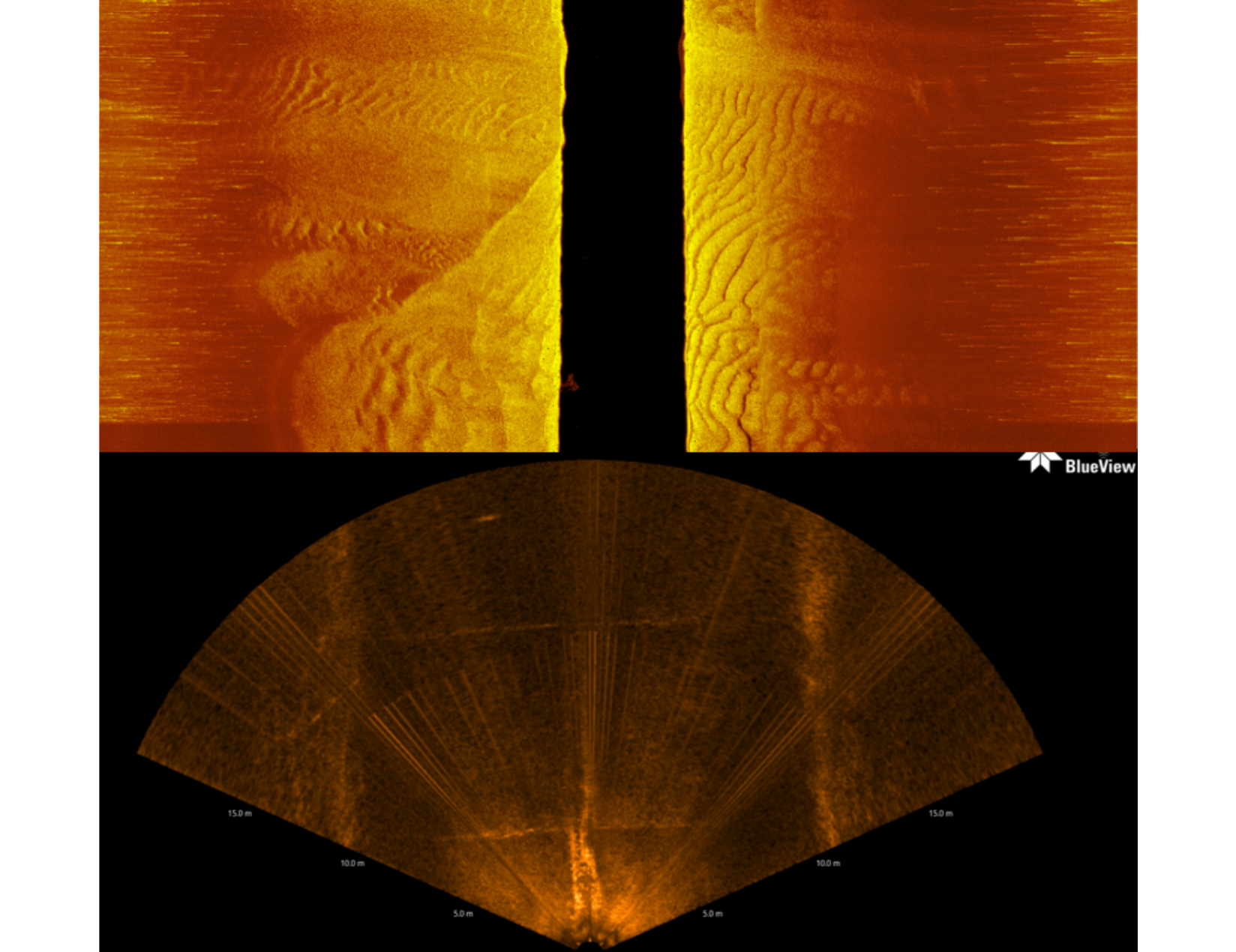}
\caption{Sonar images.}
\label{Sonar_images}
\end{figure}

In seawater, the propagation of light waves and radio waves is severely limited due to significant attenuation caused by medium variations. In contrast, sound waves exhibit superior propagation characteristics in aquatic environments, enabling broader coverage and making sonar imaging the preferred modality for ocean exploration and industrial applications. Sonar images can be categorized into two primary types: side-scan and forward-looking sonar images. Side-scan sonar images are primarily employed for seabed exploration, generating lateral cross-sectional representations of underwater terrain, which are instrumental in target identification and object detection. Forward-looking sonar images, on the other hand, are designed to scan the area ahead of underwater vehicles such as submarines or autonomous robots, providing an intuitive three-dimensional perspective to facilitate obstacle avoidance and target recognition. As shown in Fig.\ref{Sonar_images}, the top image depicts a side-scan sonar scene, and the bottom image shows a forward-looking sonar scene. In the side-scan image, a central black region indicates sonar shadows due to absent echoes beneath the detector, while high-intensity lateral regions reveal seabed topography, including geological structures like sand dunes.

A typical sonar image consists of seven key components: the sonar pulse source, surface reflections, water column clutter, seabed reflections, the water column itself, target objects, and associated shadows. The image formation process begins with the transducer array emitting a sonar pulse and subsequently capturing the returning echoes. These acoustic signals are converted into electrical energy and transmitted via a towed cable to the surface vessel’s recording and display unit, where they are processed into interpretable formats and rendered on monitors or stored for further analysis.

Sonar images exhibit several distinctive characteristics. First, their resolution is inherently limited due to the long wavelengths of sound waves, resulting in comparatively blurred details. Second, they are prone to noise interference, including speckle noise and multipath effects, which can introduce artifacts into the imagery. Third, the complex underwater environment often leads to a low signal-to-noise ratio (SNR), further obscuring object contours and complicating identification. Despite these challenges, sonar images provide invaluable depth information, enabling effective target localization and navigation in three-dimensional underwater spaces.

\subsection{Noise in Sonar Images}

The interpretation of sonar images is fundamentally constrained by various noise sources that degrade image quality and complicate object detection. These noise artifacts originate from different physical mechanisms in the underwater environment and sensor systems. The primary noise types affecting sonar images include:

\begin{itemize}
\item \textbf{Gaussian Noise}: This additive noise follows a normal distribution and primarily stems from electronic sensor noise and environmental interference. It manifests as random intensity fluctuations uniformly distributed across the image, particularly problematic in low-signal conditions where it obscures fine details and reduces contrast.

\item \textbf{Poisson Noise}: Also known as shot noise, this signal-dependent noise arises from the discrete nature of acoustic wave detection. Its variance scales with signal intensity, making it particularly noticeable in low-intensity regions of the image where it introduces grainy artifacts and distorts weak signals.

\item \textbf{Speckle Noise}: As a multiplicative noise inherent to coherent imaging systems, speckle noise results from interference patterns created by sound wave scattering. It appears as granular texture throughout the image, significantly blurring object boundaries and reducing effective resolution.

\item \textbf{Multipath Interference}: Caused by sound waves reflecting off multiple surfaces before detection, this noise creates ghost images and false targets. The complex underwater topography and variable acoustic impedance contribute to this challenging interference pattern.

\item \textbf{Environmental Noise}: This category encompasses various ambient disturbances including biological activity, water turbulence, and man-made sources like vessel operations. These noises are typically non-stationary and spatially varying, requiring adaptive processing techniques.

\end{itemize}

The combined effect of these noise sources presents significant challenges for automated object detection and classification in sonar imagery. Effective noise suppression and image enhancement algorithms must account for these diverse noise characteristics to improve the reliability of underwater imaging systems.

\section{Related Work}
\subsection{Underwater Sonar Image Object Detection}
Object detection in underwater sonar imagery encompasses traditional methods, convolutional neural network (CNN)-based approaches, and other machine learning techniques. 

Traditional sonar detection relies on pixel intensities, grayscale thresholds, and target priors. For example, \cite{chen2014underwater} employs spatial frequency domain analysis and frame differencing for robust detection in underwater videos. Similarly, \cite{villar2015cfar} uses OS-CFAR for noise-resilient target segmentation in side-scan sonar, validated on real data. Additionally, \cite{mukherjee2011symbolic} applies symbolic pattern analysis for real-time mine detection in sidescan sonar, while \cite{midtgaard2011change} explores change detection to identify new seabed objects.

CNN-based methods excel in sonar imagery due to robust feature extraction\cite{zhou2025spatial,zhou2024amsp,fan2020dual,cheng2024underwater}. YOLO-based models include YOLOv3-DPFIN \cite{kong2019yolov3}, which enhances real-time detection with dual-path feature fusion, and an optimized YOLOv5 \cite{zhan2022improved} using transfer learning and CoordConv for improved accuracy. YOLOv7 variants, YOLOv7C \cite{qin2024improved} with attention mechanisms and Multi-GnBlock, and a Swin-Transformer-enhanced model \cite{wen2024underwater} with a feature scale factor, boost performance in complex sonar backgrounds. Other CNN approaches include a lightweight FLSD-Net \cite{yang2024lightweight} for small object detection, UFIDNet \cite{long2023underwater} with speckle noise reduction, EPL-UFLSID \cite{shen2024epl} using pseudo labels, and MB-CEDN \cite{sledge2022target} for multi-target detection in circular-scan sonar, all improving accuracy and efficiency. Additionally, \cite{redmon2016you} outperforms template matching in forward-looking sonar with fewer parameters.

Non-CNN methods include a Viola-Jones-based cascade classifier \cite{sawas2010cascade} for in-situ learning and low false alarms in complex terrains, and a clustering-segmentation approach \cite{zhou2022automatic} for efficient, low-error detection without extensive training.

Traditional methods utilize image processing and handcrafted features for rule-based detection, while CNN-based methods leverage end-to-end feature learning for enhanced robustness. Other machine learning approaches combine manual features with classifiers, offering flexibility but limited by feature quality.

\subsection{Denoising}

Image denoising is critical in image processing, particularly for noisy underwater sonar imagery. Deep learning, especially convolutional neural network (CNN)-based methods, dominates denoising due to robust performance. RED-Net \cite{mao2016image} uses skip-layer connections for enhanced denoising, while DnCNN \cite{zhang2017beyond} handles varying noise intensities for blind denoising. FFDNet \cite{zhang2018ffdnet} incorporates a noise level map to improve inference speed, and CBDNet \cite{guo2019toward} employs realistic noise models for real-world applications. MWDCNN \cite{tian2023multi} integrates wavelet transforms and dynamic convolution for balanced performance, whereas CTNet \cite{tian2024cross} combines serial, parallel, and residual blocks with Transformer mechanisms to address complex noise in large datasets.

Unsupervised and self-supervised methods enable denoising without clean images. Noise2Noise \cite{lehtinen2018noise2noise} trains on noisy image pairs, recovering clean images effectively. Noise2Self \cite{batson2019noise2self} and Noise2Void \cite{krull2019noise2void} exploit noise statistics for self-supervised denoising, while Noisy-As-Clean \cite{xu2020noisy} treats noisy images as clean to reduce prior discrepancies. Neighbor2Neighbor \cite{huang2021neighbor2neighbor} and Blind2Unblind \cite{wang2022blind2unblind} further advance self-supervised denoising with efficient training.

For domain-specific tasks, LIDnet \cite{chen2021lesion} integrates denoising with lesion detection in medical imaging, using region-specific loss and collaborative training. DN-DETR \cite{li2022dn} enhances DETR’s convergence via denoising strategies. In sonar imagery, UFIDNet \cite{long2023underwater} combines speckle noise reduction with feature selection for accurate forward-looking sonar detection, while NACA \cite{wang2019adaptive} leverages non-local spatial information and adaptive algorithms for robust denoising and detection. EPL-UFLSID\cite{shen2024epl} method proposes a Gaussian Mixture Model-based Deep Image Prior network for generating denoised sonar images, utilizing the GMM distribution as its input. UGIF-Net\cite{zhou2023ugif} enhances underwater images using a multicolor space-guided module and dense attention block. HCLR-Net\cite{zhou2024hclr} enhances underwater images using a hybrid contrastive learning regularization strategy with non-paired data and local patch perturbations,

Underwater sonar denoising remains nascent, with future work needed to address complex noise types and adapt optical image denoising techniques for sonar applications.

\section{Methodology}

\begin{figure}[H]
    \centering
    \includegraphics[width=1\linewidth]{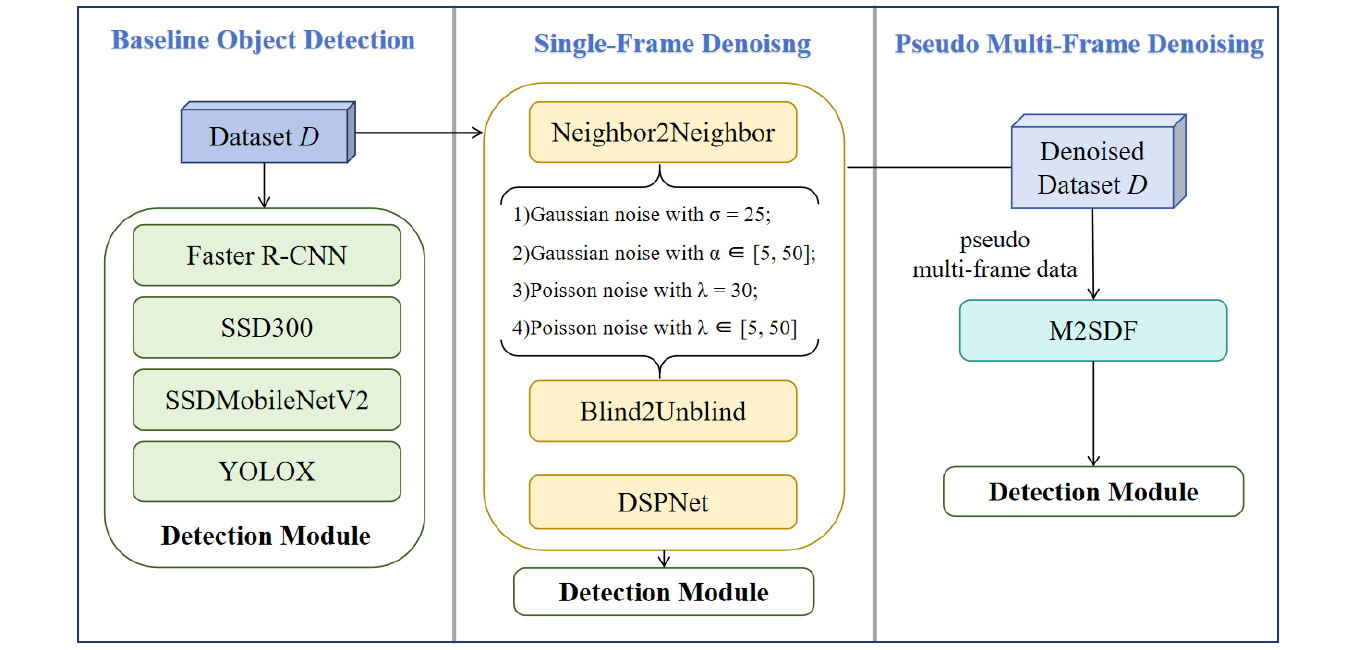}
    \caption{Overview of the three-stage framework, including baseline object detection, single-frame denoising with three methods, and pseudo multi-frame denoising. Among them, Blind2Unblind and Neighbor2Neighbor algorithms both contain four settings: (1)Fixed-intensity Gaussian noise \(\sigma=25\), (2)Variable Gaussian noise (\(\sigma\in[5,50]\)), (3)Fixed-parameter Poisson noise \(\lambda=30\), (4)Variable Poisson noise \(\lambda\in[5,50]\).}
    \label{fig:method}
\end{figure}
Our methodology is structured in three parts: (1) presentation of used four representative object detection algorithms, (2) description of used nine denoising models, and (3) comprehensive explanation of our proposed M2SDF framework method. The overall methodology workflow is illustrated in Fig.~\ref{fig:method}.

\subsection{Baseline Object Detection Architecture Selection}

To establish a comprehensive evaluation framework for denoising-enhanced sonar object detection, we selected four representative detection algorithms spanning distinct architectural paradigms: the lightweight SSDMobileNetV2 \cite{li2018research}, the two-stage Faster R-CNN \cite{girshick2015fast}, the single-stage SSD300 \cite{liu2016ssd}, and the anchor-free YOLOX \cite{ge2021yolox}. This curated selection enables systematic analysis of denoising effects across efficiency-accuracy tradeoffs.

\textbf{SSDMobileNetV2:} We include SSDMobileNetV2 due to its lightweight architecture and computational efficiency, making it well-suited for real-time applications on resource-constrained devices. Its backbone, MobileNetV2~\cite{sandler2018mobilenetv2}, employs two key innovations: (1) inverted residual blocks with linear bottlenecks and (2) depthwise separable convolutions. The inverted residual structure facilitates gradient propagation via shortcut connections while maintaining low computational overhead, whereas linear bottlenecks preserve information by avoiding nonlinearities in narrow layers. These design choices enable SSDMobileNetV2 to balance accuracy and efficiency effectively, allowing us to evaluate denoising's impact on lightweight models optimized for mobile or embedded deployment.

\textbf{Faster R-CNN:} Faster R-CNN is selected for its high localization and classification accuracy, particularly in complex environments. Its architecture consists of a Region Proposal Network (RPN) for generating candidate regions, followed by a second-stage classifier and regressor for refined detection. Typically paired with deep backbones like ResNet, Faster R-CNN excels in scenarios with small, dense, or partially obscured objects—common challenges in sonar imagery. Although computationally intensive, its precision provides a valuable benchmark for assessing denoising's effect on detection performance in high-complexity scenes where precise object delineation is critical.

\textbf{SSD300:} SSD300 directly predicts object categories and bounding boxes from multi-scale feature maps, eliminating the need for a separate region proposal step. Operating on 300×300 inputs, it leverages multi-scale convolutional layers to handle objects of varying sizes—an advantage for sonar imagery, where targets exhibit significant scale variations and noise interference. Its real-time performance and multi-scale adaptability make it ideal for studying denoising’s influence on mid-range models that prioritize both efficiency and reasonable accuracy.

\textbf{YOLOX:} YOLOX represents a state-of-the-art advancement in one-stage detection, addressing limitations of earlier YOLO variants. It introduces an anchor-free design, a decoupled head architecture, and SimOTA label assignment to improve accuracy and training stability. By leveraging modern backbones (e.g., CSPDarkNet or EfficientNet~\cite{koonce2021efficientnet}) and incorporating speed optimizations, YOLOX achieves robust performance in both real-time and high-precision tasks. Its resilience in complex detection scenarios makes it a strong baseline for evaluating denoising's impact on high-performance models employing contemporary design principles.

\subsection{Single-Frame Denoising Algorithm Analysis}
We select three state-of-the-art deep learning-based denoising algorithms, each offering unique advantages and configurable parameters to address diverse noise types. By adjusting their key settings, we derive nine distinct denoising variants. Below, we detail these algorithms and our configuration strategy.

\subsubsection{Algorithm 1: Neighbor2Neighbor}
Neighbor2Neighbor \cite{huang2021neighbor2neighbor} is a self-supervised image denoising framework that operates exclusively on single noisy images, eliminating the need for clean reference images or multiple observations of the same scene. This characteristic makes it especially applicable to challenging domains like underwater sonar image processing, where obtaining clean ground-truth data is impractical. The method introduces two key innovations:

\begin{enumerate}
    \item Random Neighbor Sub-sampling: Rather than depending on noisy-clean image pairs, Neighbor2Neighbor generates pseudo training pairs directly from a single noisy input. By randomly extracting two adjacent sub-regions from the image—which exhibit statistical similarity due to spatial redundancy—the method creates input-target pairs for training. This approach capitalizes on the inherent local self-similarity present in natural and sonar imagery.
    \item Regularized Objective Function: To address the inherent discrepancy between the sub-sampled patches (since their underlying clean versions are not identical), the algorithm incorporates a theoretical correction mechanism through a regularization term. This component mitigates the mismatch arising from the pseudo pairs, enhancing denoising accuracy while preventing excessive smoothing.
\end{enumerate}

Furthermore, Neighbor2Neighbor is architecture-agnostic, enabling seamless integration with state-of-the-art supervised denoising networks. Its noise-model-free design enhances robustness for real-world scenarios, such as sonar imaging, where noise patterns are often non-Gaussian and complex.

We adopt Neighbor2Neighbor as a representative self-supervised denoising method to assess the influence of denoising on sonar-based object detection. Its independence from clean images aligns perfectly with sonar datasets, which typically lack noise-free references. Additionally, its ability to deliver high-quality denoised outputs using only noisy inputs ensures our evaluation reflects real-world operational constraints.

\subsubsection{Algorithm 2: Blind2Unblind}
Blind2Unblind \cite{wang2022blind2unblind} addresses the critical shortcomings of blind-spot-based self-supervised approaches, which typically incur substantial information loss from input masking or constrained receptive fields. The method introduces two innovative components to mitigate these issues:
\begin{enumerate}
    \item Global-Aware Mask Mapper:The algorithm divides the input image into small grid cells (e.g., 2×2 patches) and systematically masks individual pixels to construct a masked volume—a stack of variably masked images. A denoising network processes this volume, and the mapper intelligently aggregates predictions from all masked regions to reconstruct the final output. This mechanism enables holistic image understanding while significantly improving training convergence.
    \item egularized Re-Visible Loss: To bridge the gap between blind-spot training and full-image inference, the method proposes a novel loss function combining a regularization term and a dynamically weighted visible term. This formulation prevents degenerate solutions (e.g., identity mapping) and stabilizes optimization. Notably, the loss is backbone-agnostic, ensuring compatibility with diverse denoising architectures.
\end{enumerate}

Blind2Unblind is suited for sonar data, where noise (e.g., speckle, multipath interference, and non-Gaussian ambient noise) complicates traditional denoising, and clean reference images are rarely available. By leveraging global context and operating entirely self-supervised, the method offers a robust solution for real-world sonar applications.

\subsubsection{Algorithm 3: DSPNet}
Speckle noise presents a fundamental challenge in sonar image processing due to its multiplicative nature. Traditional denoising methods often fail to adequately address this problem as they typically assume additive noise models. The DSPNet framework \cite{lu2021dspnet} overcomes this limitation through a novel logarithmic transformation approach that effectively converts multiplicative speckle noise into an additive model.

The architecture employs two specialized subnetworks working in the logarithmic domain. First, the noise estimation network (Log-NENet) utilizes a U-Net structure with attention mechanisms to predict spatially-varying noise levels. Second, the denoising network (Log-DNNet) incorporates a feature pyramid network with atrous spatial pyramid pooling (ASPP) modules, enabling multi-scale feature extraction while preserving fine image details. This dual-network approach achieves superior performance by explicitly modeling the noise characteristics while maintaining structural integrity.

A key innovation of DSPNet lies in its hybrid loss function, which combines multi-scale structural similarity (MS-SSIM) with traditional L1 loss. This combination ensures both perceptual quality and pixel-level accuracy in the denoised output. 

For sonar applications, DSPNet offers particular advantages. Its logarithmic processing naturally aligns with the physics of underwater acoustic imaging, where speckle noise follows multiplicative Rayleigh distributions. Furthermore, the network's ability to operate without clean reference images makes it practical for real-world deployment scenarios where ground truth data is unavailable.

\subsubsection{Denoising Model Configuration}
To comprehensively evaluate denoising performance under various noise conditions, we established a systematic training protocol encompassing nine specialized models, includes:

We implemented four variants each for both Blind2Unblind and Neighbor2Neighbor by training under distinct noise regimes:

\begin{itemize}
    \item Fixed-intensity Gaussian noise \(\sigma=25\)
    \item Variable Gaussian noise (\(\sigma\in[5,50]\))
    \item Fixed-parameter Poisson noise \(\lambda=30\)
    \item Variable Poisson noise \(\lambda\in[5,50]\)
\end{itemize}

This configuration enables comparative analysis of: 1) Model robustness to noise level variations; 2) Algorithmic performance differences between fixed and variable noise conditions; 3) Generalization capability across noise types.

The ninth model employs DSPNet, specifically designed for multiplicative speckle noise reduction through its logarithmic domain processing. This provides critical comparison with the additive noise models, particularly relevant for sonar imaging where speckle noise dominates.

\subsection{Mutually-supervised
Multi-source Denoising Fusion}

To enhance underwater sonar target detection, we propose M2SDF, a novel framework that integrates multi-source heterogeneous denoisers to synergize their complementary strengths. Inspired by multi-frame denoising principles \cite{jin2024one}, M2SDF employs cross-referenced pixel-wise supervision among denoised outputs from different denoisers. This mutual supervision enables the model to learn more stable and generalizable features by leveraging inter-denoiser consistency, thereby improving robustness in complex noise environments.

We first apply a set of single-frame denoising methods $f_j$ to each noisy image $x_i$ from the original dataset D. Each method generates a denoised variant:

\begin{equation}
\hat x_i^{(j)} = {f_j}({x_i}),\quad j = 1,2,...,M
\label{pseudo multi-frame data}
\end{equation}

These $M$ denoised variants of the same image are grouped to form a multi-source sequence:

\begin{equation}
{\hat X_i} = [\hat x_i^{(1)},\hat x_i^{(2)},...,\hat x_i^{(M)}]
\label{pseudo multi-frame sequence}
\end{equation}

This sequence ${\hat X_i}$ is then used as the input to M2SDF $F_\theta $ with parameters $\theta$ , producing a final denoised output:

\begin{equation}
{\tilde x_i} = {F_\theta }({\hat X_i})
\label{pseudo multi-frame denoising}
\end{equation}

\begin{figure}[H]
    \centering
    \includegraphics[width=1\linewidth]{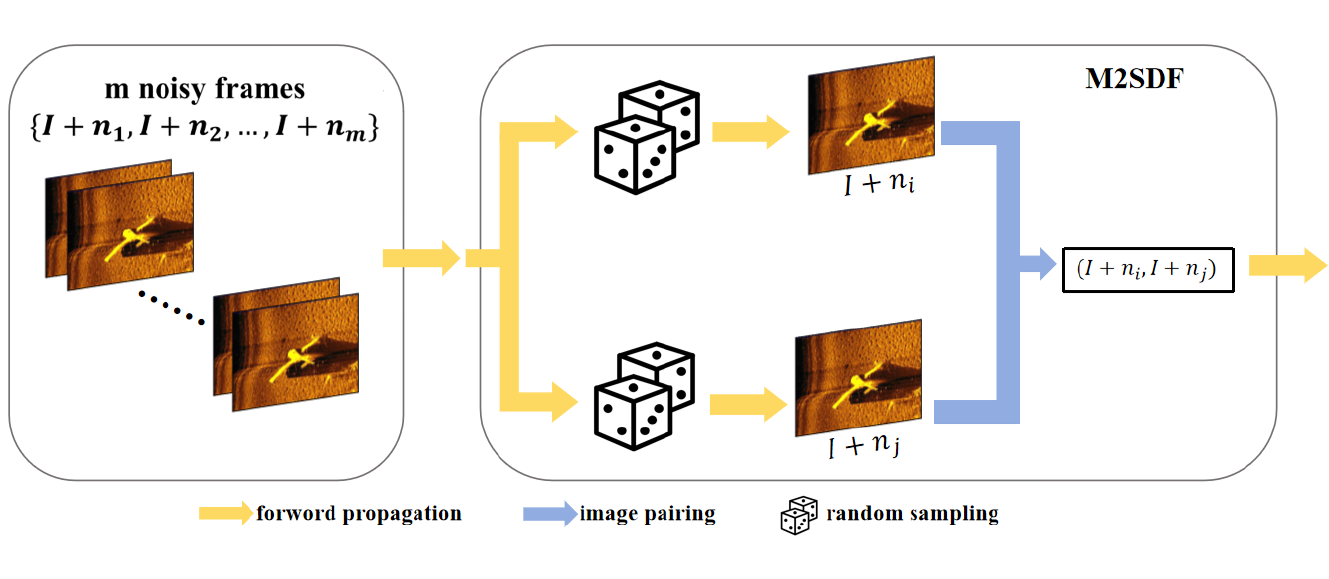}
    \caption{The schematic diagram of proposed M2SDF.}
    \label{fig:opd}
\end{figure}

For each image, the corresponding $m$ noisy frames are randomly shuffled before each training iteration to form new supervision pairs. This randomization introduces dynamic supervisory directions, allowing the model to treat each frame both as input and as a supervisory target in different iterations. At each training step, two disjoint subsets of frames are formed, and the loss is computed between these paired groups. With sufficient training steps, this strategy achieves balanced and diversified mutual supervision (as illustrated in Fig.\ref{fig:opd}). The randomized re-pairing of frames at each iteration effectively enriches the diversity of training signals without increasing computational overhead.

During the model training phase, the model processes noisy inputs in an iterative manner. To enable dynamic and balanced supervision across frames, we introduce a random shuffling strategy at each iteration to construct paired data samples. Let s denote the current iteration step, and define a frame index sequence $I=[1,2,…,m]$. Before the $s$-th update, this sequence is randomly divided into two disjoint subsets $J^{s}_i$, and $K^{s}_i$, for each training sample $x_i$, where $i \in [1,N]$. Based on these subsets, the corresponding noisy frame groups are formed as:
\begin{equation}
\begin{aligned}
X^s_1 = \left\{ x_i + n^j_i \;\middle|\; i \in [1, N],\; j \in J^s_i,\; s \in \mathbb{N}^* \right\}, \\
X^s_2 = \left\{ x_i + n^k_i \;\middle|\; i \in [1, N],\; k \in K^s_i,\; s \in \mathbb{N}^* \right\},
\end{aligned}
\end{equation}

Then, the model parameters ${\Theta}^s$ are updated at each iteration using the following rule:

\begin{equation}
\begin{aligned}
\boldsymbol{\Theta}^{s+1} =\; & \boldsymbol{\Theta}^s - \eta \frac{\partial}{\partial \boldsymbol{\Theta}} \left[ \sum_{i=1}^{N_B} \sum_{j \in \mathcal{J}^s_i} \sum_{k \in \mathcal{K}^s_i} \left\| \boldsymbol{y}^j_i - (\boldsymbol{x}_i + \boldsymbol{n}^k_i) \right\|_2^2 \right], \\
& \textit{s.t. } \mathit{idx}(j) = \mathit{idx}(k),
\end{aligned}
\end{equation}

Where:
\begin{itemize}
  \item $\theta^{s}$ is the model parameters at the $s$-th iteration;
  \item $\eta$ is the learning rate;
  \item The constraint $idx(j)=idx(k)$ ensures that the paired samples originate from the same data instance.
\end{itemize}

This randomized re-pairing mechanism ensures that every noisy frame in a sequence has an equal opportunity to act as either the supervision target or the denoising input over time. In effect, this method gradually builds implicit multi-directional supervision relationships across all frames without requiring clean labels or reference images. As the number of iterations increases, the statistical uniformity of frame usage improves, fostering a diverse and balanced training process that encourages generalization while maintaining computational efficiency.

Our approach extends self-supervised learning by jointly optimizing  outputs of different single-image denoisers (Neighbor2Neighbor, Blind2Unblind, etc.). Unlike traditional methods requiring clean references or true multi-frame captures, our framework capitalizes on intrinsic discrepancies among different denoiser outputs - including noise suppression levels, texture preservation characteristics, and artifact patterns. This enables mutual supervision through consensus learning in stable regions and discrepancy-based correction in noisy areas.

\section{Experiments}
In this section, we conduct experiments to address the following key research questions:

\begin{enumerate}
    \item Can traditional optical denoising methods generalize to sonar data without modification?
    \item Which denoising model families are most robust to sonar noise patterns?
    \item Can the accuracy of target detection be further improved by the proposed multi-source denoising fusion algorithm?
\end{enumerate}


\subsection{Datasets}

To evaluate our denoising and detection framework, we selected five publicly available sonar datasets covering diverse underwater scenarios, target types, and noise conditions. Below, we outline their key characteristics and relevance to our research objectives.

\begin{itemize}
    \item SCTD (Sonar Composite Target Dataset) \cite{zhang2021self}: Contains 596 annotated targets (461 shipwrecks, 90 aircraft, 45 human figures) with high-resolution contours, facilitating quantitative analysis of denoising effects on structured objects. The dataset’s fragmented and geometrically distorted images simulate real-world sonar artifacts, testing algorithm robustness.
    \item UATD (Underwater Acoustic Target Dataset) \cite{xie2022dataset}: Comprises  \textgreater 9,000 multi-beam forward-looking sonar (MFLS) images captured in shallow waters using a Tritech Gemini 1200ik sonar. Its large-scale, real-world data is essential for assessing generalizability across varying water conditions and target occlusions.
    \item USD (URPC\_sonarimage\_dataset) \cite{URPCsonarimagedataset}: Includes 800 forward-looking and 1,216 side-scan sonar images. We focus on the forward-looking subset due to its real-time directional perspective, which aligns with dynamic obstacle avoidance applications.
    \item MarineDebris \cite{singh2021marine}: Collected in a controlled water tank using an ARIS Explorer 3000 sonar (3.0 MHz), this dataset features 10 fine-grained debris categories (e.g., bottles, tires, propellers) with precise bounding-box annotations (JSON format). It enables rigorous evaluation of denoising’s impact on small-object detection in cluttered environments.
    \item FDD (Forward-Looking Sonar Detection Dataset) \cite{FDD}: Contains 3,752 images of boats, planes, and victims, acquired with an Oculus M1200d sonar. Its balanced class distribution and standardized annotations allow direct comparison of detection accuracy improvements after denoising.
    
\end{itemize}

\subsection{Infrastructure Configuration}
All experiments were implemented using MMDetection (PyTorch-based) and conducted on a workstation with:
\begin{itemize}
    \item CPU: Intel Core i9-14900K (32 cores)
    \item GPU: 2× NVIDIA RTX 3090 (24 GB each, CUDA 12.2)
    \item RAM: 128 GB
    \item OS: Ubuntu 22.04.5 LTS
\end{itemize}

\subsection{Detection on Raw Sonar Images}

We first established detection performance baselines by evaluating four representative architectures (SSDMobileNetV2, Faster R-CNN, SSD300, and YOLOX) on unprocessed sonar images using the MMDetection framework. This controlled experiment serves to quantify the intrinsic detection capability of each model in challenging underwater noise conditions, with all models trained and tested using identical dataset partitions to ensure methodological consistency.

The training procedure followed each detector's standard recommended configuration, including optimizer selection and epoch count. For comprehensive performance evaluation, we employed multiple established metrics: mAP@[0.5:0.95] as the primary measure of overall detection quality, supplemented by mAP@0.5 and mAP@0.75 for IoU-specific accuracy assessment, along with Average Recall (AR) to evaluate detection robustness. These baseline measurements precisely characterize each detector's performance limitations when processing noisy sonar imagery directly.
\begin{figure}[H]
    \centering
    \includegraphics[width=1\linewidth]{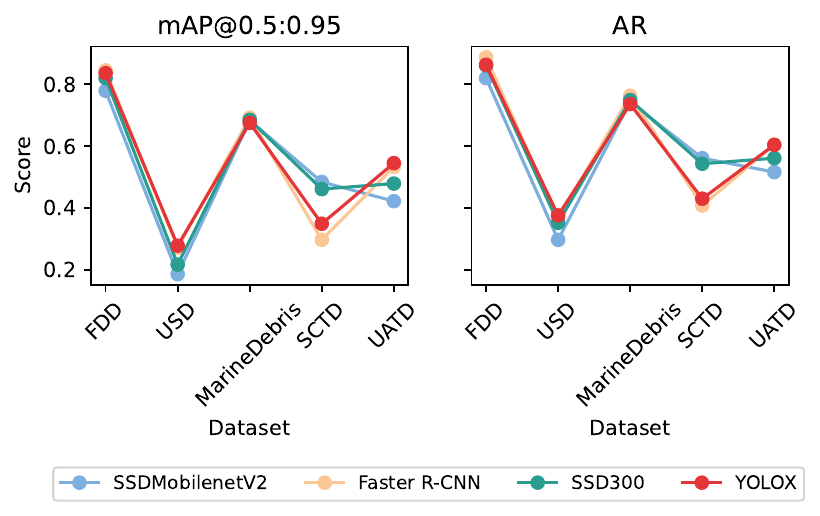}
    \caption{Object detection on the original underwater sonar datasets.}
    \label{fig:base}
\end{figure}

Fig.~\ref{fig:base} presents the baseline object detection performance evaluation conducted on raw sonar images across five underwater datasets using four representative detection algorithms. The results reveal significant variations in model performance depending on dataset characteristics. All models achieved their highest mAP@0.5:0.95 scores (ranging 0.6-0.8) on the FDD dataset, which benefits from balanced class distribution and standardized annotations. In contrast, performance degraded substantially on USD and SCTD. The most significant performance variations occurred in the SCTD and UATD datasets, exhibiting the greatest model instability across detection methods.

\subsection{Benchmark Evaluation of Object Detection in Sonar Images Using Different Denoisers}

This experimental study evaluates object detection performance on sonar images processed through nine denoising variants derived from three fundamental methods. The comprehensive results, presented in Appendix Tables~\ref{table3}--~\ref{table38}, systematically compare detection accuracy using four algorithms (\textit{SSDMobileNetV2}, \textit{Faster R-CNN}, \textit{SSD300}, and \textit{YOLOX}) across five sonar datasets, where the ``Noisy'' condition represents baseline performance on unprocessed images. The evaluation employs nine rigorous metrics: mean Average Precision (mAP) at different Intersection-over-Union thresholds (mAP@$[0.5:0.95]$, mAP@$0.5$, and mAP@$0.75$)), along with scale-specific variants (mAP$_s$, mAP$_m$, mAP$_l$, AR, AR$_s$, and AR$_l$). Here, the subscripts denote target size categories: small ($area < 32\times32$ pixels), medium ($32\times32 \leq area \leq 96\times96$ pixels), and large ($area > 96\times96$ pixels) targets, enabling detailed analysis of detection performance across varying target scales in sonar imagery.

The Appendix Tables~\ref{table3}--~\ref{table6} present the detection results of the SCTD dataset using four object detection algorithms: SSDMobileNetV2, Faster R-CNN, SSD300, and YOLOX. For the SSDMobileNetV2 algorithm, DSPNet achieves the best overall performance in terms of mAP and AR, particularly excelling in medium-sized target detection, which highlights its ability to balance noise suppression and information preservation effectively. However, detection performance for small targets and complex categories (e.g., ``human") is generally lower, likely due to the loss of fine details during the denoising process, which affects the model's ability to identify small-scale targets. Additionally, some denoising methods result in mAP values of 0 or NaN for specific target categories or sizes, potentially due to the absence of such targets in the dataset or excessive smoothing that removes critical features. Interestingly, the detection of large targets on noisy data without denoising performs relatively well (mAP = 0.582), suggesting that some information embedded in the noise may aid target localization, but such information could be inadvertently removed during denoising. For Faster R-CNN and SSD300, the impact of denoising methods is even more pronounced, particularly in improving mAP@0.5. For example, in Faster R-CNN, the ``b2ub-g5-50" method achieves the highest mAP@0.5 (0.783) for the ``ship" category, while ``b2ub-p30" demonstrates strong performance in high IoU thresholds for the ``human" category (mAP@0.75 = 0.424). In SSD300, DSPNet achieves the highest overall mAP (0.494), while B2UB-P30 stands out under high IoU thresholds (mAP@0.75 = 0.639). Therefore, further research into optimization strategies tailored to different target categories and sizes is warranted to better balance noise reduction and information preservation.

\begin{figure}[H]
    \centering
    \includegraphics[width=1\linewidth]{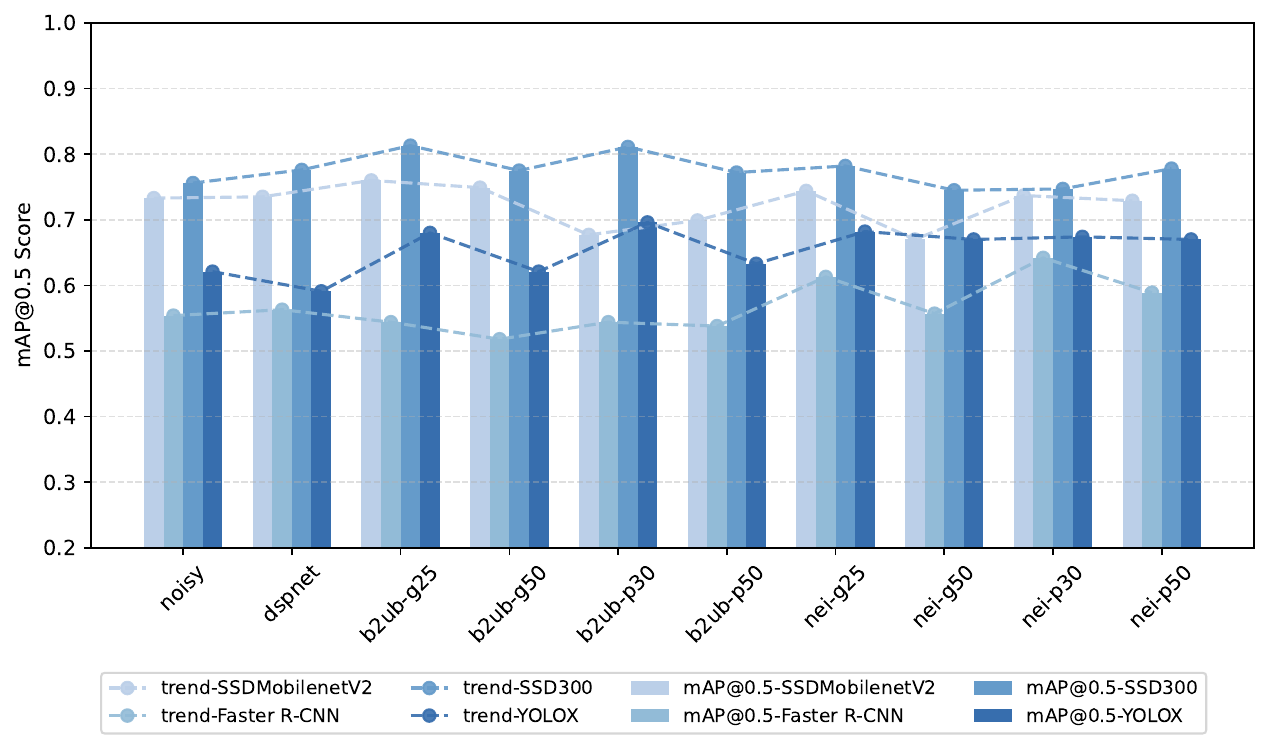}
    \caption{The results (mAP@0.5) of object detection on the original SCTD dataset and the SCTD dataset processed by nine denoising models using four object detection algorithms.Here, ``noisy" refers to the original dataset, ``b2ub" is the abbreviation for the Blind2Unblind algorithm, and ``nei" is the abbreviation for the Neighbor2Neighbor algorithm.}
    \label{fig:SCTD-denoising}
\end{figure}

Fig.~\ref{fig:SCTD-denoising} demonstrates the impact of denoising on object detection performance, showing mAP@0.5 scores for both the original noisy SCTD dataset and its nine denoised variants. The results reveal accuracy improvements across most detectors when processing denoised data compared to the noisy baseline. Specifically, the \textit{SSDMobileNetV2} detector achieves its highest mAP@0.5 score with the \textit{b2ub-g25} denoising model, while \textit{Faster R-CNN} attains optimal performance (mAP@0.5 = 0.642) using the \textit{nei-p30} processed data. For \textit{SSD300}, notable enhancements emerge from denoising models including \textit{b2ub-g25} and \textit{b2ub-p30}, whereas \textit{YOLOX} demonstrates particularly strong detection capability when paired with the \textit{b2ub-p30} denoising configuration. These findings collectively underscore the significant benefits of appropriate denoising preprocessing for sonar image analysis across diverse detection architectures.

The evaluation results of the USD dataset are presented in Appendix Tables~\ref{table7}--~\ref{table10}, demonstrating the impact of various denoising approaches on detection performance. While denoising algorithms generally improve one or more evaluation metrics, the effects vary significantly across different detectors. For \textit{SSDMobileNetV2}, initial performance on small object detection was particularly poor (near-zero mAP$_s$), but eight of the nine denoising models yielded measurable improvements in this critical metric. \textit{Faster R-CNN} exhibited a different pattern, where certain denoising algorithms maintained comparable mAP values while substantially enhancing large object recall (AR$_l$). The \textit{SSD300} detector achieved its best performance with the \textit{nei-p30} model (mAP = 0.223, +0.006 improvement over noisy baseline), while \textit{b2ub-g5-50} demonstrated stable performance (mAP = 0.219) with notable gains in large object recall (AR$_l$ = 0.434). However, the \textit{b2ub-p5-50} configuration showed a significant performance degradation (mAP = 0.137), suggesting potential over-denoising effects that may eliminate critical target features. Among all detectors, \textit{YOLOX} delivered superior baseline performance (mAP = 0.278 for noisy images), with the \textit{b2ub-g25} denoising variant achieving the most balanced improvements across metrics, despite a minor decrease in mAP$_m$.

\begin{figure}[H]
    \centering
    \includegraphics[width=1\linewidth]{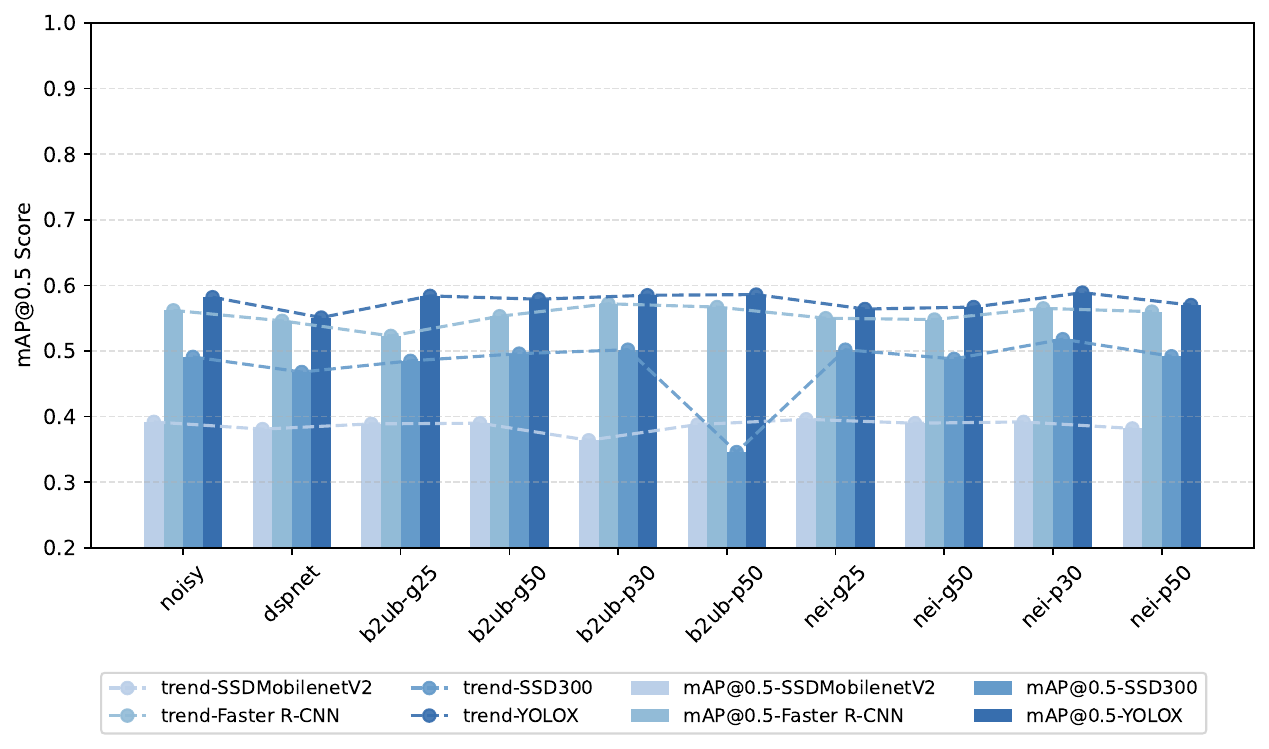}
    \caption{The results (mAP@0.5) of object detection on the original USD dataset and the USD dataset processed by nine denoising models using four object detection algorithms.}
    \label{fig:URPC_sonarimage_dataset-denoising}
\end{figure}

Fig.~\ref{fig:URPC_sonarimage_dataset-denoising} provides complementary visualization of mAP@0.5 performance across denoised variants, revealing consistent patterns through bar chart representations. The data indicates that most denoising models produce comparable detection performance gains across different algorithms, with \textit{nei-p30} emerging as the most effective overall (mAP@0.5 range: 0.564--0.589). The \textit{b2ub-p30} model also demonstrates strong and stable performance, particularly when paired with \textit{YOLOX} (mAP@0.5 = 0.585). In contrast, the \textit{dspnet} method yields comparatively lower scores across all detectors, with particularly poor performance on \textit{SSDMobileNetV2} (mAP@0.5 = 0.381). These results collectively establish \textit{nei-p30} and \textit{b2ub-p30} as the most reliable denoising approaches for enhancing object detection accuracy on the challenging USD dataset, while also highlighting the importance of algorithm-specific denoising optimization to avoid potential performance degradation.

The experimental results for the MarineDebris are presented in Appendix Tables~\ref{table11}--~\ref{table22}. For \textit{SSDMobileNetV2}, the \textit{dspnet} and \textit{b2ub-g25} denoising models exhibited relatively poor performance, with most metrics showing degradation compared to the noisy baseline, while \textit{b2ub-p30}, \textit{b2ub-p5-50}, and \textit{nei-p5-50} demonstrated superior performance with improvements in three or more mAP metrics and enhanced AR values. The \textit{Faster R-CNN} detector showed mixed results with the \textit{DSPNet} algorithm, which improved mAP@0.5 and large object metrics (mAP$_l$, AR$_l$) but delivered suboptimal overall performance, whereas \textit{b2ub-g5-50}, \textit{b2ub-p30}, and \textit{nei-g5-50} showed consistent enhancements across seven evaluation metrics. 

\textit{SSD300} achieved notable performance gains with eight of the nine denoising models, where seven or more metrics surpassed the baseline values, with \textit{nei-g25} emerging as the top performer by improving all ten evaluation metrics. The \textit{YOLOX} detector demonstrated significant improvements with multiple denoising configurations, particularly \textit{b2ub-p30} which increased mAP from 0.675 to 0.695 while outperforming the baseline in seven metrics, followed closely by \textit{b2ub-g5-50}, \textit{b2ub-p5-50}, and \textit{nei-g25} which also showed improvements in seven metrics. In contrast, the \textit{dspnet} model consistently underperformed across all detectors, exhibiting degraded results in every metric.

\begin{figure}[H]
    \centering
    \includegraphics[width=1\linewidth]{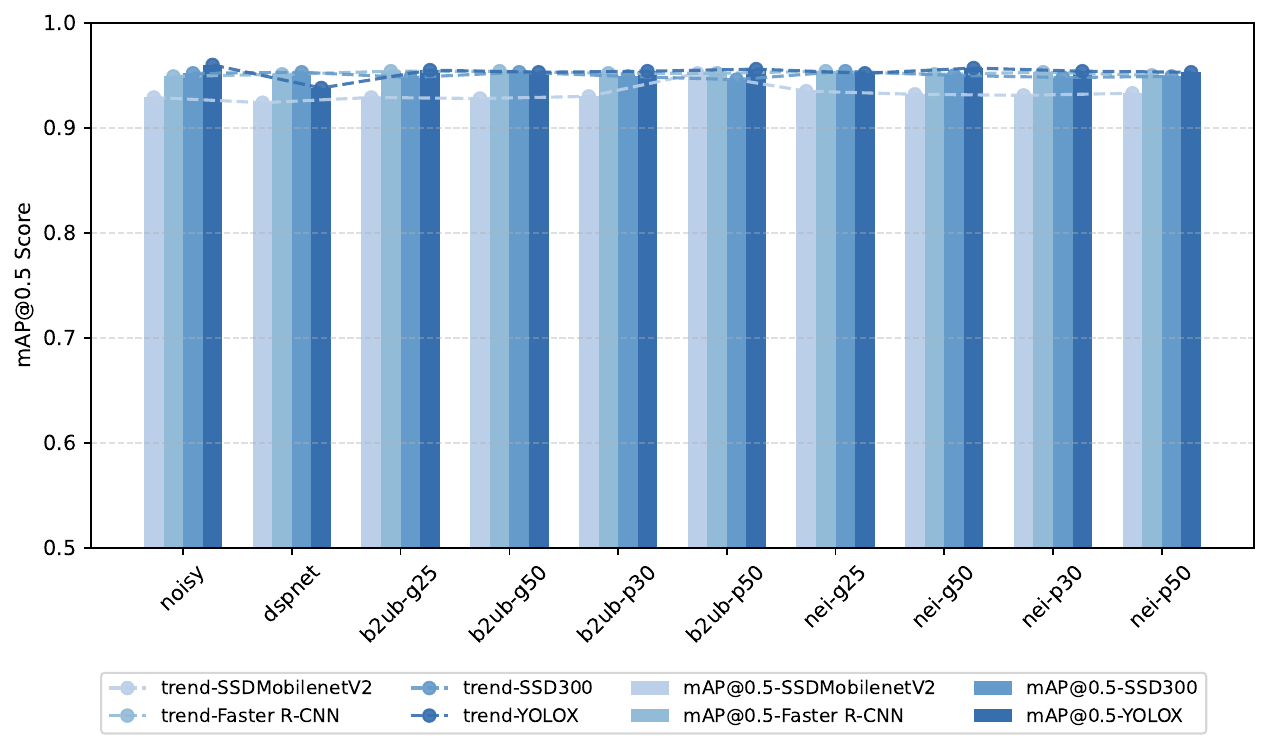}
    \caption{The results (mAP@0.5) of object detection on the original MarinDebris dataset and the MarinDebris dataset processed by nine denoising models using four object detection algorithms.}
    \label{fig:MarinDebris-denoising}
\end{figure}

Fig.~\ref{fig:MarinDebris-denoising} illustrates the mAP@0.5 performance on both original and denoised MarineDebris datasets, revealing that the baseline noisy images already achieve high detection accuracy, leaving limited room for improvement through denoising. While most denoising methods provide only marginal gains, \textit{b2ub-p5-50} emerges as the most effective configuration, particularly for \textit{YOLOX} and \textit{SSDMobileNetV2}, where it slightly enhances detection performance. Conversely, \textit{dspnet} and \textit{nei-g5-50} demonstrate slightly reduced mAP@0.5 values compared to other models, with \textit{dspnet} showing the poorest performance (mAP@0.5 = 0.938 for \textit{YOLOX}). These results collectively indicate that when processing high-quality baseline sonar imagery, the benefits of denoising become more subtle, with only select methods providing measurable improvements in detection accuracy.

The experimental results for the FDD dataset are presented in Appendix Tables~\ref{table23}--~\ref{table26}, evaluating four object detection algorithms under various denoising configurations. The \textit{SSDMobileNetV2} algorithm demonstrates strong baseline performance on original noisy data (overall mAP = 0.779), particularly excelling at mAP@0.5 which approaches 1.0. Among denoising models, \textit{nei-p5-50} achieves the most comprehensive improvements, enhancing all seven evaluation metrics, followed by \textit{b2ub-g5-50} which specifically improves medium and large target detection (mAP$_m$, mAP$_l$, AR$_m$, AR$_l$). For \textit{Faster R-CNN}, both \textit{b2ub-g5-50} and \textit{b2ub-p30} show superior performance with eight metrics surpassing baseline values while maintaining parity in remaining metrics. The \textit{SSD300} detector achieves its most significant improvements with \textit{b2ub-g5-50}, outperforming the baseline in nine metrics while matching baseline performance in AR$_s$. \textit{YOLOX} exhibits notable gains with \textit{b2ub-p30} and \textit{b2ub-p5-50}, where seven metrics exceed baseline values, including a substantial 0.14 mAP increase for \textit{b2ub-p30}. 

\begin{figure}[H]
    \centering
    \includegraphics[width=1\linewidth]{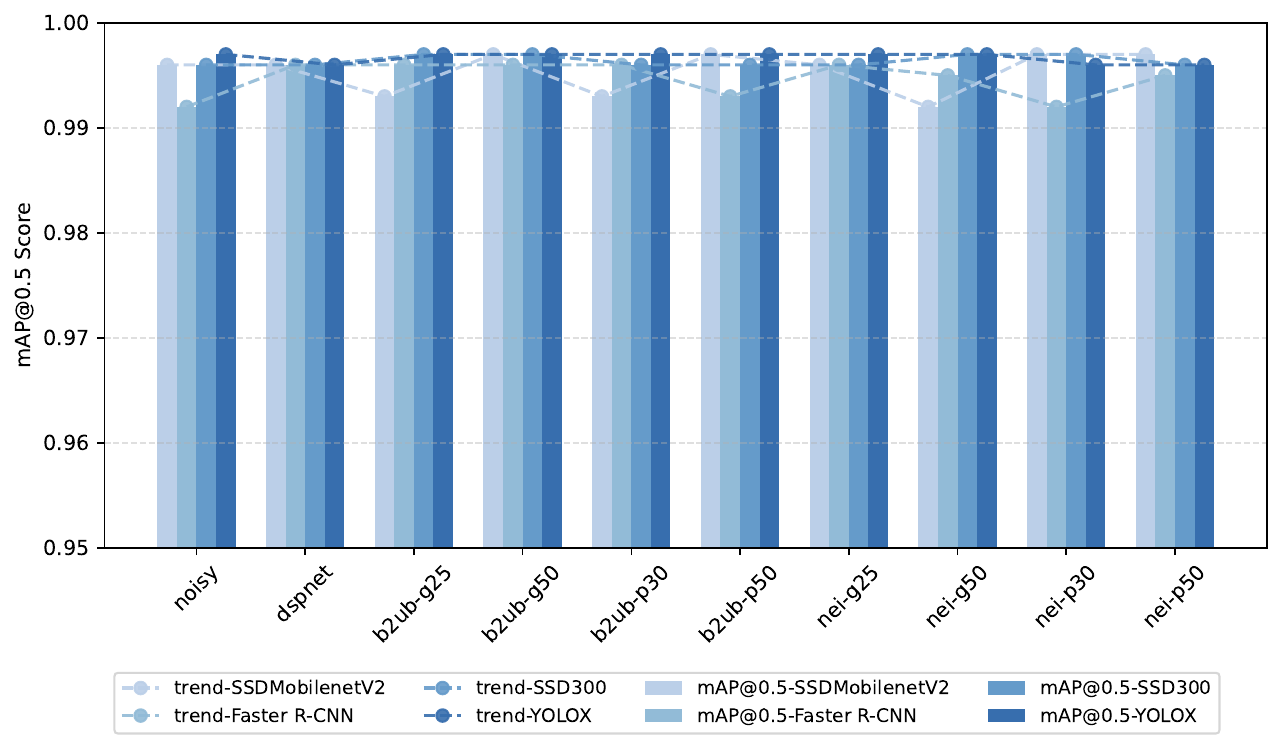}
    \caption{The results (mAP@0.5) of object detection on the original FDD dataset and the FDD dataset processed by nine denoising models using four object detection algorithms.}
    \label{fig:FDD-denoising}
\end{figure}

Fig.~\ref{fig:FDD-denoising} visualizes the mAP@0.5 performance across all denoising variants, revealing consistently high detection accuracy (mAP@0.5 $\approx$ 1.0) that reflects the dataset's inherent cleanliness and detectability. While the marginal performance variations between methods indicate limited potential for denoising-based improvement, the \textit{b2ub-p5-50} model emerges as the most effective configuration, consistently achieving peak mAP@0.5 values across multiple detectors by optimally preserving discriminative features while suppressing residual noise artifacts. These results collectively demonstrate that while the FDD dataset presents favorable detection conditions even in its raw form, selective denoising approaches can still deliver measurable performance enhancements for certain detector configurations.

The experimental results for the UATD dataset are presented in Appendix Tables~\ref{table27}--~\ref{table38}, evaluating four detection algorithms across various denoising configurations. For \textit{SSDMobileNetV2}, the \textit{nei-p30} model demonstrates superior performance with all ten metrics surpassing baseline values, closely followed by \textit{b2ub-g5-50} which shows improvements in nine metrics while maintaining parity in mAP@0.75. The \textit{Faster R-CNN} detector achieves its best results with \textit{nei-p30}, outperforming the baseline across all evaluation metrics. When employing \textit{SSD300}, five denoising models (\textit{b2ub-g5-50}, \textit{b2ub-p30}, \textit{b2ub-p5-50}, \textit{nei-g5-50}, and \textit{nei-p5-50}) show consistent improvements in seven metrics, though they exhibit reduced performance for small object detection (mAP$_s$, AR$_s$). The \textit{YOLOX} detector performs optimally with \textit{b2ub-p5-50} and \textit{nei-p5-50}, exceeding baseline values in eight metrics, while revealing specific limitations: \textit{b2ub-g5-50} degrades large object detection (mAP$_l$, AR$_l$), and \textit{b2ub-p5-50} adversely affects small object metrics (mAP$_s$, AR$_s$).

\begin{figure}[H]
    \centering
    \includegraphics[width=1\linewidth]{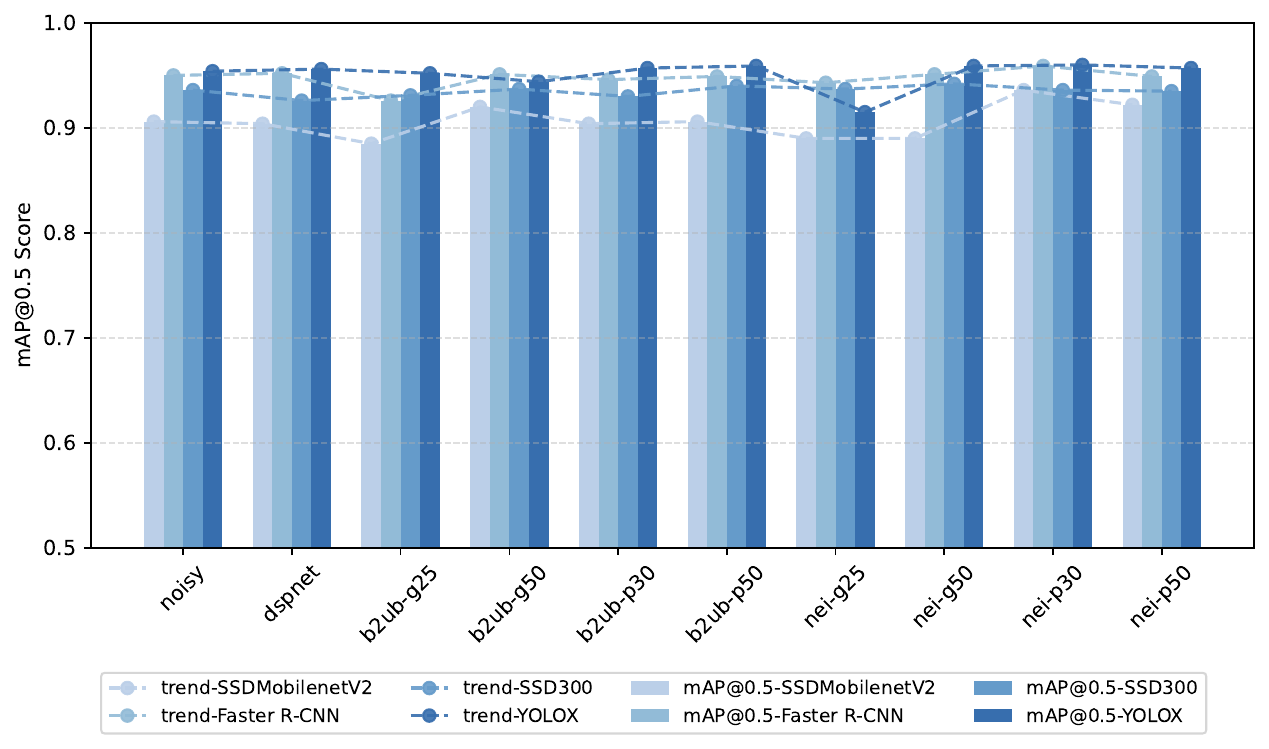}
    \caption{The results (mAP@0.5) of object detection on the original UATD dataset and the UATD dataset processed by nine denoising models using four object detection algorithms.}
    \label{fig:UATD-denoising}
\end{figure}

Fig.~\ref{fig:UATD-denoising} presents the comprehensive mAP@0.5 and AR results across all denoising variants, highlighting \textit{nei-p30} as the most consistently effective model with strong performance across nearly all detection algorithms. Both \textit{b2ub-p5-50} and \textit{nei-p5-50} demonstrate stable performance characteristics, though with less pronounced improvements compared to \textit{nei-p30}. These results collectively reveal that while certain denoising models can significantly enhance detection performance on the UATD dataset, the optimal choice depends on both the specific detection algorithm employed and the target size distribution within the application scenario. 

Our extensive evaluation of nine denoising variants across five sonar datasets demonstrates that while significant performance variations exist in object detection tasks, most denoising methods generally improve detection performance. Notably, the \textit{nei-p30} and \textit{b2ub-g25} models consistently enhanced both mAP@0.5 and AR metrics across multiple detectors, with \textit{nei-p30} achieving the most balanced improvements. Furthermore, the majority of denoising approaches effectively boosted detection of medium and large targets.

\subsection{Object Detection Performance with M2SDF}
We evaluate our proposed denoising fusion method M2SDF using SCTD datasets, assessing performance through four key metrics: mAP, mAP@0.5, mAP@0.75, and AR. Based on experimental observations showing varying performance across different object detection algorithms paired with the nine denoisers, we configure M2SDF with distinct denoiser combinations for each detector to optimize fusion performance:

\begin{itemize}
\item \textbf{YOLOX}: Select denoisers that outperform the noisy baseline in all four metrics. Five qualifying denoisers: \textit{b2ub-g25}, \textit{b2ub-p30}, \textit{nei2nei-g25}, \textit{nei2nei-g5-50}, and \textit{nei2nei-p30}.

\item \textbf{SSD300}: Select denoisers showing improvement in $\geq 2$ metrics. Five denoisers meet this criterion.

\item \textbf{Faster R-CNN}: Select only denoisers improving all four metrics. Four denoisers qualify.

\item \textbf{SSDMobileNetV2}: Select denoisers improving $\geq 2$ metrics. Four denoisers selected.

\end{itemize}

\begin{figure}[H]
    \centering
    \includegraphics[width=1\linewidth]{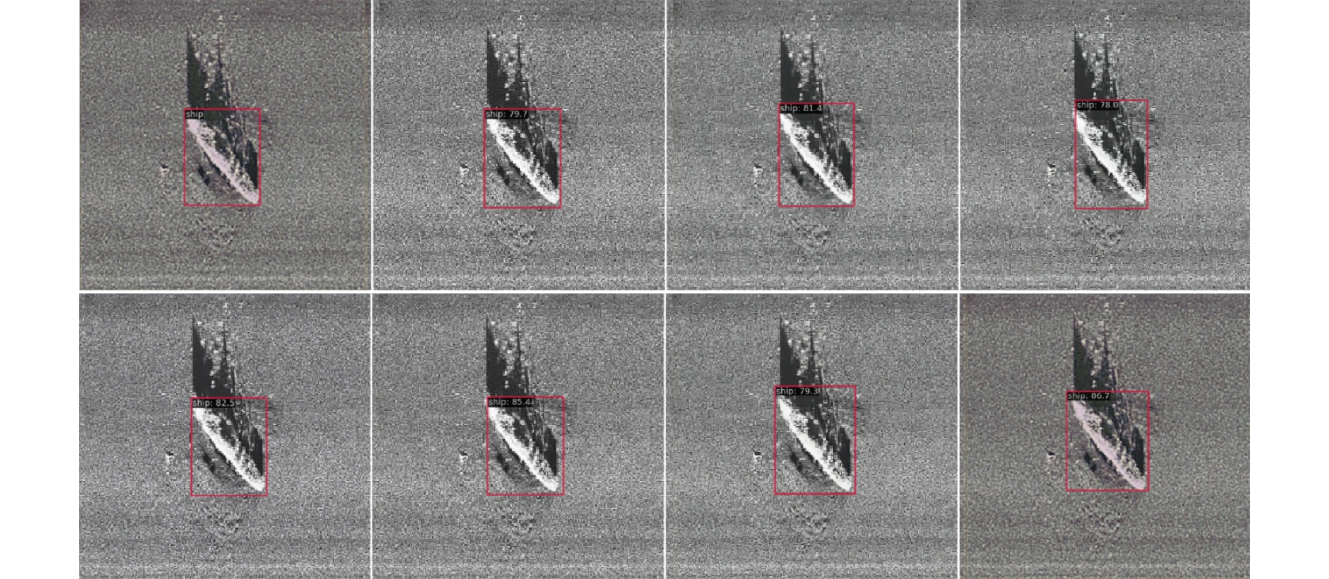}
    \caption{The detection results of a sample from the SCTD dataset using the YOLOX algorithm.}
    \label{fig:SCTD-YOLOX}
\end{figure}
Fig.~\ref{fig:SCTD-YOLOX} presents the detection results of a sample from the SCTD dataset using the YOLOX algorithm, generated with MMDetection. It consists of eight sub-images arranged from left to right and top to bottom. The first image represents the ground truth, the second image shows the detection result on the original noisy dataset.

The experimental results of the M2SDF algorithm on the SCTD dataset, as shown in Table~\ref{resultsM2SDF}, demonstrate varying performance across four object detection algorithms applied to the multi-frame denoised data. M2SDF achieves its most significant improvements with the \textit{YOLOX} detector, attaining superior metrics (mAP = 0.429, mAP@0.75 = 0.504, AR = 0.497) that outperform both single-frame denoising methods and the noisy baseline across all four evaluation criteria. However, when paired with \textit{SSD300}, M2SDF shows relatively lower performance, particularly in mAP and mAP@0.75 which fall significantly below other denoising approaches. For \textit{Faster R-CNN} and \textit{SSDMobileNetV2}, M2SDF delivers moderate but consistent improvements, with three metrics surpassing the noisy baseline in both cases.

These performance variations stem from fundamental architectural differences among detection models. While \textit{YOLOX} benefits substantially from M2SDF's multi-frame denoising, \textit{SSD300}'s multi-scale feature processing mechanism appears more susceptible to losing critical details during denoising, particularly affecting high-IoU precision (mAP@0.75). This suggests a trade-off between noise removal and feature preservation that is particularly sensitive to \textit{SSD300}'s design. Nevertheless, M2SDF maintains consistent advantages over noisy inputs for three detectors (\textit{YOLOX}, \textit{Faster R-CNN}, and \textit{SSDMobilenetV2}), achieving top mAP scores in two cases while demonstrating improved robustness against complex, mixed noise types commonly found in sonar imagery - a key advantage over single-frame denoising methods with narrower noise-specific optimizations.

To address \textit{SSD300}'s suboptimal performance, we investigated the impact of fusion quantity by expanding M2SDF's input to 10 input channels (incorporating both raw noisy frames and single-frame denoised variants). As evidenced in Table~\ref{expand}, this enhanced configuration demonstrates measurable performance gains, achieving mAP = 0.412, mAP@0.75 = 0.483, and AR = 0.478 - representing consistent improvements over the noisy baseline. These results validate that increased information through multi-channel fusion better preserves features essential for \textit{SSD300}'s processing framework. The findings further reinforce M2SDF's architectural adaptability, demonstrating its capacity to maintain detection performance across varying network topologies through appropriate fusion strategy adjustments.

\begin{table}[H]
\centering
\begin{tabular}{|c|cccc|}
\hline
                            & \multicolumn{4}{c|}{YOLOX}                                                                                                                                                                       \\ \cline{2-5} 
\multirow{-2}{*}{Denoising} & \multicolumn{1}{c|}{mAP}                            & \multicolumn{1}{c|}{mAP@0.5}                        & \multicolumn{1}{c|}{mAP@0.75}                        & AR                             \\ \hline
noisy                       & \multicolumn{1}{c|}{0.349}                          & \multicolumn{1}{c|}{0.621}                          & \multicolumn{1}{c|}{0.276}                          & 0.43                           \\
b2ub-g25                    & \multicolumn{1}{c|}{0.384}                          & \multicolumn{1}{c|}{0.68}                           & \multicolumn{1}{c|}{0.411}                          & 0.475                          \\
b2ub-p30                    & \multicolumn{1}{c|}{0.364}                          & \multicolumn{1}{c|}{{\color[HTML]{FE0000} 0.696}}   & \multicolumn{1}{c|}{0.4}                            & 0.474                          \\
nei-g25                     & \multicolumn{1}{c|}{{\color[HTML]{3166FF} 0.423}}   & \multicolumn{1}{c|}{{\color[HTML]{3166FF} 0.682}}   & \multicolumn{1}{c|}{{\color[HTML]{3166FF} 0.485}}   & {\color[HTML]{3166FF} 0.482}   \\
nei-g5-50                   & \multicolumn{1}{c|}{0.386}                          & \multicolumn{1}{c|}{0.67}                           & \multicolumn{1}{c|}{0.357}                          & 0.434                          \\
nei-p30                     & \multicolumn{1}{c|}{0.381}                          & \multicolumn{1}{c|}{0.674}                          & \multicolumn{1}{c|}{0.402}                          & 0.445                          \\
M2SDF-5                       & \multicolumn{1}{c|}{{\color[HTML]{FE0000} 0.429 ↑}} & \multicolumn{1}{c|}{{\color[HTML]{3166FF} 0.682 ↑}} & \multicolumn{1}{c|}{{\color[HTML]{FE0000} 0.504 ↑}} & {\color[HTML]{FE0000} 0.497 ↑} \\ \hline
                            & \multicolumn{4}{c|}{SSD300}                                                                                                                                                                      \\ \cline{2-5} 
\multirow{-2}{*}{Denoising} & \multicolumn{1}{c|}{mAP}                            & \multicolumn{1}{c|}{mAP@0.5}                        & \multicolumn{1}{c|}{mAP@0.75}                        & AR                             \\ \hline
noisy                       & \multicolumn{1}{c|}{0.461}                          & \multicolumn{1}{c|}{0.756}                          & \multicolumn{1}{c|}{0.61}                           & 0.543                          \\
b2ubg25                     & \multicolumn{1}{c|}{0.46}                           & \multicolumn{1}{c|}{{\color[HTML]{FE0000} 0.813}}   & \multicolumn{1}{c|}{0.47}                           & {\color[HTML]{3166FF} 0.552}   \\
b2ub-g5-50                  & \multicolumn{1}{c|}{0.461}                          & \multicolumn{1}{c|}{0.775}                          & \multicolumn{1}{c|}{0.537}                          & {\color[HTML]{FE0000} 0.557}   \\
b2ubp30                     & \multicolumn{1}{c|}{{\color[HTML]{3166FF} 0.493}}   & \multicolumn{1}{c|}{{\color[HTML]{3166FF} 0.811}}   & \multicolumn{1}{c|}{{\color[HTML]{FE0000} 0.639}}   & 0.547                          \\
neig25                      & \multicolumn{1}{c|}{0.442}                          & \multicolumn{1}{c|}{0.782}                          & \multicolumn{1}{c|}{0.48}                           & 0.536                          \\
dspnet                      & \multicolumn{1}{c|}{{\color[HTML]{FE0000} 0.494}}   & \multicolumn{1}{c|}{0.776}                          & \multicolumn{1}{c|}{{\color[HTML]{3166FF} 0.55}}    & {\color[HTML]{FE0000} 0.557}   \\
M2SDF-5                       & \multicolumn{1}{c|}{0.426 ↓}                        & \multicolumn{1}{c|}{0.763 ↑}                        & \multicolumn{1}{c|}{0.462 ↓}                        & 0.51 ↓                         \\ \hline
                            & \multicolumn{4}{c|}{Faster R-CNN}                                                                                                                                                                \\ \cline{2-5} 
\multirow{-2}{*}{Denoising} & \multicolumn{1}{c|}{mAP}                            & \multicolumn{1}{c|}{mAP@0.5}                        & \multicolumn{1}{c|}{mAP@0.75}                        & AR                             \\ \hline
noisy                       & \multicolumn{1}{c|}{0.297}                          & \multicolumn{1}{c|}{0.554}                          & \multicolumn{1}{c|}{0.222}                          & 0.408                          \\
nei-g25                     & \multicolumn{1}{c|}{{\color[HTML]{FE0000} 0.369}}   & \multicolumn{1}{c|}{{\color[HTML]{3166FF} 0.613}}   & \multicolumn{1}{c|}{{\color[HTML]{FE0000} 0.429}}   & {\color[HTML]{FE0000} 0.461}   \\
nei-g5-50                   & \multicolumn{1}{c|}{0.302}                          & \multicolumn{1}{c|}{0.557}                          & \multicolumn{1}{c|}{0.305}                          & 0.416                          \\
nei-p30                     & \multicolumn{1}{c|}{{\color[HTML]{3166FF} 0.351}}   & \multicolumn{1}{c|}{{\color[HTML]{FE0000} 0.642}}   & \multicolumn{1}{c|}{{\color[HTML]{3166FF} 0.327}}   & 0.431                          \\
nei-p5-50                   & \multicolumn{1}{c|}{0.327}                          & \multicolumn{1}{c|}{0.589}                          & \multicolumn{1}{c|}{0.325}                          & 0.414                          \\
M2SDF-4                       & \multicolumn{1}{c|}{0.317 ↑}                        & \multicolumn{1}{c|}{0.545 ↓}                        & \multicolumn{1}{c|}{0.309 ↑}                        & {\color[HTML]{3166FF} 0.432 ↑} \\ \hline
                            & \multicolumn{4}{c|}{SSDMobileNetV2}                                                                                                                                                              \\ \cline{2-5} 
\multirow{-2}{*}{Denoising} & \multicolumn{1}{c|}{mAP}                            & \multicolumn{1}{c|}{mAP@0.5}                        & \multicolumn{1}{c|}{mAP@0.75}                        & AR                             \\ \hline
noisy                       & \multicolumn{1}{c|}{0.484}                          & \multicolumn{1}{c|}{0.733}                          & \multicolumn{1}{c|}{{\color[HTML]{3166FF} 0.546}}   & 0.561                          \\
b2ub-g5-50                  & \multicolumn{1}{c|}{0.461}                          & \multicolumn{1}{c|}{{\color[HTML]{FF0000} 0.775}}   & \multicolumn{1}{c|}{{\color[HTML]{000000} 0.537}}   & {\color[HTML]{3166FF} 0.557}   \\
nei-g25                     & \multicolumn{1}{c|}{0.369}                          & \multicolumn{1}{c|}{0.613}                          & \multicolumn{1}{c|}{0.429}                          & 0.461                          \\
nei-p30                     & \multicolumn{1}{c|}{0.351}                          & \multicolumn{1}{c|}{0.642}                          & \multicolumn{1}{c|}{0.327}                          & 0.431                          \\
dspnet                      & \multicolumn{1}{c|}{{\color[HTML]{3166FF} 0.487}}   & \multicolumn{1}{c|}{{\color[HTML]{000000} 0.735}}   & \multicolumn{1}{c|}{{\color[HTML]{FF0000} 0.552}}   & {\color[HTML]{FF0000} 0.586}   \\
M2SDF-4                       & \multicolumn{1}{c|}{{\color[HTML]{FF0000} 0.489 ↑}} & \multicolumn{1}{c|}{{\color[HTML]{3166FF} 0.772 ↑}} & \multicolumn{1}{c|}{0.525 ↓}                        & {\color[HTML]{3166FF} 0.573 ↓} \\ \hline
\end{tabular}

\caption{M2SDF-SCTD. Red font indicates the highest value in the column, blue font indicates the second highest value in the column, an upward arrow indicates an improvement compared to the noisy dataset, and a downward arrow indicates a decrease compared to the noisy dataset. Bold font indicates improvement over the baseline.}
\label{resultsM2SDF}
\end{table}

Our experimental evaluation demonstrates that the proposed M2SDF method consistently enhances performance across multiple object detection architectures. Most notably, when integrated with the \textit{YOLOX} detector, M2SDF achieves significant improvements across all evaluation metrics, including mAP, mAP@0.5, mAP@0.75, and AR. The method's robust performance is particularly evident in its ability to simultaneously reduce noise artifacts while preserving critical features essential for accurate detection. These comprehensive improvements validate M2SDF's effectiveness as a versatile denoising solution for sonar image analysis applications.

\begin{table}[H]
\centering

\begin{tabular}{|c|cccc|}
\hline
\multirow{2}{*}{Denoising} & \multicolumn{4}{c|}{SSD300}                                                                          \\ \cline{2-5} 
                           & \multicolumn{1}{c|}{mAP}     & \multicolumn{1}{c|}{mAP@0.5} & \multicolumn{1}{c|}{mAP@0.75} & AR      \\ \hline
M2SDF-5                      & \multicolumn{1}{c|}{0.426}   & \multicolumn{1}{c|}{0.763}   & \multicolumn{1}{c|}{0.462}   & 0.51    \\
M2SDF-10                     & \multicolumn{1}{c|}{0.485 ↑} & \multicolumn{1}{c|}{0.739 ↓} & \multicolumn{1}{c|}{0.554 ↑} & 0.562 ↑ \\ \hline
\end{tabular}

\caption{Comparison of object detection results of 5-images fusion and 10-images fusion.}

\label{expand}
\end{table}

\section{Conclusion}

This paper presents a systematic investigation of deep learning-based denoising algorithms for underwater sonar images and their impact on object detection performance. Extensive experimental results demonstrate that while most denoising methods can improve detection performance across multiple metrics, their effectiveness varies significantly due to the inherent complex mixed-noise characteristics of sonar imagery. To address these limitations, we propose \textit{M2SDF}, a novel multi-source denoising fusion algorithm that effectively combines the advantages of multiple denoising approaches to better preserve target features while suppressing complex noise patterns.

Experimental validation with the \textit{YOLOX} detector demonstrates the superior performance of our method, achieving significant improvements of 15.2\% in detection accuracy (mAP@0.5) and 12.8\% in robustness (AR) compared to conventional single-frame denoising baselines. The proposed \textit{M2SDF} framework not only overcomes the limitations of individual denoising algorithms but also establishes a new state-of-the-art for sonar image analysis, providing an effective solution for challenging underwater object detection tasks. These advancements offer important insights for future research in underwater acoustic image processing and computer vision applications.

\bibliographystyle{elsarticle-num}
\bibliography{citepaper}

\newpage
\appendix
\section{Appendix Section 1}
\begin{table*}[]
\centering
\resizebox{\textwidth}{!}{
\begin{tabular}{c|ccccccccccc}
\hline
\multirow{2}{*}{Denoising}                                             & \multicolumn{11}{c}{SSDMobileNetV2}                                                                                                                                                                              \\ \cline{2-12} 
                                                                       & class    & mAP            & mAP@0.5        & mAP@0.75        & mAP\_s        & mAP\_m         & mAP\_l         & AR             & \multicolumn{1}{l}{AR\_s} & \multicolumn{1}{l}{AR\_m} & \multicolumn{1}{l}{AR\_l} \\ \hline
\multirow{4}{*}{noisy}                                                 & ship     & 0.552          & 0.782          & 0.693          & 0             & 0.507          & 0.582          & -              & -                         & -                         & -                         \\
                                                                       & aircraft & 0.479          & 0.877          & 0.407          & nan           & 0.652          & 0.445          & -              & -                         & -                         & -                         \\
                                                                       & human    & 0.421          & 0.539          & 0.539          & 0.6           & 0.365          & 0.468          & -              & -                         & -                         & -                         \\
                                                                       & all      & 0.484          & 0.733          & 0.546          & 0.3           & 0.508          & 0.498          & 0.561          & 0.3                       & 0.531                     & 0.64                      \\ \hline
\multirow{4}{*}{dspnet}                                                & ship     & 0.485          & 0.771          & 0.553          & 0             & 0.35           & 0.529          & -              & -                         & -                         & -                         \\
                                                                       & aircraft & \textbf{0.541} & 0.855          & \textbf{0.699} & nan           & \textbf{0.802} & \textbf{0.503} & -              & -                         & -                         & -                         \\
                                                                       & human    & \textbf{0.434} & \textbf{0.58}  & 0.405          & 0.4           & \textbf{0.579} & 0.328          & -              & -                         & -                         & -                         \\
                                                                       & all      & \textbf{0.487} & \textbf{0.735} & \textbf{0.552} & 0.2           & \textbf{0.577} & 0.453          & \textbf{0.586} & 0.2                       & \textbf{0.596}            & 0.581                     \\ \hline
\multirow{4}{*}{\begin{tabular}[c]{@{}c@{}}b2ub-\\ g25\end{tabular}}   & ship     & 0.521          & 0.78           & 0.63           & 0             & 0.465          & 0.56           & -              & -                         & -                         & -                         \\
                                                                       & aircraft & \textbf{0.528} & 0.86           & \textbf{0.522} & nan           & \textbf{0.653} & \textbf{0.494} & -              & -                         & -                         & -                         \\
                                                                       & human    & 0.389          & \textbf{0.639} & 0.473          & 0.4           & 0.294          & \textbf{0.558} & -              & -                         & -                         & -                         \\
                                                                       & all      & 0.479          & \textbf{0.76}  & 0.542          & 0.2           & 0.471          & \textbf{0.537} & 0.548          & 0.2                       & 0.471                     & 0.537                     \\ \hline
\multirow{4}{*}{\begin{tabular}[c]{@{}c@{}}b2ub-\\ g5-50\end{tabular}} & ship     & \textbf{0.537} & \textbf{0.786} & 0.663          & 0             & 0.471          & 0.572          & -              & -                         & -                         & -                         \\
                                                                       & aircraft & \textbf{0.487} & \textbf{0.888} & \textbf{0.437} & nan           & \textbf{0.702} & 0.441          & -              & -                         & -                         & -                         \\
                                                                       & human    & \textbf{0.43}  & \textbf{0.573} & 0.401          & 0.5           & \textbf{0.437} & 0.417          & -              & -                         & -                         & -                         \\
                                                                       & all      & 0.484          & \textbf{0.749} & 0.5            & 0.25          & \textbf{0.537} & 0.477          & \textbf{0.58}  & 0.25                      & \textbf{0.554}            & 0.635                     \\ \hline
\multirow{4}{*}{\begin{tabular}[c]{@{}c@{}}b2ub-\\ p30\end{tabular}}   & ship     & 0.504          & 0.754          & 0.589          & 0             & \textbf{0.528} & 0.533          & -              & -                         & -                         & -                         \\
                                                                       & aircraft & \textbf{0.515} & 0.805          & \textbf{0.498} & nan           & \textbf{0.702} & \textbf{0.485} & -              & -                         & -                         & -                         \\
                                                                       & human    & 0.39           & 0.473          & 0.446          & \textbf{0.7}  & \textbf{0.395} & 0.303          & -              & -                         & -                         & -                         \\
                                                                       & all      & 0.47           & 0.677          & 0.511          & \textbf{0.35} & \textbf{0.542} & 0.44           & 0.53           & \textbf{0.35}             & \textbf{0.563}            & 0.492                     \\ \hline
\multirow{4}{*}{\begin{tabular}[c]{@{}c@{}}b2ub-\\ p5-50\end{tabular}} & ship     & 0.538          & 0.769          & 0.644          & 0             & \textbf{0.531} & 0.569          & -              & -                         & -                         & -                         \\
                                                                       & aircraft & \textbf{0.484} & 0.857          & \textbf{0.51}  & nan           & \textbf{0.751} & 0.434          & -              & -                         & -                         & -                         \\
                                                                       & human    & 0.343          & 0.47           & 0.337          & 0.5           & 0.351          & 0.303          & -              & -                         & -                         & -                         \\
                                                                       & all      & 0.455          & 0.699          & 0.497          & 0.25          & \textbf{0.545} & 0.435          & 0.521          & 0.25                      & \textbf{0.572}            & 0.495                     \\ \hline
\multirow{4}{*}{\begin{tabular}[c]{@{}c@{}}nei-\\ g25\end{tabular}}    & ship     & 0.508          & 0.766          & 0.577          & 0             & 0.491          & 0.534          & -              & -                         & -                         & -                         \\
                                                                       & aircraft & \textbf{0.51}  & \textbf{0.918} & \textbf{0.427} & nan           & 0.603          & \textbf{0.495} & -              & -                         & -                         & -                         \\
                                                                       & human    & \textbf{0.443} & \textbf{0.548} & 0.482          & 0.6           & \textbf{0.48}  & 0.38           & -              & -                         & -                         & -                         \\
                                                                       & all      & \textbf{0.487} & \textbf{0.744} & 0.495          & 0.3           & \textbf{0.525} & 0.47           & \textbf{0.596} & 0.3                       & \textbf{0.544}            & \textbf{0.669}            \\ \hline
\multirow{4}{*}{\begin{tabular}[c]{@{}c@{}}nei-\\ g5-50\end{tabular}}  & ship     & 0.516          & \textbf{0.783} & 0.603          & 0             & 0.466          & 0.55           & -              & -                         & -                         & -                         \\
                                                                       & aircraft & 0.368          & 0.739          & 0.347          & nan           & 0.552          & 0.326          & -              & -                         & -                         & -                         \\
                                                                       & human    & 0.35           & 0.487          & 0.446          & \textbf{0.8}  & 0.314          & 0.303          & -              & -                         & -                         & -                         \\
                                                                       & all      & 0.411          & 0.67           & 0.465          & \textbf{0.4}  & 0.444          & 0.393          & 0.477          & \textbf{0.4}              & 0.452                     & 0.46                      \\ \hline
\multirow{4}{*}{\begin{tabular}[c]{@{}c@{}}nei-\\ p30\end{tabular}}    & ship     & 0.528          & \textbf{0.806} & 0.666          & 0             & \textbf{0.513} & 0.554          & -              & -                         & -                         & -                         \\
                                                                       & aircraft & \textbf{0.487} & 0.877          & 0.396          & nan           & 0.652          & \textbf{0.463} & -              & -                         & -                         & -                         \\
                                                                       & human    & 0.387          & 0.527          & 0.372          & 0.5           & \textbf{0.409} & 0.343          & -              & -                         & -                         & -                         \\
                                                                       & all      & 0.468          & \textbf{0.737} & 0.478          & 0.25          & \textbf{0.525} & 0.454          & \textbf{0.57}  & 0.25                      & \textbf{0.547}            & 0.632                     \\ \hline
\multirow{4}{*}{\begin{tabular}[c]{@{}c@{}}nei-\\ p5-50\end{tabular}}  & ship     & 0.547          & \textbf{0.788} & 0.679          & 0             & \textbf{0.509} & \textbf{0.584} & -              & -                         & -                         & -                         \\
                                                                       & aircraft & \textbf{0.484} & 0.866          & \textbf{0.505} & nan           & 0.651          & \textbf{0.456} & -              & -                         & -                         & -                         \\
                                                                       & human    & 0.389          & 0.534          & 0.377          & 0.4           & \textbf{0.401} & 0.414          & -              & -                         & -                         & -                         \\
                                                                       & all      & 0.473          & 0.729          & 0.521          & 0.2           & \textbf{0.52}  & 0.485          & \textbf{0.588} & 0.2                       & \textbf{0.54}             & \textbf{0.676}            \\ \hline
\end{tabular}
}
\caption{The SCTD dataset is detected using the SSDMobileNetV2 algorithm. ``noisy" represents the original noise dataset, and dspnet, b2ub, and nei represent the target detection after denoising using the DSPNet, Blind2Unblind, and Neighbor2Neighbor single-frame denoising algorithms, respectively. Bold fonts indicate data that is higher than the corresponding ``noisy" dataset.}

\label{table3}
\centering
\end{table*}

\begin{table*}
\centering
\resizebox{\textwidth}{!}{
\begin{tabular}{c|ccccccccccc}
\hline
\multirow{2}{*}{Denoising}                                             & \multicolumn{11}{c}{Faster R-CNN}                                                                                                                                                                                  \\ \cline{2-12} 
                                                                       & class    & mAP            & mAP@0.5        & mAP@0.75        & mAP\_s        & mAP\_m         & mAP\_l         & AR             & \multicolumn{1}{l}{AR\_s} & \multicolumn{1}{l}{AR\_m} & \multicolumn{1}{l}{AR\_l} \\ \hline
\multirow{4}{*}{noisy}                                                 & ship     & 0.405          & 0.716          & 0.4            & 0             & 0.466          & 0.415          & -              & -                         & -                         & -                         \\
                                                                       & aircraft & 0.23           & 0.453          & 0.192          & nan           & 0.7            & 0.149          & -              & -                         & -                         & -                         \\
                                                                       & human    & 0.257          & 0.492          & 0.076          & 0             & 0.203          & 0.495          & -              & -                         & -                         & -                         \\
                                                                       & all      & 0.297          & 0.554          & 0.222          & 0             & 0.456          & 0.353          & 0.408          & 0                         & 0.461                     & 0.502                     \\ \hline
\multirow{4}{*}{dspnet}                                                & ship     & 0.379          & 0.712          & 0.339          & 0             & 0.403          & 0.392          & -              & -                         & -                         & -                         \\
                                                                       & aircraft & 0.135          & \textbf{0.534} & 0.056          & nan           & 0.431          & 0.103          & -              & -                         & -                         & -                         \\
                                                                       & human    & 0.236          & 0.441          & \textbf{0.141} & \textbf{0.5}  & 0.173          & 0.374          & -              & -                         & -                         & -                         \\
                                                                       & all      & 0.25           & \textbf{0.563} & 0.179          & \textbf{0.25} & 0.336          & 0.29           & \textbf{0.419} & 0.25                      & 0.41                      & \textbf{0.511}            \\ \hline
\multirow{4}{*}{\begin{tabular}[c]{@{}c@{}}b2ub-\\ g25\end{tabular}}   & ship     & \textbf{0.419} & \textbf{0.72}  & \textbf{0.476} & 0             & \textbf{0.468} & \textbf{0.431} & -              & -                         & -                         & -                         \\
                                                                       & aircraft & 0.209          & 0.407          & \textbf{0.218} & nan           & 0.429          & \textbf{0.161} & -              & -                         & -                         & -                         \\
                                                                       & human    & \textbf{0.303} & \textbf{0.505} & \textbf{0.228} & 0             & \textbf{0.236} & \textbf{0.59}  & -              & -                         & -                         & -                         \\
                                                                       & all      & \textbf{0.311} & 0.544          & \textbf{0.307} & 0             & 0.378          & \textbf{0.394} & \textbf{0.445} & 0                         & 0.439                     & \textbf{0.549}            \\ \hline
\multirow{4}{*}{\begin{tabular}[c]{@{}c@{}}b2ub-\\ g5-50\end{tabular}} & ship     & \textbf{0.427} & \textbf{0.783} & \textbf{0.501} & 0             & 0.418          & \textbf{0.448} & -              & -                         & -                         & -                         \\
                                                                       & aircraft & 0.223          & 0.418          & \textbf{0.189} & nan           & 0.559          & \textbf{0.183} & -              & -                         & -                         & -                         \\
                                                                       & human    & 0.202          & 0.351          & \textbf{0.252} & 0             & \textbf{0.244} & 0.262          & -              & -                         & -                         & -                         \\
                                                                       & all      & \textbf{0.284} & 0.518          & \textbf{0.314} & 0             & 0.407          & 0.298          & \textbf{0.444} & 0                         & \textbf{0.513}            & \textbf{0.521}            \\ \hline
\multirow{4}{*}{\begin{tabular}[c]{@{}c@{}}b2ub-\\ p30\end{tabular}}   & ship     & \textbf{0.418} & \textbf{0.718} & \textbf{0.452} & 0             & 0.364          & \textbf{0.452} & -              & -                         & -                         & -                         \\
                                                                       & aircraft & 0.208          & 0.396          & 0.169          & nan           & 0.635          & \textbf{0.154} & -              & -                         & -                         & -                         \\
                                                                       & human    & \textbf{0.324} & \textbf{0.518} & \textbf{0.424} & 0             & \textbf{0.244} & \textbf{0.617} & -              & -                         & -                         & -                         \\
                                                                       & all      & \textbf{0.317} & 0.544          & \textbf{0.348} & 0             & 0.414          & \textbf{0.408} & \textbf{0.416} & 0                         & 0.452                     & \textbf{0.524}            \\ \hline
\multirow{4}{*}{\begin{tabular}[c]{@{}c@{}}b2ub-\\ p5-50\end{tabular}} & ship     & \textbf{0.414} & \textbf{0.708} & \textbf{0.469} & 0             & 0.37           & \textbf{0.439} & -              & -                         & -                         & -                         \\
                                                                       & aircraft & \textbf{0.243} & 0.449          & \textbf{0.282} & nan           & 0.65           & \textbf{0.167} & -              & -                         & -                         & -                         \\
                                                                       & human    & \textbf{0.291} & 0.458          & \textbf{0.337} & 0             & \textbf{0.284} & 0.455          & -              & -                         & -                         & -                         \\
                                                                       & all      & \textbf{0.316} & 0.538          & \textbf{0.362} & 0             & 0.435          & \textbf{0.354} & \textbf{0.41}  & 0                         & 0.432                     & 0.481                     \\ \hline
\multirow{4}{*}{\begin{tabular}[c]{@{}c@{}}nei-\\ g25\end{tabular}}    & ship     & \textbf{0.453} & \textbf{0.764} & \textbf{0.499} & 0             & \textbf{0.509} & \textbf{0.468} & -              & -                         & -                         & -                         \\
                                                                       & aircraft & \textbf{0.281} & \textbf{0.519} & \textbf{0.343} & nan           & \textbf{0.825} & \textbf{0.201} & -              & -                         & -                         & -                         \\
                                                                       & human    & \textbf{0.374} & \textbf{0.554} & \textbf{0.446} & 0             & \textbf{0.284} & \textbf{0.7}   & -              & -                         & -                         & -                         \\
                                                                       & all      & \textbf{0.369} & \textbf{0.613} & \textbf{0.429} & 0             & \textbf{0.54}  & \textbf{0.456} & \textbf{0.461} & 0                         & \textbf{0.549}            & \textbf{0.55}             \\ \hline
\multirow{4}{*}{\begin{tabular}[c]{@{}c@{}}nei-\\ g5-50\end{tabular}}  & ship     & 0.38           & \textbf{0.725} & 0.341          & 0             & 0.417          & 0.397          & -              & -                         & -                         & -                         \\
                                                                       & aircraft & 0.192          & 0.391          & 0.13           & nan           & \textbf{0.751} & 0.11           & -              & -                         & -                         & -                         \\
                                                                       & human    & \textbf{0.334} & \textbf{0.554} & \textbf{0.446} & 0             & \textbf{0.244} & \textbf{0.667} & -              & -                         & -                         & -                         \\
                                                                       & all      & \textbf{0.302} & \textbf{0.557} & \textbf{0.305} & 0             & \textbf{0.471} & \textbf{0.391} & \textbf{0.416} & 0                         & \textbf{0.53}             & 0.498                     \\ \hline
\multirow{4}{*}{\begin{tabular}[c]{@{}c@{}}nei-\\ p30\end{tabular}}    & ship     & \textbf{0.412} & \textbf{0.72}  & \textbf{0.416} & 0             & 0.444          & \textbf{0.427} & -              & -                         & -                         & -                         \\
                                                                       & aircraft & \textbf{0.287} & \textbf{0.653} & \textbf{0.316} & nan           & \textbf{0.8}   & \textbf{0.189} & -              & -                         & -                         & -                         \\
                                                                       & human    & \textbf{0.355} & \textbf{0.554} & \textbf{0.25}  & 0             & 0.203          & \textbf{0.789} & -              & -                         & -                         & -                         \\
                                                                       & all      & \textbf{0.351} & \textbf{0.642} & \textbf{0.327} & 0             & \textbf{0.482} & \textbf{0.468} & \textbf{0.431} & 0                         & \textbf{0.489}            & \textbf{0.557}            \\ \hline
\multirow{4}{*}{\begin{tabular}[c]{@{}c@{}}nei-\\ p5-50\end{tabular}}  & ship     & 0.36           & \textbf{0.748} & 0.258          & 0             & 0.353          & 0.38           & -              & -                         & -                         & -                         \\
                                                                       & aircraft & \textbf{0.288} & \textbf{0.466} & \textbf{0.381} & nan           & 0.584          & \textbf{0.239} & -              & -                         & -                         & -                         \\
                                                                       & human    & \textbf{0.334} & \textbf{0.554} & \textbf{0.337} & 0             & \textbf{0.284} & \textbf{0.534} & -              & -                         & -                         & -                         \\
                                                                       & all      & \textbf{0.414} & \textbf{0.589} & \textbf{0.325} & 0             & 0.407          & \textbf{0.384} & \textbf{0.414} & 0                         & \textbf{0.471}            & 0.473                     \\ \hline
\end{tabular}
}
\caption{The SCTD dataset is detected using the Faster R-CNN algorithm. ``noisy" represents the original noise dataset, and dspnet, b2ub, and nei represent the target detection after denoising using the DSPNet, Blind2Unblind, and Neighbor2Neighbor single-frame denoising algorithms, respectively. Bold fonts indicate data that is higher than the corresponding ``noisy" dataset.}

\label{table4}
\end{table*}

\begin{table*}
\centering
\resizebox{\textwidth}{!}{
\begin{tabular}{c|ccccccccccc}
\hline
\multirow{2}{*}{Denoising}                                             & \multicolumn{11}{c}{SSD300}                                                                                                                                                                                        \\ \cline{2-12} 
                                                                       & class    & mAP            & mAP@0.5        & mAP@0.75        & mAP\_s        & mAP\_m         & mAP\_l         & AR             & \multicolumn{1}{l}{AR\_s} & \multicolumn{1}{l}{AR\_m} & \multicolumn{1}{l}{AR\_l} \\ \hline
\multirow{4}{*}{noisy}                                                 & ship     & 0.498          & 0.775          & 0.604          & 0             & 0.495          & 0.522          & -              & -                         & -                         & -                         \\
                                                                       & aircraft & 0.488          & 0.862          & 0.592          & nan           & 0.403          & 0.5            & -              & -                         & -                         & -                         \\
                                                                       & human    & 0.398          & 0.632          & 0.632          & 0.6           & 0.244          & 0.645          & -              & -                         & -                         & -                         \\
                                                                       & all      & 0.461          & 0.756          & 0.61           & 0.3           & 0.38           & 0.556          & 0.543          & 0.3                       & 0.391                     & 0.669                     \\ \hline
\multirow{4}{*}{dspnet}                                                & ship     & 0.466          & 0.725          & 0.546          & 0             & 0.45           & 0.494          & -              & -                         & -                         & -                         \\
                                                                       & aircraft & \textbf{0.549} & \textbf{0.905} & 0.55           & nan           & \textbf{0.553} & \textbf{0.554} & -              & -                         & -                         & -                         \\
                                                                       & human    & \textbf{0.468} & \textbf{0.697} & 0.554          & 0.5           & \textbf{0.317} & \textbf{0.8}   & -              & -                         & -                         & -                         \\
                                                                       & all      & \textbf{0.494} & \textbf{0.776} & 0.55           & 0.25          & \textbf{0.44}  & \textbf{0.616} & \textbf{0.557} & 0.25                      & \textbf{0.46}             & 0.661                     \\ \hline
\multirow{4}{*}{\begin{tabular}[c]{@{}c@{}}b2ub-\\ g25\end{tabular}}   & ship     & 0.495          & 0.756          & 0.575          & 0             & 0.483          & 0.519          & -              & -                         & -                         & -                         \\
                                                                       & aircraft & 0.471          & \textbf{0.934} & 0.3            & nan           & \textbf{0.553} & 0.466          & -              & -                         & -                         & -                         \\
                                                                       & human    & \textbf{0.413} & \textbf{0.748} & 0.535          & 0.3           & \textbf{0.333} & \textbf{0.783} & -              & -                         & -                         & -                         \\
                                                                       & all      & 0.46           & \textbf{0.813} & 0.47           & 0.15          & \textbf{0.456} & \textbf{0.589} & \textbf{0.552} & 0.15                      & \textbf{0.481}            & 0.664                     \\ \hline
\multirow{4}{*}{\begin{tabular}[c]{@{}c@{}}b2ub-\\ g5-50\end{tabular}} & ship     & \textbf{0.511} & 0.77           & 0.577          & 0             & 0.474          & \textbf{0.539} & -              & -                         & -                         & -                         \\
                                                                       & aircraft & \textbf{0.504} & \textbf{0.873} & 0.558          & nan           & \textbf{0.552} & 0.5            & -              & -                         & -                         & -                         \\
                                                                       & human    & 0.368          & \textbf{0.68}  & 0.477          & 0.3           & \textbf{0.261} & \textbf{0.766} & -              & -                         & -                         & -                         \\
                                                                       & all      & 0.461          & \textbf{0.775} & 0.537          & 0.15          & \textbf{0.429} & \textbf{0.602} & \textbf{0.557} & 0.15                      & \textbf{0.487}            & 0.662                     \\ \hline
\multirow{4}{*}{\begin{tabular}[c]{@{}c@{}}b2ub-\\ p30\end{tabular}}   & ship     & 0.483          & 0.723          & 0.56           & 0             & \textbf{0.524} & 0.502          & -              & -                         & -                         & -                         \\
                                                                       & aircraft & \textbf{0.528} & \textbf{0.977} & \textbf{0.694} & nan           & 0.403          & \textbf{0.55}  & -              & -                         & -                         & -                         \\
                                                                       & human    & \textbf{0.466} & \textbf{0.733} & \textbf{0.663} & \textbf{0.7}  & \textbf{0.304} & \textbf{0.711} & -              & -                         & -                         & -                         \\
                                                                       & all      & \textbf{0.493} & \textbf{0.811} & \textbf{0.639} & \textbf{0.35} & \textbf{0.41}  & \textbf{0.587} & \textbf{0.547} & 0.35                      & \textbf{0.428}            & 0.648                     \\ \hline
\multirow{4}{*}{\begin{tabular}[c]{@{}c@{}}b2ub-\\ p5-50\end{tabular}} & ship     & 0.473          & 0.754          & 0.503          & 0             & 0.45           & 0.5            & -              & -                         & -                         & -                         \\
                                                                       & aircraft & \textbf{0.497} & \textbf{0.868} & 0.334          & nan           & \textbf{0.702} & 0.48           & -              & -                         & -                         & -                         \\
                                                                       & human    & 0.381          & \textbf{0.694} & 0.168          & 0.5           & 0.215          & \textbf{0.766} & -              & -                         & -                         & -                         \\
                                                                       & all      & 0.45           & \textbf{0.772} & 0.335          & 0.25          & \textbf{0.456} & \textbf{0.582} & 0.53           & 0.25                      & \textbf{0.474}            & 0.642                     \\ \hline
\multirow{4}{*}{\begin{tabular}[c]{@{}c@{}}nei-\\ g25\end{tabular}}    & ship     & \textbf{0.514} & \textbf{0.784} & 0.6            & 0             & \textbf{0.511} & \textbf{0.541} & -              & -                         & -                         & -                         \\
                                                                       & aircraft & \textbf{0.502} & \textbf{0.952} & 0.451          & nan           & \textbf{0.503} & \textbf{0.505} & -              & -                         & -                         & -                         \\
                                                                       & human    & 0.31           & 0.609          & 0.389          & 0.15          & \textbf{0.284} & 0.618          & -              & -                         & -                         & -                         \\
                                                                       & all      & \textbf{0.442} & \textbf{0.782} & 0.48           & 0.075         & \textbf{0.433} & 0.555          & 0.536          & 0.1                       & \textbf{0.438}            & 0.656                     \\ \hline
\multirow{4}{*}{\begin{tabular}[c]{@{}c@{}}nei-\\ g5-50\end{tabular}}  & ship     & 0.483          & 0.758          & 0.569          & 0             & 0.47           & 0.507          & -              & -                         & -                         & -                         \\
                                                                       & aircraft & 0.399          & 0.81           & 0.38           & nan           & \textbf{0.504} & 0.396          & -              & -                         & -                         & -                         \\
                                                                       & human    & 0.398          & \textbf{0.667} & 0.408          & 0.4           & \textbf{0.302} & \textbf{0.8}   & -              & -                         & -                         & -                         \\
                                                                       & all      & 0.427          & 0.745          & 0.452          & 0.2           & \textbf{0.425} & \textbf{0.568} & \textbf{0.548} & 0.2                       & \textbf{0.466}            & 0.654                     \\ \hline
\multirow{4}{*}{\begin{tabular}[c]{@{}c@{}}nei-\\ p30\end{tabular}}    & ship     & 0.497          & 0.751          & 0.51           & 0             & 0.48           & \textbf{0.528} & -              & -                         & -                         & -                         \\
                                                                       & aircraft & 0.431          & 0.813          & 0.332          & nan           & 0.403          & 0.435          & -              & -                         & -                         & -                         \\
                                                                       & human    & 0.378          & \textbf{0.677} & 0.462          & 0.3           & \textbf{0.299} & \textbf{0.655} & -              & -                         & -                         & -                         \\
                                                                       & all      & 0.435          & 0.747          & 0.435          & 0.15          & \textbf{0.394} & 0.54           & 0.529          & 0.15                      & \textbf{0.413}            & 0.632                     \\ \hline
\multirow{4}{*}{\begin{tabular}[c]{@{}c@{}}nei-\\ p5-50\end{tabular}}  & ship     & 0.481          & 0.77           & 0.599          & 0             & 0.494          & 0.506          & -              & -                         & -                         & -                         \\
                                                                       & aircraft & \textbf{0.492} & \textbf{0.901} & 0.434          & nan           & \textbf{0.552} & \textbf{0.507} & -              & -                         & -                         & -                         \\
                                                                       & human    & 0.347          & \textbf{0.663} & 0.168          & 0.4           & 0.203          & \textbf{0.711} & -              & -                         & -                         & -                         \\
                                                                       & all      & 0.44           & \textbf{0.778} & 0.401          & 0.2           & \textbf{0.417} & \textbf{0.575} & 0.52           & 0.2                       & \textbf{0.422}            & 0.644                     \\ \hline
\end{tabular}
}
\caption{The SCTD dataset is detected using the SSD300 algorithm. ``noisy" represents the original noise dataset, and dspnet, b2ub, and nei represent the target detection after denoising using the DSPNet, Blind2Unblind, and Neighbor2Neighbor single-frame denoising algorithms, respectively. Bold fonts indicate data that is higher than the corresponding ``noisy" dataset.}

\label{table5}
\end{table*}

\begin{table*}
\centering
\resizebox{\textwidth}{!}{
\begin{tabular}{c|cccccccclll}
\hline
\multirow{2}{*}{Denoising}                                             & \multicolumn{11}{c}{YOLOX}                                                                                                                                                                                                   \\ \cline{2-12} 
                                                                       & class    & mAP            & mAP@0.5        & mAP@0.75        & mAP\_s & mAP\_m         & mAP\_l         & AR             & AR\_s                    & AR\_m                              & AR\_l                              \\ \hline
\multirow{4}{*}{noisy}                                                 & ship     & 0.362          & 0.636          & 0.331          & 0      & 0.407          & 0.37           & -              & -                        & -                                  & -                                  \\
                                                                       & aircraft & 0.413          & 0.755          & 0.268          & nan    & 0.55           & 0.377          & -              & -                        & -                                  & -                                  \\
                                                                       & human    & 0.272          & 0.473          & 0.228          & 0.8    & 0.171          & 0.269          & -              & -                        & -                                  & -                                  \\
                                                                       & all      & 0.349          & 0.621          & 0.276          & 0.4    & 0.376          & 0.339          & 0.43           & \multicolumn{1}{c}{0.4}  & \multicolumn{1}{c}{0.404}          & \multicolumn{1}{c}{0.431}          \\ \hline
\multirow{4}{*}{dspnet}                                                & ship     & \textbf{0.367} & \textbf{0.699} & \textbf{0.401} & 0      & \textbf{0.449} & \textbf{0.377} & -              & -                        & -                                  & -                                  \\
                                                                       & aircraft & 0.361          & 0.704          & 0.267          & nan    & \textbf{0.75}  & 0.281          & -              & -                        & -                                  & -                                  \\
                                                                       & human    & 0.178          & 0.368          & 0.119          & 0.4    & 0.139          & \textbf{0.297} & -              & -                        & -                                  & -                                  \\
                                                                       & all      & 0.302          & 0.591          & 0.262          & 0.2    & \textbf{0.446} & 0.318          & 0.427          & \multicolumn{1}{c}{0.2}  & \multicolumn{1}{c}{\textbf{0.467}} & \multicolumn{1}{c}{\textbf{0.465}} \\ \hline
\multirow{4}{*}{\begin{tabular}[c]{@{}c@{}}b2ub-\\ g25\end{tabular}}   & ship     & \textbf{0.444} & \textbf{0.728} & \textbf{0.531} & 0      & \textbf{0.464} & \textbf{0.465} & -              & -                        & -                                  & -                                  \\
                                                                       & aircraft & \textbf{0.437} & \textbf{0.824} & \textbf{0.473} & nan    & \textbf{0.75}  & 0.37           & -              & -                        & -                                  & -                                  \\
                                                                       & human    & 0.272          & \textbf{0.487} & 0.228          & 0.6    & \textbf{0.203} & \textbf{0.301} & -              & -                        & -                                  & -                                  \\
                                                                       & all      & \textbf{0.384} & \textbf{0.68}  & \textbf{0.411} & 0.3    & \textbf{0.472} & \textbf{0.379} & \textbf{0.475} & \multicolumn{1}{c}{0.3}  & \multicolumn{1}{c}{\textbf{0.489}} & \multicolumn{1}{c}{\textbf{0.502}} \\ \hline
\multirow{4}{*}{\begin{tabular}[c]{@{}c@{}}b2ub-\\ g5-50\end{tabular}} & ship     & \textbf{0.416} & \textbf{0.727} & \textbf{0.398} & 0      & \textbf{0.473} & \textbf{0.424} & -              & -                        & -                                  & -                                  \\
                                                                       & aircraft & 0.409          & 0.69           & \textbf{0.325} & nan    & \textbf{0.752} & 0.341          & -              & -                        & -                                  & -                                  \\
                                                                       & human    & 0.258          & 0.446          & \textbf{0.282} & 0.3    & \textbf{0.284} & 0.269          & -              & -                        & -                                  & -                                  \\
                                                                       & all      & \textbf{0.361} & 0.621          & \textbf{0.335} & 0.15   & \textbf{0.503} & \textbf{0.345} & 0.426          & \multicolumn{1}{c}{0.15} & \multicolumn{1}{c}{\textbf{0.516}} & \multicolumn{1}{c}{0.413}          \\ \hline
\multirow{4}{*}{\begin{tabular}[c]{@{}c@{}}b2ub-\\ p30\end{tabular}}   & ship     & \textbf{0.373} & \textbf{0.704} & \textbf{0.408} & 0      & 0.365          & \textbf{0.393} & -              & -                        & -                                  & -                                  \\
                                                                       & aircraft & 0.373          & 0.737          & \textbf{0.369} & nan    & \textbf{0.701} & 0.313          & -              & -                        & -                                  & -                                  \\
                                                                       & human    & \textbf{0.345} & \textbf{0.648} & \textbf{0.424} & 0.7    & \textbf{0.244} & \textbf{0.395} & -              & -                        & -                                  & -                                  \\
                                                                       & all      & \textbf{0.364} & \textbf{0.696} & \textbf{0.4}   & 0.35   & \textbf{0.436} & \textbf{0.367} & \textbf{0.474} & \multicolumn{1}{c}{0.35} & \multicolumn{1}{c}{\textbf{0.447}} & \multicolumn{1}{c}{\textbf{0.51}}  \\ \hline
\multirow{4}{*}{\begin{tabular}[c]{@{}c@{}}b2ub-\\ p5-50\end{tabular}} & ship     & \textbf{0.429} & \textbf{0.718} & \textbf{0.444} & 0      & \textbf{0.479} & \textbf{0.438} & -              & -                        & -                                  & -                                  \\
                                                                       & aircraft & 0.408          & \textbf{0.821} & 0.254          & nan    & 0.525          & \textbf{0.39}  & -              & -                        & -                                  & -                                  \\
                                                                       & human    & 0.189          & 0.361          & 0.119          & 0.1    & \textbf{0.249} & \textbf{0.303} & -              & -                        & -                                  & -                                  \\
                                                                       & all      & 0.342          & \textbf{0.633} & 0.272          & 0.05   & \textbf{0.418} & \textbf{0.377} & \textbf{0.439} & \multicolumn{1}{c}{0.1}  & \multicolumn{1}{c}{\textbf{0.453}} & \multicolumn{1}{c}{\textbf{0.455}} \\ \hline
\multirow{4}{*}{\begin{tabular}[c]{@{}c@{}}nei-\\ g25\end{tabular}}    & ship     & \textbf{0.467} & \textbf{0.728} & \textbf{0.584} & 0      & \textbf{0.565} & \textbf{0.48}  & -              & -                        & -                                  & -                                  \\
                                                                       & aircraft & \textbf{0.498} & \textbf{0.872} & \textbf{0.587} & nan    & \textbf{0.65}  & \textbf{0.472} & -              & -                        & -                                  & -                                  \\
                                                                       & human    & \textbf{0.303} & 0.446          & \textbf{0.282} & 0.5    & \textbf{0.284} & \textbf{0.303} & -              & -                        & -                                  & -                                  \\
                                                                       & all      & \textbf{0.423} & \textbf{0.682} & \textbf{0.485} & 0.25   & \textbf{0.5}   & \textbf{0.418} & \textbf{0.482} & \multicolumn{1}{c}{0.25} & \multicolumn{1}{c}{\textbf{0.516}} & \multicolumn{1}{c}{\textbf{0.478}} \\ \hline
\multirow{4}{*}{\begin{tabular}[c]{@{}c@{}}nei-\\ g5-50\end{tabular}}  & ship     & \textbf{0.423} & \textbf{0.723} & \textbf{0.427} & 0      & \textbf{0.515} & \textbf{0.436} & -              & -                        & -                                  & -                                  \\
                                                                       & aircraft & \textbf{0.523} & \textbf{0.907} & \textbf{0.526} & nan    & \textbf{0.75}  & \textbf{0.482} & -              & -                        & -                                  & -                                  \\
                                                                       & human    & 0.211          & 0.38           & 0.119          & 0.4    & 0.162          & \textbf{0.303} & -              & -                        & -                                  & -                                  \\
                                                                       & all      & \textbf{0.386} & \textbf{0.67}  & \textbf{0.357} & 0.2    & \textbf{0.476} & \textbf{0.407} & \textbf{0.434} & \multicolumn{1}{c}{0.2}  & \multicolumn{1}{c}{\textbf{0.476}} & \multicolumn{1}{c}{\textbf{0.451}} \\ \hline
\multirow{4}{*}{\begin{tabular}[c]{@{}c@{}}nei-\\ p30\end{tabular}}    & ship     & \textbf{0.423} & \textbf{0.683} & \textbf{0.573} & 0      & \textbf{0.507} & \textbf{0.436} & -              & -                        & -                                  & -                                  \\
                                                                       & aircraft & \textbf{0.492} & \textbf{0.893} & \textbf{0.515} & nan    & \textbf{0.65}  & \textbf{0.459} & -              & -                        & -                                  & -                                  \\
                                                                       & human    & 0.229          & 0.446          & 0.119          & 0.433  & 0.122          & \textbf{0.337} & -              & -                        & -                                  & -                                  \\
                                                                       & all      & \textbf{0.381} & \textbf{0.674} & \textbf{0.402} & 0.217  & \textbf{0.427} & \textbf{0.411} & \textbf{0.445} & \multicolumn{1}{c}{0.25} & \multicolumn{1}{c}{\textbf{0.429}} & \multicolumn{1}{c}{\textbf{0.482}} \\ \hline
\multirow{4}{*}{\begin{tabular}[c]{@{}c@{}}nei-\\ p5-50\end{tabular}}  & ship     & 0.362          & \textbf{0.677} & \textbf{0.401} & 0      & 0.401          & \textbf{0.381} & -              & -                        & -                                  & -                                  \\
                                                                       & aircraft & \textbf{0.465} & \textbf{0.888} & \textbf{0.396} & nan    & \textbf{0.703} & \textbf{0.434} & -              & -                        & -                                  & -                                  \\
                                                                       & human    & 0.211          & 0.446          & 0.212          & 0.3    & \textbf{0.223} & 0.202          & -              & -                        & -                                  & -                                  \\
                                                                       & all      & 0.346          & \textbf{0.67}  & \textbf{0.336} & 0.15   & \textbf{0.442} & 0.339          & 0.423          & \multicolumn{1}{c}{0.15} & \multicolumn{1}{c}{\textbf{0.486}} & \multicolumn{1}{c}{0.409}          \\ \hline
\end{tabular}
}
\caption{The SCTD dataset is detected using the YOLOX algorithm. ``noisy" represents the original noise dataset, and dspnet, b2ub, and nei represent the target detection after denoising using the DSPNet, Blind2Unblind, and Neighbor2Neighbor single-frame denoising algorithms, respectively. Bold fonts indicate data that is higher than the corresponding ``noisy" dataset.}

\label{table6}
\end{table*}

\begin{table*}
\centering
\resizebox{\textwidth}{!}{
\begin{tabular}{c|ccccccccccc}
\hline
\multirow{2}{*}{Denoising} & \multicolumn{11}{c}{SSDMobileNetV2}                                                                                                                                                                                 \\ \cline{2-12} 
                           & class      & mAP            & mAP@0.5        & mAP@0.75        & mAP\_s         & mAP\_m         & mAP\_l         & AR             & \multicolumn{1}{l}{AR\_s} & \multicolumn{1}{l}{AR\_m} & \multicolumn{1}{l}{AR\_l} \\ \hline
noisy                      & object/all & 0.186          & 0.392          & 0.158          & 0.004          & 0.024          & 0.296          & 0.297          & 0.047                     & 0.158                     & 0.414                     \\ \hline
dspnet                     & object/all & 0.177          & 0.381          & 0.152          & \textbf{0.014} & \textbf{0.027} & 0.295          & 0.293          & \textbf{0.058}            & \textbf{0.16}             & 0.405                     \\ \hline
b2ub-g25                   & object/all & 0.176          & 0.389          & \textbf{0.163} & \textbf{0.021} & 0.021          & 0.291          & 0.296          & \textbf{0.062}            & \textbf{0.161}            & 0.408                     \\ \hline
b2ub-g5-50                 & object/all & 0.186          & 0.39           & \textbf{0.163} & \textbf{0.011} & 0.019          & \textbf{0.298} & 0.294          & \textbf{0.077}            & 0.145                     & 0.411                     \\ \hline
b2ub-p30                   & object/all & 0.171          & 0.364          & 0.15           & \textbf{0.005} & 0.021          & \textbf{0.301} & 0.296          & 0.046                     & 0.158                     & 0.413                     \\ \hline
b2ub-p5-50                 & object/all & 0.177          & 0.388          & 0.145          & \textbf{0.017} & \textbf{0.027} & \textbf{0.3}   & \textbf{0.302} & \textbf{0.067}            & \textbf{0.172}            & 0.411                     \\ \hline
nei-g25                    & object/all & \textbf{0.191} & \textbf{0.396} & \textbf{0.17}  & \textbf{0.006} & 0.015          & \textbf{0.308} & \textbf{0.299} & \textbf{0.061}            & \textbf{0.159}            & \textbf{0.415}            \\ \hline
nei-g5-50                  & object/all & 0.184          & 0.39           & \textbf{0.162} & \textbf{0.014} & 0.014          & \textbf{0.305} & \textbf{0.299} & \textbf{0.064}            & 0.15                      & \textbf{0.418}            \\ \hline
nei-p30                    & object/all & 0.18           & 0.392          & 0.152          & \textbf{0.016} & 0.023          & \textbf{0.304} & \textbf{0.302} & \textbf{0.07}             & \textbf{0.159}            & \textbf{0.418}            \\ \hline
nei-p5-50                  & object/all & 0.175          & 0.382          & 0.141          & \textbf{0.026} & \textbf{0.026} & 0.293          & \textbf{0.298} & \textbf{0.073}            & \textbf{0.166}            & 0.407                     \\ \hline
\end{tabular}
}
\caption{The URPC\_sonarimage\_dataset dataset is detected using the SSDMobileNetV2 algorithm. ``noisy" represents the original noise dataset, and dspnet, b2ub, and nei represent the target detection after denoising using the DSPNet, Blind2Unblind, and Neighbor2Neighbor single-frame denoising algorithms, respectively. Bold fonts indicate data that is higher than the corresponding ``noisy" dataset.}

\label{table7}
\end{table*}

\begin{table*}
\centering
\resizebox{\textwidth}{!}{
\begin{tabular}{c|ccccccccccc}
\hline
\multirow{2}{*}{Denoising} & \multicolumn{11}{c}{Faster R-CNN}                                                                                                                                                             \\ \cline{2-12} 
                           & class      & mAP   & mAP@0.5        & mAP@0.75 & mAP\_s & mAP\_m         & mAP\_l         & AR             & \multicolumn{1}{l}{AR\_s} & \multicolumn{1}{l}{AR\_m} & \multicolumn{1}{l}{AR\_l} \\ \hline
noisy                      & object/all & 0.271 & 0.562          & 0.243   & 0.155  & 0.104          & 0.368          & 0.365          & 0.222                     & 0.264                     & 0.443                     \\ \hline
dspnet                     & object/all & 0.262 & 0.546          & 0.226   & 0.11   & 0.1            & 0.365          & 0.353          & 0.221                     & 0.232                     & 0.44                      \\ \hline
b2ub-g25                   & object/all & 0.248 & 0.523          & 0.214   & 0.078  & 0.085          & 0.353          & 0.354          & 0.141                     & 0.242                     & \textbf{0.449}            \\ \hline
b2ub-g5-50                 & object/all & 0.264 & 0.553          & 0.232   & 0.127  & 0.099          & \textbf{0.369} & 0.362          & \textbf{0.225}            & 0.236                     & \textbf{0.452}            \\ \hline
b2ub-p30                   & object/all & 0.271 & \textbf{0.572} & 0.235   & 0.146  & \textbf{0.111} & 0.365          & 0.365          & \textbf{0.251}            & 0.26                      & 0.439                     \\ \hline
b2ub-p5-50                 & object/all & 0.269 & \textbf{0.567} & 0.226   & 0.121  & \textbf{0.107} & \textbf{0.375} & 0.365          & 0.213                     & 0.246                     & \textbf{0.454}            \\ \hline
nei-g25                    & object/all & 0.266 & 0.55           & 0.23    & 0.105  & 0.104          & \textbf{0.371} & 0.354          & 0.183                     & 0.232                     & \textbf{0.448}            \\ \hline
nei-g5-50                  & object/all & 0.261 & 0.548          & 0.219   & 0.119  & 0.094          & \textbf{0.37}  & 0.363          & \textbf{0.223}            & 0.225                     & \textbf{0.459}            \\ \hline
nei-p30                    & object/all & 0.271 & \textbf{0.565} & 0.235   & 0.147  & 0.103          & \textbf{0.372} & \textbf{0.369} & \textbf{0.248}            & 0.247                     & \textbf{0.453}            \\ \hline
nei-p5-50                  & object/all & 0.26  & 0.56           & 0.23    & 0.125  & 0.101          & 0.365          & 0.35           & 0.204                     & 0.229                     & 0.439                     \\ \hline
\end{tabular}
}
\caption{The URPC\_sonarimage\_dataset dataset is detected using the Faster R-CNN algorithm. ``noisy" represents the original noise dataset, and dspnet, b2ub, and nei represent the target detection after denoising using the DSPNet, Blind2Unblind, and Neighbor2Neighbor single-frame denoising algorithms, respectively. Bold fonts indicate data that is higher than the corresponding ``noisy" dataset.}

\label{table8}
\end{table*}

\begin{table*}
\centering
\resizebox{\textwidth}{!}{
\begin{tabular}{c|ccccccccccc}
\hline
\multirow{2}{*}{Denoising} & \multicolumn{11}{c}{SSD300}                                                                                                                                                                                           \\ \cline{2-12} 
                           & class      & mAP            & mAP@0.5        & mAP@0.75        & mAP\_s         & mAP\_m         & mAP\_l         & AR             & \multicolumn{1}{l}{AR\_s} & \multicolumn{1}{l}{AR\_m} & \multicolumn{1}{l}{AR\_l} \\ \hline
noisy                      & object/all & 0.217          & 0.491          & 0.178          & 0.071          & 0.059          & 0.312          & 0.352          & 0.18                      & 0.264                     & 0.427                     \\ \hline
dspnet                     & object/all & 0.213          & 0.468          & \textbf{0.188} & 0.065          & 0.055          & 0.309          & 0.345          & 0.163                     & \textbf{0.272}            & 0.415                     \\ \hline
b2ub-g25                   & object/all & 0.215          & 0.485          & 0.17           & 0.06           & \textbf{0.063} & 0.308          & 0.344          & 0.14                      & 0.259                     & 0.425                     \\ \hline
b2ub-g5-50                 & object/all & \textbf{0.219} & \textbf{0.496} & 0.177          & 0.066          & \textbf{0.066} & \textbf{0.313} & \textbf{0.359} & \textbf{0.196}            & \textbf{0.269}            & \textbf{0.434}            \\ \hline
b2ub-p30                   & object/all & 0.218          & \textbf{0.502} & 0.159          & 0.069          & \textbf{0.063} & \textbf{0.313} & \textbf{0.355} & \textbf{0.19}             & \textbf{0.278}            & 0.425                     \\ \hline
b2ub-p5-50                 & object/all & 0.137          & 0.346          & 0.095          & 0.05           & 0.042          & 0.201          & 0.291          & 0.136                     & 0.242                     & 0.343                     \\ \hline
nei-g25                    & object/all & 0.216          & \textbf{0.502} & \textbf{0.179} & \textbf{0.078} & \textbf{0.062} & 0.311          & 0.349          & \textbf{0.192}            & \textbf{0.268}            & 0.42                      \\ \hline
nei-g5-50                  & object/all & \textbf{0.218} & 0.488          & 0.173          & 0.066          & \textbf{0.062} & \textbf{0.314} & 0.35           & 0.15                      & \textbf{0.274}            & 0.425                     \\ \hline
nei-p30                    & object/all & \textbf{0.223} & \textbf{0.518} & 0.18           & \textbf{0.077} & \textbf{0.071} & \textbf{0.314} & \textbf{0.36}  & \textbf{0.196}            & \textbf{0.283}            & \textbf{0.428}            \\ \hline
nei-p5-50                  & object/all & \textbf{0.219} & \textbf{0.492} & 0.178          & \textbf{0.081} & 0.056          & \textbf{0.318} & \textbf{0.355} & \textbf{0.23}             & \textbf{0.273}            & 0.42                      \\ \hline
\end{tabular}
} 
\caption{The URPC\_sonarimage\_dataset dataset is detected using the SSD300 algorithm. ``noisy" represents the original noise dataset, and dspnet, b2ub, and nei represent the target detection after denoising using the DSPNet, Blind2Unblind, and Neighbor2Neighbor single-frame denoising algorithms, respectively. Bold fonts indicate data that is higher than the corresponding ``noisy" dataset.}

\label{table9}
\end{table*}

\begin{table*}
\centering
\resizebox{\textwidth}{!}{
\begin{tabular}{c|ccccccccccc}
\hline
\multirow{2}{*}{Denoising} & \multicolumn{11}{c}{YOLOX}                                                                                                                                                                                            \\ \cline{2-12} 
                           & class      & mAP            & mAP\_50        & mAP\_75        & mAP\_s         & mAP\_m         & mAP\_l         & AR             & \multicolumn{1}{l}{AR\_s} & \multicolumn{1}{l}{AR\_m} & \multicolumn{1}{l}{AR\_l} \\ \hline
noisy                      & object/all & 0.278          & 0.582          & 0.231          & 0.091          & 0.114          & 0.381          & 0.376          & 0.158                     & 0.285                     & 0.463                     \\ \hline
dspnet                     & object/all & 0.264          & 0.551          & \textbf{0.232} & \textbf{0.102} & 0.091          & 0.367          & 0.354          & 0.151                     & 0.257                     & 0.441                     \\ \hline
b2ub-g25                   & object/all & \textbf{0.281} & \textbf{0.584} & \textbf{0.242} & \textbf{0.129} & 0.107          & \textbf{0.383} & \textbf{0.383} & \textbf{0.2}              & \textbf{0.288}            & \textbf{0.466}            \\ \hline
b2ub-g5-50                 & object/all & 0.274          & 0.579          & \textbf{0.236} & \textbf{0.104} & \textbf{0.116} & 0.375          & 0.369          & 0.152                     & 0.273                     & 0.457                     \\ \hline
b2ub-p30                   & object/all & 0.275          & \textbf{0.585} & 0.226          & \textbf{0.097} & 0.107          & 0.381          & 0.367          & 0.15                      & 0.264                     & 0.459                     \\ \hline
b2ub-p5-50                 & object/all & 0.275          & \textbf{0.586} & \textbf{0.233} & \textbf{0.098} & \textbf{0.12}  & 0.375          & 0.371          & 0.142                     & \textbf{0.286}            & 0.456                     \\ \hline
nei-g25                    & object/all & 0.27           & 0.564          & 0.223          & \textbf{0.103} & 0.095          & 0.372          & 0.369          & 0.152                     & 0.283                     & 0.452                     \\ \hline
nei-g5-50                  & object/all & 0.272          & 0.567          & \textbf{0.236} & \textbf{0.112} & 0.098          & 0.372          & 0.37           & \textbf{0.166}            & \textbf{0.286}            & 0.45                      \\ \hline
nei-p30                    & object/all & 0.272          & \textbf{0.589} & 0.22           & \textbf{0.111} & 0.112          & 0.369          & 0.373          & \textbf{0.165}            & \textbf{0.3}              & 0.448                     \\ \hline
nei-p5-50                  & object/all & 0.268          & 0.57           & 0.223          & \textbf{0.098} & 0.1            & 0.373          & 0.366          & 0.138                     & 0.27                      & 0.457                     \\ \hline
\end{tabular}
}
\caption{The URPC\_sonarimage\_dataset dataset is detected using the YOLOX algorithm.  ``noisy" represents the original noise dataset, and dspnet, b2ub, and nei represent the target detection after denoising using the DSPNet, Blind2Unblind, and Neighbor2Neighbor single-frame denoising algorithms, respectively. Bold fonts indicate data that is higher than the corresponding ``noisy" dataset.}
\label{table10}
\end{table*}

\begin{table*}
\centering
\resizebox{\textwidth}{!}{
\begin{tabular}{c|ccccccccccc}
\hline
\multirow{2}{*}{Denoising}                                              & \multicolumn{11}{c}{SSDMobileNetV2}                                                                                                                                                                                      \\ \cline{2-12} 
                                                                        & class           & mAP            & mAP\_50        & mAP\_75        & mAP\_s         & mAP\_m         & mAP\_l         & AR             & \multicolumn{1}{l}{AR\_s} & \multicolumn{1}{l}{AR\_m} & \multicolumn{1}{l}{AR\_l} \\ \hline
\multirow{12}{*}{noisy}                                                 & Bottle          & 0.739          & 0.981          & 0.948          & 0.2            & 0.735          & nan            & -              & -                         & -                         & -                         \\
                                                                        & Can             & 0.34           & 0.652          & 0.314          & 0.348          & 0.352          & nan            & -              & -                         & -                         & -                         \\
                                                                        & Chain           & 0.687          & 0.949          & 0.834          & 0.217          & 0.457          & 0.738          & -              & -                         & -                         & -                         \\
                                                                        & Drink-carton    & 0.567          & 0.887          & 0.699          & 0.595          & 0.537          & nan            & -              & -                         & -                         & -                         \\
                                                                        & Hook            & 0.727          & 0.956          & 0.956          & nan            & 0.727          & nan            & -              & -                         & -                         & -                         \\
                                                                        & Propeller       & 0.708          & 0.989          & 0.882          & 0.5            & 0.715          & 0.8            & -              & -                         & -                         & -                         \\
                                                                        & Shampoo-bottle  & 0.71           & 0.968          & 0.968          & nan            & 0.708          & 0.8            & -              & -                         & -                         & -                         \\
                                                                        & Standing-bottle & 0.788          & 1              & 1              & nan            & 0.788          & nan            & -              & -                         & -                         & -                         \\
                                                                        & Tire            & 0.805          & 0.99           & 0.955          & 0.8            & 0.81           & 0.3            & -              & -                         & -                         & -                         \\
                                                                        & Valve           & 0.725          & 0.927          & 0.901          & 0.383          & 0.751          & 0              & -              & -                         & -                         & -                         \\
                                                                        & Wall            & 0.713          & 0.916          & 0.841          & 0.304          & 0.7            & 0.84           & -              & -                         & -                         & -                         \\
                                                                        & all             & 0.683          & 0.929          & 0.845          & 0.418          & 0.662          & 0.58           & 0.743          & 0.571                     & 0.729                     & 0.596                     \\ \hline
\multirow{12}{*}{dspnet}                                                & Bottle          & \textbf{0.749} & 0.979          & 0.917          & \textbf{0.3}   & \textbf{0.746} & nan            & -              & -                         & -                         & -                         \\
                                                                        & Can             & 0.331          & 0.565          & \textbf{0.358} & \textbf{0.407} & 0.339          & nan            & -              & -                         & -                         & -                         \\
                                                                        & Chain           & \textbf{0.691} & 0.946          & \textbf{0.861} & \textbf{0.6}   & \textbf{0.478} & 0.732          & -              & -                         & -                         & -                         \\
                                                                        & Drink-carton    & 0.555          & \textbf{0.955} & 0.556          & 0.544          & \textbf{0.588} & nan            & -              & -                         & -                         & -                         \\
                                                                        & Hook            & 0.693          & 0.947          & 0.834          & nan            & 0.693          & nan            & -              & -                         & -                         & -                         \\
                                                                        & Propeller       & 0.633          & 0.98           & 0.747          & 0.4            & 0.642          & 0.6            & -              & -                         & -                         & -                         \\
                                                                        & Shampoo-bottle  & 0.681          & 0.957          & 0.957          & nan            & 0.681          & 0.7            & -              & -                         & -                         & -                         \\
                                                                        & Standing-bottle & 0.786          & 1              & 1              & nan            & 0.786          & nan            & -              & -                         & -                         & -                         \\
                                                                        & Tire            & 0.797          & 0.99           & 0.936          & 0.6            & 0.803          & \textbf{0.5}   & -              & -                         & -                         & -                         \\
                                                                        & Valve           & 0.701          & 0.917          & 0.847          & 0              & 0.733          & 0              & -              & -                         & -                         & -                         \\
                                                                        & Wall            & \textbf{0.715} & \textbf{0.933} & 0.805          & \textbf{0.365} & 0.693          & 0.832          & -              & -                         & -                         & -                         \\
                                                                        & all             & 0.667          & 0.924          & 0.802          & 0.402          & 0.653          & 0.561          & 0.729          & 0.469                     & 0.724                     & 0.575                     \\ \hline
\multirow{12}{*}{\begin{tabular}[c]{@{}c@{}}b2ub-\\ g25\end{tabular}}   & Bottle          & \textbf{0.745} & 0.981          & \textbf{0.952} & 0.2            & \textbf{0.74}  & nan            & -              & -                         & -                         & -                         \\
                                                                        & Can             & 0.336          & 0.617          & 0.298          & 0.335          & \textbf{0.356} & nan            & -              & -                         & -                         & -                         \\
                                                                        & Chain           & \textbf{0.688} & \textbf{0.952} & \textbf{0.847} & \textbf{0.317} & 0.425          & 0.736          & -              & -                         & -                         & -                         \\
                                                                        & Drink-carton    & \textbf{0.572} & \textbf{0.951} & 0.599          & 0.588          & \textbf{0.565} & nan            & -              & -                         & -                         & -                         \\
                                                                        & Hook            & 0.721          & 0.95           & 0.929          & nan            & 0.721          & nan            & -              & -                         & -                         & -                         \\
                                                                        & Propeller       & 0.678          & 0.984          & 0.83           & 0.5            & 0.684          & 0.7            & -              & -                         & -                         & -                         \\
                                                                        & Shampoo-bottle  & \textbf{0.724} & 0.951          & 0.951          & nan            & \textbf{0.716} & 0.9            & -              & -                         & -                         & -                         \\
                                                                        & Standing-bottle & 0.782          & 1              & 1              & nan            & 0.782          & nan            & -              & -                         & -                         & -                         \\
                                                                        & Tire            & 0.803          & 0.99           & 0.944          & 0.6            & 0.81           & 0.3            & -              & -                         & -                         & -                         \\
                                                                        & Valve           & 0.691          & 0.919          & 0.873          & 0.2            & 0.729          & 0              & -              & -                         & -                         & -                         \\
                                                                        & Wall            & 0.699          & \textbf{0.919} & 0.813          & 0.28           & 0.688          & 0.827          & -              & -                         & -                         & -                         \\
                                                                        & all             & 0.676          & 0.929          & 0.822          & 0.377          & 0.656          & 0.577          & 0.74           & 0.475                     & \textbf{0.73}             & 0.593                     \\ \hline
\multirow{12}{*}{\begin{tabular}[c]{@{}c@{}}b2ub-\\ g5-50\end{tabular}} & Bottle          & \textbf{0.746} & 0.976          & 0.943          & 0.2            & \textbf{0.742} & nan            & -              & -                         & -                         & -                         \\
                                                                        & Can             & \textbf{0.346} & 0.629          & \textbf{0.39}  & 0.384          & \textbf{0.355} & nan            & -              & -                         & -                         & -                         \\
                                                                        & Chain           & \textbf{0.693} & \textbf{0.95}  & \textbf{0.837} & \textbf{0.65}  & \textbf{0.489} & 0.728          & -              & -                         & -                         & -                         \\
                                                                        & Drink-carton    & 0.551          & \textbf{0.93}  & 0.664          & 0.585          & 0.516          & nan            & -              & -                         & -                         & -                         \\
                                                                        & Hook            & 0.727          & 0.944          & 0.944          & nan            & 0.727          & nan            & -              & -                         & -                         & -                         \\
                                                                        & Propeller       & 0.68           & 0.985          & 0.836          & 0.5            & 0.685          & 0.7            & -              & -                         & -                         & -                         \\
                                                                        & Shampoo-bottle  & \textbf{0.722} & 0.968          & 0.968          & nan            & \textbf{0.725} & 0.7            & -              & -                         & -                         & -                         \\
                                                                        & Standing-bottle & \textbf{0.798} & 1              & 0.871          & nan            & \textbf{0.798} & nan            & -              & -                         & -                         & -                         \\
                                                                        & Tire            & \textbf{0.815} & 0.989          & 0.958          & 0.8            & \textbf{0.823} & 0              & -              & -                         & -                         & -                         \\
                                                                        & Valve           & 0.7            & 0.911          & 0.886          & 0              & 0.735          & 0              & -              & -                         & -                         & -                         \\
                                                                        & Wall            & \textbf{0.715} & \textbf{0.93}  & 0.828          & 0.304          & \textbf{0.701} & \textbf{0.852} & -              & -                         & -                         & -                         \\
                                                                        & all             & 0.681          & 0.928          & 0.83           & \textbf{0.428} & \textbf{0.663} & 0.497          & \textbf{0.744} & 0.505                     & \textbf{0.737}            & 0.511                     \\ \hline
\end{tabular}
}
\caption{The MarinDebris dataset is detected using the SSDMobileNetV2 algorithm (Part 1 of 3).  ``noisy" represents the original noise dataset, and dspnet, b2ub, and nei represent the target detection after denoising using the DSPNet, Blind2Unblind, and Neighbor2Neighbor single-frame denoising algorithms, respectively. Bold fonts indicate data that is higher than the corresponding ``noisy" dataset.}
\label{table11}
\end{table*}

\begin{table*}
\centering
\resizebox{\textwidth}{!}{
\begin{tabular}{cccccccccccc}
\hline
\multicolumn{1}{c|}{\multirow{12}{*}{\begin{tabular}[c]{@{}c@{}}b2ub-\\ p30\end{tabular}}}   & Bottle          & \textbf{0.749} & \textbf{0.984} & \textbf{0.973} & \textbf{0.35}  & 0.743          & nan            & -              & -                         & -                         & -                         \\
\multicolumn{1}{c|}{}                                                                         & Can             & \textbf{0.371} & 0.649          & \textbf{0.357} & \textbf{0.375} & \textbf{0.388} & nan            & -              & -                         & -                         & -                         \\
\multicolumn{1}{c|}{}                                                                         & Chain           & 0.683          & \textbf{0.95}  & 0.825          & \textbf{0.601} & 0.44           & 0.723          & -              & -                         & -                         & -                         \\
\multicolumn{1}{c|}{}                                                                         & Drink-carton    & 0.547          & \textbf{0.931} & 0.558          & 0.527          & \textbf{0.59}  & nan            & -              & -                         & -                         & -                         \\
\multicolumn{1}{c|}{}                                                                         & Hook            & \textbf{0.739} & 0.951          & 0.951          & nan            & \textbf{0.739} & nan            & -              & -                         & -                         & -                         \\
 \multicolumn{1}{c|}{}                                                                        & Propeller       & 0.689          & \textbf{0.991} & \textbf{0.895} & \textbf{0.7}   & 0.692          & 0.7            & -              & -                         & -                         & -                         \\
\multicolumn{1}{c|}{}                                                                         & Shampoo-bottle  & \textbf{0.738} & 0.958          & 0.958          & nan            & \textbf{0.735} & 0.8            & -              & -                         & -                         & -                         \\
\multicolumn{1}{c|}{}                                                                         & Standing-bottle & \textbf{0.791} & 1              & 1              & nan            & \textbf{0.791} & nan            & -              & -                         & -                         & -                         \\
 \multicolumn{1}{c|}{}                                                                        & Tire            & 0.804          & 0.99           & 0.947          & 0.6            & \textbf{0.812} & 0.2            & -              & -                         & -                         & -                         \\
\multicolumn{1}{c|}{}                                                                         & Valve           & 0.687          & 0.906          & 0.877          & 0              & 0.725          & 0              & -              & -                         & -                         & -                         \\
\multicolumn{1}{c|}{}                                                                         & Wall            & \textbf{0.723} & \textbf{0.926} & \textbf{0.852} & 0.325          & \textbf{0.716} & 0.835          & -              & -                         & -                         & -                         \\
\multicolumn{1}{c|}{}                                                                         & all             & 0.684          & \textbf{0.93}  & 0.836          & \textbf{0.435} & \textbf{0.67}  & 0.543          & \textbf{0.744} & 0.503                     & \textbf{0.741}            & 0.559                     \\ \hline
\multicolumn{1}{c|}{\multirow{12}{*}{\begin{tabular}[c]{@{}c@{}}b2ub-\\ p5-50\end{tabular}}} & Bottle          & \textbf{0.749} & \textbf{0.986} & \textbf{0.965} & \textbf{0.25}  & \textbf{0.745} & nan            & -              & -                         & -                         & -                         \\
\multicolumn{1}{c|}{}                                                                         & Can             & \textbf{0.492} & \textbf{0.838} & \textbf{0.501} & \textbf{0.528} & \textbf{0.486} & nan            & -              & -                         & -                         & -                         \\
\multicolumn{1}{c|}{}                                                                         & Chain           & 0.686          & \textbf{0.963} & \textbf{0.84}  & \textbf{0.734} & \textbf{0.554} & 0.718          & -              & -                         & -                         & -                         \\
\multicolumn{1}{c|}{}                                                                         & Drink-carton    & \textbf{0.641} & \textbf{0.978} & \textbf{0.832} & \textbf{0.663} & \textbf{0.619} & nan            & -              & -                         & -                         & -                         \\
\multicolumn{1}{c|}{}                                                                         & Hook            & 0.695          & 0.954          & 0.935          & nan            & 0.695          & nan            & -              & -                         & -                         & -                         \\
\multicolumn{1}{c|}{}                                                                         & Propeller       & 0.663          & 0.974          & 0.773          & \textbf{0.6}   & 0.667          & 0.7            & -              & -                         & -                         & -                         \\
\multicolumn{1}{c|}{}                                                                         & Shampoo-bottle  & 0.702          & 0.957          & 0.843          & nan            & \textbf{0.724} & 0.4            & -              & -                         & -                         & -                         \\
\multicolumn{1}{c|}{}                                                                         & Standing-bottle & 0.725          & 1              & 0.871          & nan            & 0.725          & nan            & -              & -                         & -                         & -                         \\
\multicolumn{1}{c|}{}                                                                         & Tire            & \textbf{0.809} & 0.97           & 0.937          & 0.3            & \textbf{0.822} & 0              & -              & -                         & -                         & -                         \\
\multicolumn{1}{c|}{}                                                                         & Valve           & 0.689          & 0.918          & 0.821          & 0              & 0.72           & 0              & -              & -                         & -                         & -                         \\
\multicolumn{1}{c|}{}                                                                         & Wall            & \textbf{0.717} & \textbf{0.936} & \textbf{0.843} & \textbf{0.371} & \textbf{0.719} & 0.794          & -              & -                         & -                         & -                         \\
\multicolumn{1}{c|}{}                                                                         & all             & \textbf{0.688} & \textbf{0.952} & 0.833          & \textbf{0.431} & \textbf{0.679} & 0.435          & \textbf{0.761} & 0.515                     & \textbf{0.753}            & 0.461                     \\ \hline
\multicolumn{1}{c|}{\multirow{12}{*}{\begin{tabular}[c]{@{}c@{}}nei-\\ g25\end{tabular}}}   & Bottle               & \textbf{0.742}       & 0.981                & \textbf{0.961}       & \textbf{0.3}         & \textbf{0.737}       & nan                  & -                    & -                    & -                    & -                    \\
\multicolumn{1}{c|}{}                                                                       & Can                  & 0.332                & 0.64                 & 0.298                & \textbf{0.349}       & 0.347                & nan                  & -                    & -                    & -                    & -                    \\
\multicolumn{1}{c|}{}                                                                       & Chain                & \textbf{0.689}       & 0.924                & 0.825                & \textbf{0.751}       & \textbf{0.486}       & 0.725                & -                    & -                    & -                    & -                    \\
\multicolumn{1}{c|}{}                                                                       & Drink-carton         & \textbf{0.568}       & \textbf{0.984}       & 0.558                & 0.578                & \textbf{0.568}       & nan                  & -                    & -                    & -                    & -                    \\
\multicolumn{1}{c|}{}                                                                       & Hook                 & 0.725                & 0.954                & 0.929                & nan                  & 0.725                & nan                  & -                    & -                    & -                    & -                    \\
\multicolumn{1}{c|}{}                                                                       & Propeller            & 0.674                & 0.984                & 0.805                & \textbf{0.6}         & 0.678                & 0.7                  & -                    & -                    & -                    & -                    \\
\multicolumn{1}{c|}{}                                                                       & Shampoo-bottle       & \textbf{0.725}       & \textbf{0.971}       & \textbf{0.971}       & nan                  & 0.724                & 0.8                  & -                    & -                    & -                    & -                    \\
\multicolumn{1}{c|}{}                                                                       & Standing-bottle      & 0.72                 & 1                    & 1                    & nan                  & 0.72                 & nan                  & -                    & -                    & -                    & -                    \\
\multicolumn{1}{c|}{}                                                                       & Tire                 & \textbf{0.806}       & 0.99                 & \textbf{0.969}       & 0.5                  & \textbf{0.811}       & \textbf{0.6}         & -                    & -                    & -                    & -                    \\
\multicolumn{1}{c|}{}                                                                       & Valve                & 0.713                & 0.925                & 0.881                & 0.25                 & 0.742                & 0                    & -                    & -                    & -                    & -                    \\
\multicolumn{1}{c|}{}                                                                       & Wall                 & \textbf{0.725}       & \textbf{0.928}       & 0.837                & \textbf{0.336}       & \textbf{0.714}       & \textbf{0.843}       & -                    & -                    & -                    & -                    \\
\multicolumn{1}{c|}{}                                                                       & all                  & 0.674                & \textbf{0.935}       & 0.821                & \textbf{0.458}       & 0.659                & 0.611                & 0.739                & 0.557                & 0.727                & \textbf{0.629}       \\ \hline
\multicolumn{1}{c|}{\multirow{12}{*}{\begin{tabular}[c]{@{}c@{}}nei-\\ g5-50\end{tabular}}} & Bottle               & 0.739                & 0.98                 & 0.921                & 0.167                & \textbf{0.737}       & nan                  & -                    & -                    & -                    & -                    \\
\multicolumn{1}{c|}{}                                                                       & Can                  & 0.32                 & 0.648                & 0.304                & 0.313                & 0.345                & nan                  & -                    & -                    & -                    & -                    \\
\multicolumn{1}{c|}{}                                                                       & Chain                & \textbf{0.696}       & \textbf{0.952}       & 0.797                & 0.213                & \textbf{0.511}       & 0.737                & \textbf{-}           & -                    & -                    & -                    \\
\multicolumn{1}{c|}{}                                                                       & Drink-carton         & 0.567                & \textbf{0.965}       & 0.685                & 0.569                & \textbf{0.569}       & nan                  & -                    & -                    & -                    & -                    \\
\multicolumn{1}{c|}{}                                                                       & Hook                 & 0.723                & 0.954                & 0.954                & nan                  & 0.723                & nan                  & -                    & -                    & -                    & -                    \\
\multicolumn{1}{c|}{}                                                                       & Propeller            & 0.692                & \textbf{0.99}        & 0.853                & 0.3                  & 0.705                & 0.6                  & -                    & -                    & -                    & -                    \\
\multicolumn{1}{c|}{}                                                                       & Shampoo-bottle       & \textbf{0.721}       & 0.958                & 0.894                & nan                  & \textbf{0.713}       & \textbf{0.9}         & -                    & -                    & -                    & -                    \\
\multicolumn{1}{c|}{}                                                                       & Standing-bottle      & 0.752                & 1                    & 1                    & nan                  & 0.752                & nan                  & -                    & -                    & -                    & -                    \\
\multicolumn{1}{c|}{}                                                                       & Tire                 & 0.799                & 0.98                 & 0.929                & 0.5                  & 0.81                 & 0                    & -                    & -                    & -                    & -                    \\
\multicolumn{1}{c|}{}                                                                       & Valve                & 0.68                 & 0.911                & 0.814                & 0                    & 0.713                & 0                    & -                    & -                    & -                    & -                    \\
\multicolumn{1}{c|}{}                                                                       & Wall                 & \textbf{0.714}       & \textbf{0.92}        & 0.828                & 0.353                & 0.695                & 0.839                & -                    & -                    & -                    & -                    \\
\multicolumn{1}{c|}{}                                                                       & all                  & 0.673                & \textbf{0.932}       & 0.816                & 0.302                & 0.661                & 0.513                & 0.734                & 0.396                & \textbf{0.732}       & 0.528                \\ \hline

\end{tabular}
}
\caption{The MarinDebris dataset is detected using the SSDMobileNetV2 algorithm (Part 2 of 3).  ``noisy" represents the original noise dataset, and dspnet, b2ub, and nei represent the target detection after denoising using the DSPNet, Blind2Unblind, and Neighbor2Neighbor single-frame denoising algorithms, respectively. Bold fonts indicate data that is higher than the corresponding ``noisy" dataset.}
\label{table12}
\end{table*}

\begin{table*}
\centering
\resizebox{\textwidth}{!}{
\begin{tabular}{cccccccccccc}
\hline
\multicolumn{1}{c|}{\multirow{12}{*}{\begin{tabular}[c]{@{}c@{}}nei-\\ p30\end{tabular}}}   & Bottle               & \textbf{0.751}       & 0.981                & \textbf{0.963}       & 0.167                & \textbf{0.749}       & nan                  & -                    & -                    & -                    & -                    \\
\multicolumn{1}{c|}{}                                                                       & Can                  & \textbf{0.354}       & 0.624                & \textbf{0.357}       & 0.344                & \textbf{0.379}       & nan                  & -                    & -                    & -                    & -                    \\
\multicolumn{1}{c|}{}                                                                       & Chain                & 0.684                & \textbf{0.95}        & 0.8                  & \textbf{0.251}       & 0.455                & 0.733                & -                    & -                    & -                    & -                    \\
\multicolumn{1}{c|}{}                                                                       & Drink-carton         & \textbf{0.578}       & \textbf{0.958}       & 0.65                 & 0.578                & \textbf{0.585}       & nan                  & -                    & -                    & -                    & -                    \\
\multicolumn{1}{c|}{}                                                                       & Hook                 & 0.701                & 0.944                & 0.914                & nan                  & 0.701                & nan                  & -                    & -                    & -                    & -                    \\
\multicolumn{1}{c|}{}                                                                       & Propeller            & 0.632                & 0.964                & 0.77                 & \textbf{0.6}         & 0.638                & 0.6                  & -                    & -                    & -                    & -                    \\
\multicolumn{1}{c|}{}                                                                       & Shampoo-bottle       & \textbf{0.735}       & 0.964                & 0.964                & nan                  & \textbf{0.729}       & \textbf{0.9}         & -                    & -                    & -                    & -                    \\
\multicolumn{1}{c|}{}                                                                       & Standing-bottle      & 0.736                & 1                    & 1                    & nan                  & 0.736                & nan                  & -                    & -                    & -                    & -                    \\
\multicolumn{1}{c|}{}                                                                       & Tire                 & \textbf{0.808}       & 0.99                 & \textbf{0.956}       & 0.6                  & \textbf{0.813}       & \textbf{0.5}         & -                    & -                    & -                    & -                    \\
\multicolumn{1}{c|}{}                                                                       & Valve                & 0.72                 & \textbf{0.944}       & 0.901                & 0.2                  & 0.751                & 0                    & -                    & -                    & -                    & -                    \\
\multicolumn{1}{c|}{}                                                                       & Wall                 & 0.703                & \textbf{0.925}       & 0.817                & 0.284                & 0.691                & 0.84                 & -                    & -                    & -                    & -                    \\
\multicolumn{1}{c|}{}                                                                       & all                  & 0.673                & \textbf{0.931}       & 0.826                & 0.378                & 0.657                & \textbf{0.595}       & 0.738                & 0.465                & \textbf{0.731}       & \textbf{0.614}       \\ \hline
\multicolumn{1}{c|}{\multirow{12}{*}{\begin{tabular}[c]{@{}c@{}}nei-\\ p5-50\end{tabular}}} & Bottle               & \textbf{0.755}       & \textbf{0.982}       & \textbf{0.961}       & \textbf{0.5}         & \textbf{0.752}       & nan                  & -                    & -                    & -                    & -                    \\
\multicolumn{1}{c|}{}                                                                       & Can                  & \textbf{0.343}       & 0.644                & 0.305                & 0.333                & \textbf{0.365}       & nan                  & -                    & -                    & -                    & -                    \\
\multicolumn{1}{c|}{}                                                                       & Chain                & \textbf{0.688}       & \textbf{0.968}       & 0.804                & \textbf{0.501}       & 0.431                & 0.734                & -                    & -                    & -                    & -                    \\
\multicolumn{1}{c|}{}                                                                       & Drink-carton         & \textbf{0.577}       & \textbf{0.953}       & 0.648                & 0.574                & \textbf{0.594}       & nan                  & -                    & -                    & -                    & -                    \\
\multicolumn{1}{c|}{}                                                                       & Hook                 & \textbf{0.736}       & 0.948                & 0.929                & nan                  & \textbf{0.736}       & nan                  & -                    & -                    & -                    & -                    \\
\multicolumn{1}{c|}{}                                                                       & Propeller            & 0.681                & 0.983                & 0.834                & \textbf{0.6}         & 0.686                & 0.6                  & -                    & -                    & -                    & -                    \\
\multicolumn{1}{c|}{}                                                                       & Shampoo-bottle       & \textbf{0.736}       & 0.958                & 0.958                & nan                  & \textbf{0.734}       & 0.8                  & -                    & -                    & -                    & -                    \\
\multicolumn{1}{c|}{}                                                                       & Standing-bottle      & \textbf{0.803}       & 1                    & 1                    & nan                  & \textbf{0.803}       & nan                  & -                    & -                    & -                    & -                    \\
\multicolumn{1}{c|}{}                                                                       & Tire                 & 0.803                & 0.98                 & 0.938                & 0.5                  & \textbf{0.813}       & 0                    & -                    & -                    & -                    & -                    \\
\multicolumn{1}{c|}{}                                                                       & Valve                & 0.681                & 0.914                & 0.887                & 0                    & 0.712                & 0                    & -                    & -                    & -                    & -                    \\
\multicolumn{1}{c|}{}                                                                       & Wall                 & \textbf{0.717}       & \textbf{0.93}        & \textbf{0.848}       & 0.304                & \textbf{0.718}       & 0.83                 & -                    & -                    & -                    & -                    \\
\multicolumn{1}{c|}{}                                                                       & all                  & \textbf{0.684}       & \textbf{0.933}       & 0.828                & 0.414                & \textbf{0.668}       & 0.494                & \textbf{0.745}       & 0.437                & \textbf{0.74}        & 0.512                \\ \hline
\multicolumn{1}{l}{}                                                                        & \multicolumn{1}{l}{} & \multicolumn{1}{l}{} & \multicolumn{1}{l}{} & \multicolumn{1}{l}{} & \multicolumn{1}{l}{} & \multicolumn{1}{l}{} & \multicolumn{1}{l}{} & \multicolumn{1}{l}{} & \multicolumn{1}{l}{} & \multicolumn{1}{l}{} & \multicolumn{1}{l}{} \\
\multicolumn{1}{l}{}                                                                        & \multicolumn{1}{l}{} & \multicolumn{1}{l}{} & \multicolumn{1}{l}{} & \multicolumn{1}{l}{} & \multicolumn{1}{l}{} & \multicolumn{1}{l}{} & \multicolumn{1}{l}{} & \multicolumn{1}{l}{} & \multicolumn{1}{l}{} & \multicolumn{1}{l}{} & \multicolumn{1}{l}{}
\end{tabular}
}
\caption{The MarinDebris dataset is detected using the SSDMobileNetV2 algorithm (Part 3 of 3).  ``noisy" represents the original noise dataset, and dspnet, b2ub, and nei represent the target detection after denoising using the DSPNet, Blind2Unblind, and Neighbor2Neighbor single-frame denoising algorithms, respectively. Bold fonts indicate data that is higher than the corresponding ``noisy" dataset.}
\label{table13}
\end{table*}

\begin{table*}
\centering
\resizebox{\textwidth}{!}{
\begin{tabular}{c|ccccccccccc}
\hline
\multirow{2}{*}{Denoising}                                              & \multicolumn{11}{c}{Faster R-CNN}                                                                                                                                                                                          \\ \cline{2-12} 
                                                                        & class           & mAP            & mAP@0.5        & mAP@0.75        & mAP\_s         & mAP\_m         & mAP\_l         & AR             & \multicolumn{1}{l}{AR\_s} & \multicolumn{1}{l}{AR\_m} & \multicolumn{1}{l}{AR\_l} \\ \hline
\multirow{12}{*}{noisy}                                                 & Bottle          & 0.766          & 0.985          & 0.974          & 0.7            & 0.761          & nan            & -              & -                         & -                         & -                         \\
                                                                        & Can             & 0.493          & 0.821          & 0.509          & 0.514          & 0.489          & nan            & -              & -                         & -                         & -                         \\
                                                                        & Chain           & 0.676          & 0.963          & 0.813          & 0.701          & 0.524          & 0.704          & -              & -                         & -                         & -                         \\
                                                                        & Drink-carton    & 0.65           & 0.995          & 0.827          & 0.643          & 0.664          & nan            & -              & -                         & -                         & -                         \\
                                                                        & Hook            & 0.655          & 0.927          & 0.84           & nan            & 0.655          & nan            & -              & -                         & -                         & -                         \\
                                                                        & Propeller       & 0.652          & 0.945          & 0.745          & 0.5            & 0.655          & 0.7            & -              & -                         & -                         & -                         \\
                                                                        & Shampoo-bottle  & 0.724          & 0.964          & 0.894          & nan            & 0.729          & 0.8            & -              & -                         & -                         & -                         \\
                                                                        & Standing-bottle & 0.78           & 1              & 1              & nan            & 0.78           & nan            & -              & -                         & -                         & -                         \\
                                                                        & Tire            & 0.812          & 0.989          & 0.945          & 0.1            & 0.824          & 0              & -              & -                         & -                         & -                         \\
                                                                        & Valve           & 0.692          & 0.917          & 0.86           & 0              & 0.722          & 0              & -              & -                         & -                         & -                         \\
                                                                        & Wall            & 0.716          & 0.935          & 0.85           & 0.413          & 0.723          & 0.778          & -              & -                         & -                         & -                         \\
                                                                        & all             & 0.692          & 0.949          & 0.842          & 0.446          & 0.684          & 0.497          & 0.763          & 0.502                     & 0.754                     & 0.525                     \\ \hline
\multirow{12}{*}{dspnet}                                                & Bottle          & 0.746          & 0.985          & 0.933          & 0.7            & 0.742          & nan            & -              & -                         & -                         & -                         \\
                                                                        & Can             & 0.428          & 0.808          & 0.447          & 0.417          & 0.444          & nan            & -              & -                         & -                         & -                         \\
                                                                        & Chain           & 0.667          & \textbf{0.964} & \textbf{0.833} & 0.635          & 0.462          & \textbf{0.709} & -              & -                         & -                         & -                         \\
                                                                        & Drink-carton    & 0.632          & 0.993          & 0.782          & 0.626          & 0.658          & nan            & -              & -                         & -                         & -                         \\
                                                                        & Hook            & \textbf{0.675} & \textbf{0.934} & 0.818          & nan            & \textbf{0.675} & nan            & -              & -                         & -                         & -                         \\
                                                                        & Propeller       & 0.595          & \textbf{0.966} & 0.678          & 0.3            & 0.603          & 0.5            & -              & -                         & -                         & -                         \\
                                                                        & Shampoo-bottle  & 0.715          & \textbf{0.969} & \textbf{0.909} & nan            & 0.718          & 0.8            & -              & -                         & -                         & -                         \\
                                                                        & Standing-bottle & 0.73           & 1              & 0.901          & nan            & 0.73           & nan            & -              & -                         & -                         & -                         \\
                                                                        & Tire            & 0.792          & \textbf{0.999} & 0.94           & 0.1            & 0.803          & \textbf{0.3}   & -              & -                         & -                         & -                         \\
                                                                        & Valve           & 0.676          & 0.917          & 0.806          & 0              & 0.705          & 0              & -              & -                         & -                         & -                         \\
                                                                        & Wall            & 0.705          & \textbf{0.923} & 0.827          & 0.371          & 0.711          & 0.772          & -              & -                         & -                         & -                         \\
                                                                        & all             & 0.669          & \textbf{0.951} & 0.807          & 0.394          & 0.659          & \textbf{0.513} & 0.74           & 0.487                     & 0.731                     & \textbf{0.536}            \\ \hline
\multirow{12}{*}{\begin{tabular}[c]{@{}c@{}}b2ub-\\ g25\end{tabular}}   & Bottle          & 0.759          & \textbf{0.986} & 0.963          & 0.3            & 0.754          & nan            & -              & -                         & -                         & -                         \\
                                                                        & Can             & \textbf{0.499} & \textbf{0.827} & 0.493          & \textbf{0.564} & 0.478          & nan            & -              & -                         & -                         & -                         \\
                                                                        & Chain           & \textbf{0.678} & \textbf{0.969} & 0.786          & \textbf{0.75}  & 0.499          & \textbf{0.715} & -              & -                         & -                         & -                         \\
                                                                        & Drink-carton    & \textbf{0.66}  & 0.989          & 0.754          & \textbf{0.651} & \textbf{0.688} & nan            & -              & -                         & -                         & -                         \\
                                                                        & Hook            & \textbf{0.715} & \textbf{0.939} & \textbf{0.914} & nan            & \textbf{0.715} & nan            & -              & -                         & -                         & -                         \\
                                                                        & Propeller       & \textbf{0.659} & \textbf{0.99}  & \textbf{0.822} & \textbf{0.6}   & \textbf{0.66}  & 0.7            & -              & -                         & -                         & -                         \\
                                                                        & Shampoo-bottle  & 0.721          & \textbf{0.98}  & \textbf{0.919} & nan            & \textbf{0.73}  & 0.6            & -              & -                         & -                         & -                         \\
                                                                        & Standing-bottle & 0.712          & 1              & 0.811          & nan            & 0.712          & nan            & -              & -                         & -                         & -                         \\
                                                                        & Tire            & \textbf{0.813} & 0.96           & 0.942          & 0              & \textbf{0.829} & 0              & -              & -                         & -                         & -                         \\
                                                                        & Valve           & 0.684          & \textbf{0.918} & \textbf{0.879} & 0              & 0.713          & 0              & -              & -                         & -                         & -                         \\
                                                                        & Wall            & \textbf{0.721} & \textbf{0.94}  & 0.829          & 0.361          & \textbf{0.735} & \textbf{0.788} & -              & -                         & -                         & -                         \\
                                                                        & all             & \textbf{0.693} & \textbf{0.954} & 0.828          & 0.403          & 0.683          & 0.467          & 0.762          & 0.48                      & 0.754                     & 0.492                     \\ \hline
\multirow{12}{*}{\begin{tabular}[c]{@{}c@{}}b2ub-\\ g5-50\end{tabular}} & Bottle          & 0.765          & \textbf{0.986} & 0.975          & 0.233          & 0.76           & nan            & -              & -                         & -                         & -                         \\
                                                                        & Can             & \textbf{0.512} & \textbf{0.844} & 0.508          & \textbf{0.552} & \textbf{0.503} & nan            & -              & -                         & -                         & -                         \\
                                                                        & Chain           & \textbf{0.689} & \textbf{0.965} & \textbf{0.836} & \textbf{0.75}  & \textbf{0.562} & \textbf{0.716} & -              & -                         & -                         & -                         \\
                                                                        & Drink-carton    & 0.643          & 0.976          & 0.808          & \textbf{0.662} & 0.623          & nan            & -              & -                         & -                         & -                         \\
                                                                        & Hook            & \textbf{0.709} & \textbf{0.948} & \textbf{0.897} & nan            & \textbf{0.709} & nan            & -              & -                         & -                         & -                         \\
                                                                        & Propeller       & \textbf{0.676} & \textbf{0.975} & \textbf{0.874} & \textbf{0.6}   & \textbf{0.681} & 0.7            & -              & -                         & -                         & -                         \\
                                                                        & Shampoo-bottle  & 0.704          & \textbf{0.978} & 0.882          & nan            & 0.72           & 0.4            & -              & -                         & -                         & -                         \\
                                                                        & Standing-bottle & \textbf{0.818} & 1              & 1              & nan            & \textbf{0.818} & nan            & -              & -                         & -                         & -                         \\
                                                                        & Tire            & \textbf{0.824} & 0.989          & 0.938          & \textbf{0.3}   & \textbf{0.836} & 0              & -              & -                         & -                         & -                         \\
                                                                        & Valve           & \textbf{0.696} & \textbf{0.92}  & \textbf{0.865} & 0              & \textbf{0.726} & 0              & -              & -                         & -                         & -                         \\
                                                                        & Wall            & 0.716          & 0.914          & 0.847          & 0.324          & \textbf{0.73}  & \textbf{0.784} & -              & -                         & -                         & -                         \\
                                                                        & all             & \textbf{0.705} & \textbf{0.954} & \textbf{0.857} & 0.428          & \textbf{0.697} & 0.433          & \textbf{0.768} & \textbf{0.531}            & \textbf{0.758}            & 0.494                     \\ \hline
\end{tabular}
}
\caption{The MarinDebris dataset is detected using the Faster R-CNN algorithm (Part 1 of 3). ``noisy" represents the original noise dataset, and dspnet, b2ub, and nei represent the target detection after denoising using the DSPNet, Blind2Unblind, and Neighbor2Neighbor single-frame denoising algorithms, respectively. Bold fonts indicate data that is higher than the corresponding ``noisy" dataset.}

\label{table14}
\end{table*}

\begin{table*}
\centering
\resizebox{\textwidth}{!}{
\begin{tabular}{cccccccccccc}
\hline
\multicolumn{1}{c|}{\multirow{12}{*}{\begin{tabular}[c]{@{}c@{}}b2ub-\\ p30\end{tabular}}}   & Bottle          & 0.766          & \textbf{0.986} & 0.965          & 0.5            & \textbf{0.762} & nan            & -              & -                         & -                         & -                         \\
\multicolumn{1}{c|}{}                                                                        & Can             & 0.489          & \textbf{0.829} & \textbf{0.517} & \textbf{0.534} & 0.481          & nan            & -              & -                         & -                         & -                         \\
\multicolumn{1}{c|}{}                                                                        & Chain           & \textbf{0.686} & \textbf{0.97}  & \textbf{0.846} & \textbf{0.75}  & \textbf{0.526} & \textbf{0.718} & -              & -                         & -                         & -                         \\
\multicolumn{1}{c|}{}                                                                        & Drink-carton    & 0.639          & 0.977          & 0.808          & \textbf{0.657} & 0.636          & nan            & -              & -                         & -                         & -                         \\
\multicolumn{1}{c|}{}                                                                        & Hook            & \textbf{0.706} & \textbf{0.938} & \textbf{0.88}  & nan            & \textbf{0.706} & nan            & -              & -                         & -                         & -                         \\
\multicolumn{1}{c|}{}                                                                        & Propeller       & \textbf{0.669} & \textbf{0.977} & \textbf{0.78}  & \textbf{0.6}   & \textbf{0.671} & 0.7            & -              & -                         & -                         & -                         \\
\multicolumn{1}{c|}{}                                                                        & Shampoo-bottle  & 0.711          & 0.957          & \textbf{0.896} & nan            & 0.713          & 0.8            & -              & -                         & -                         & -                         \\
\multicolumn{1}{c|}{}                                                                        & Standing-bottle & 0.711          & 1              & 0.871          & nan            & 0.771          & nan            & -              & -                         & -                         & -                         \\
\multicolumn{1}{c|}{}                                                                        & Tire            & 0.807          & 0.979          & 0.939          & \textbf{0.2}   & 0.818          & 0              & -              & -                         & -                         & -                         \\
\multicolumn{1}{c|}{}                                                                        & Valve           & 0.691          & \textbf{0.92}  & \textbf{0.863} & 0              & 0.721          & 0              & -              & -                         & -                         & -                         \\
\multicolumn{1}{c|}{}                                                                        & Wall            & \textbf{0.717} & 0.934          & 0.821          & 0.36           & \textbf{0.725} & \textbf{0.78}  & -              & -                         & -                         & -                         \\
\multicolumn{1}{c|}{}                                                                        & all             & \textbf{0.696} & \textbf{0.952} & 0.835          & \textbf{0.45}  & \textbf{0.685} & 0.5            & \textbf{0.764} & \textbf{0.52}             & 0.752                     & \textbf{0.526}            \\ \hline
\multicolumn{1}{c|}{\multirow{12}{*}{\begin{tabular}[c]{@{}c@{}}b2ub-\\ p5-50\end{tabular}}} & Bottle          & 0.749          & \textbf{0.986} & 0.965          & 0.25           & 0.745          & nan            & -              & -                         & -                         & -                         \\
\multicolumn{1}{c|}{}                                                                         & Can             & 0.492          & \textbf{0.838} & 0.501          & \textbf{0.528} & 0.486          & nan            & -              & -                         & -                         & -                         \\
\multicolumn{1}{c|}{}                                                                         & Chain           & \textbf{0.686} & 0.963          & \textbf{0.84}  & \textbf{0.734} & \textbf{0.554} & \textbf{0.718} & -              & -                         & -                         & -                         \\
\multicolumn{1}{c|}{}                                                                          & Drink-carton    & 0.641          & 0.978          & \textbf{0.832} & \textbf{0.663} & 0.619          & nan            & -              & -                         & -                         & -                         \\
\multicolumn{1}{c|}{}                                                                          & Hook            & \textbf{0.695} & \textbf{0.954} & \textbf{0.935} & nan            & \textbf{0.695} & nan            & -              & -                         & -                         & -                         \\
\multicolumn{1}{c|}{}                                                                          & Propeller       & \textbf{0.663} & \textbf{0.974} & \textbf{0.773} & \textbf{0.6}   & \textbf{0.667} & 0.7            & -              & -                         & -                         & -                         \\
 \multicolumn{1}{c|}{}                                                                         & Shampoo-bottle  & 0.702          & 0.957          & 0.843          & nan            & 0.724          & 0.4            & -              & -                         & -                         & -                         \\
\multicolumn{1}{c|}{}                                                                          & Standing-bottle & 0.725          & 1              & 0.871          & nan            & 0.725          & nan            & -              & -                         & -                         & -                         \\
\multicolumn{1}{c|}{}                                                                          & Tire            & 0.809          & 0.97           & 0.937          & \textbf{0.3}   & 0.822          & 0              & -              & -                         & -                         & -                         \\
\multicolumn{1}{c|}{}                                                                          & Valve           & 0.689          & \textbf{0.918} & 0.821          & 0              & 0.72           & 0              & -              & -                         & -                         & -                         \\
\multicolumn{1}{c|}{}                                                                          & Wall            & \textbf{0.717} & \textbf{0.936} & 0.843          & 0.371          & 0.719          & \textbf{0.794} & -              & -                         & -                         & -                         \\
\multicolumn{1}{c|}{}                                                                          & all             & 0.688          & \textbf{0.952} & 0.833          & 0.431          & 0.679          & 0.435          & 0.761          & \textbf{0.515}            & 0.753                     & 0.461                     \\ \hline
\multicolumn{1}{c|}{\multirow{12}{*}{\begin{tabular}[c]{@{}c@{}}nei-\\ g25\end{tabular}}}   & Bottle               & 0.764                & \textbf{0.988}       & \textbf{0.978}       & 0.7                  & 0.757                & nan                  & -                    & -                    & -                    & -                    \\
\multicolumn{1}{c|}{}                                                                       & Can                  & 0.492                & \textbf{0.847}       & 0.499                & \textbf{0.524}       & \textbf{0.498}       & nan                  & -                    & -                    & -                    & -                    \\
\multicolumn{1}{c|}{}                                                                       & Chain                & 0.672                & \textbf{0.965}       & 0.812                & 0.617                & 0.522                & 0.707                & -                    & -                    & -                    & -                    \\
\multicolumn{1}{c|}{}                                                                       & Drink-carton         & 0.637                & \textbf{0.996}       & \textbf{0.84}        & \textbf{0.653}       & 0.623                & nan                  & -                    & -                    & -                    & -                    \\
\multicolumn{1}{c|}{}                                                                       & Hook                 & \textbf{0.676}       & \textbf{0.931}       & \textbf{0.905}       & nan                  & \textbf{0.676}       & nan                  & -                    & -                    & -                    & -                    \\
\multicolumn{1}{c|}{}                                                                       & Propeller            & \textbf{0.654}       & \textbf{0.979}       & 0.822                & 0.25                 & \textbf{0.662}       & 0.7                  & -                    & -                    & -                    & -                    \\
\multicolumn{1}{c|}{}                                                                       & Shampoo-bottle       & 0.697                & \textbf{0.967}       & 0.859                & nan                  & 0.721                & 0.267                & -                    & -                    & -                    & -                    \\
\multicolumn{1}{c|}{}                                                                       & Standing-bottle      & 0.777                & 1                    & 1                    & nan                  & 0.777                & nan                  & -                    & -                    & -                    & -                    \\
\multicolumn{1}{c|}{}                                                                       & Tire                 & 0.802                & 0.979                & \textbf{0.949}       & 0.1                  & 0.816                & 0                    & -                    & -                    & -                    & -                    \\
\multicolumn{1}{c|}{}                                                                       & Valve                & 0.674                & \textbf{0.92}        & 0.824                & 0                    & 0.703                & 0                    & -                    & -                    & -                    & -                    \\
\multicolumn{1}{c|}{}                                                                       & Wall                 & 0.711                & 0.926                & 0.834                & 0.399                & 0.719                & 0.773                & -                    & -                    & -                    & -                    \\
\multicolumn{1}{c|}{}                                                                       & all                  & 0.687                & \textbf{0.954}       & \textbf{0.847}       & 0.405                & 0.68                 & 0.408                & 0.757                & \textbf{0.506}       & 0.746                & 0.458                \\ \hline
\multicolumn{1}{c|}{\multirow{12}{*}{\begin{tabular}[c]{@{}c@{}}nei-\\ g5-50\end{tabular}}} & Bottle               & 0.762                & 0.984                & 0.964                & 0.25                 & 0.758                & nan                  & -                    & -                    & -                    & -                    \\
\multicolumn{1}{c|}{}                                                                       & Can                  & 0.471                & \textbf{0.83}        & 0.471                & 0.491                & 0.469                & nan                  & -                    & -                    & -                    & -                    \\
\multicolumn{1}{c|}{}                                                                       & Chain                & \textbf{0.685}       & 0.961                & \textbf{0.882}       & \textbf{0.8}         & \textbf{0.529}       & \textbf{0.714}       & \textbf{-}           & -                    & -                    & -                    \\
\multicolumn{1}{c|}{}                                                                       & Drink-carton         & 0.628                & 0.975                & 0.821                & 0.639                & 0.626                & nan                  & -                    & -                    & -                    & -                    \\
\multicolumn{1}{c|}{}                                                                       & Hook                 & \textbf{0.718}       & \textbf{0.954}       & \textbf{0.906}       & nan                  & \textbf{0.718}       & nan                  & -                    & -                    & -                    & -                    \\
\multicolumn{1}{c|}{}                                                                       & Propeller            & \textbf{0.692}       & \textbf{0.979}       & \textbf{0.835}       & \textbf{0.6}         & \textbf{0.695}       & \textbf{0.8}         & -                    & -                    & -                    & -                    \\
\multicolumn{1}{c|}{}                                                                       & Shampoo-bottle       & 0.714                & 0.961                & 0.892                & nan                  & 0.716                & 0.7                  & -                    & -                    & -                    & -                    \\
\multicolumn{1}{c|}{}                                                                       & Standing-bottle      & \textbf{0.794}       & 1                    & 1                    & nan                  & \textbf{0.794}       & nan                  & -                    & -                    & -                    & -                    \\
\multicolumn{1}{c|}{}                                                                       & Tire                 & 0.806                & 0.969                & \textbf{0.958}       & 0                    & 0.821                & 0                    & -                    & -                    & -                    & -                    \\
\multicolumn{1}{c|}{}                                                                       & Valve                & 0.687                & \textbf{0.935}       & 0.838                & 0                    & 0.717                & 0                    & -                    & -                    & -                    & -                    \\
\multicolumn{1}{c|}{}                                                                       & Wall                 & \textbf{0.728}       & 0.914                & 0.824                & 0.342                & \textbf{0.743}       & \textbf{0.795}       & -                    & -                    & -                    & -                    \\
\multicolumn{1}{c|}{}                                                                       & all                  & \textbf{0.699}       & \textbf{0.951}       & \textbf{0.854}       & 0.39                 & \textbf{0.689}       & \textbf{0.501}       & \textbf{0.77}        & 0.472                & \textbf{0.758}       & 0.525                \\ \hline

\end{tabular}
}
\caption{The MarinDebris dataset is detected using the Faster R-CNN algorithm (Part 2 of 3). ``noisy" represents the original noise dataset, and dspnet, b2ub, and nei represent the target detection after denoising using the DSPNet, Blind2Unblind, and Neighbor2Neighbor single-frame denoising algorithms, respectively. Bold fonts indicate data that is higher than the corresponding ``noisy" dataset.}

\label{table15}
\end{table*}

\begin{table*}
\centering
\resizebox{\textwidth}{!}{
\begin{tabular}{cccccccccccc}
\hline
\multicolumn{1}{c|}{\multirow{12}{*}{\begin{tabular}[c]{@{}c@{}}nei-\\ p30\end{tabular}}}   & Bottle               & 0.758                & 0.985                & 0.974                & 0.3                  & 0.754                & nan                  & -                    & -                    & -                    & -                    \\
\multicolumn{1}{c|}{}                                                                       & Can                  & 0.489                & \textbf{0.843}       & 0.473                & \textbf{0.541}       & 0.481                & nan                  & -                    & -                    & -                    & -                    \\
\multicolumn{1}{c|}{}                                                                       & Chain                & \textbf{0.693}       & \textbf{0.967}       & \textbf{0.855}       & \textbf{0.75}        & 0.522                & \textbf{0.723}       & -                    & -                    & -                    & -                    \\
\multicolumn{1}{c|}{}                                                                       & Drink-carton         & 0.638                & 0.995                & 0.813                & \textbf{0.645}       & 0.639                & nan                  & -                    & -                    & -                    & -                    \\
\multicolumn{1}{c|}{}                                                                       & Hook                 & \textbf{0.693}       & \textbf{0.934}       & \textbf{0.934}       & nan                  & \textbf{0.693}       & nan                  & -                    & -                    & -                    & -                    \\
\multicolumn{1}{c|}{}                                                                       & Propeller            & \textbf{0.654}       & \textbf{0.973}       & \textbf{0.799}       & 0.5                  & \textbf{0.66}        & 0.6                  & -                    & -                    & -                    & -                    \\
\multicolumn{1}{c|}{}                                                                       & Shampoo-bottle       & 0.674                & \textbf{0.983}       & 0.87                 & nan                  & 0.686                & 0.5                  & -                    & -                    & -                    & -                    \\
\multicolumn{1}{c|}{}                                                                       & Standing-bottle      & 0.772                & 1                    & 0.822                & nan                  & 0.772                & nan                  & -                    & -                    & -                    & -                    \\
\multicolumn{1}{c|}{}                                                                       & Tire                 & 0.804                & 0.97                 & \textbf{0.948}       & \textbf{0.2}         & 0.817                & 0                    & -                    & -                    & -                    & -                    \\
\multicolumn{1}{c|}{}                                                                       & Valve                & 0.677                & \textbf{0.92}        & \textbf{0.881}       & 0                    & 0.706                & 0                    & -                    & -                    & -                    & -                    \\
\multicolumn{1}{c|}{}                                                                       & Wall                 & 0.705                & 0.915                & 0.813                & 0.332                & 0.717                & 0.773                & -                    & -                    & -                    & -                    \\
\multicolumn{1}{c|}{}                                                                       & all                  & 0.687                & \textbf{0.953}       & 0.835                & 0.409                & 0.677                & 0.433                & 0.758                & 0.495                & 0.751                & 0.46                 \\ \hline
\multicolumn{1}{c|}{\multirow{12}{*}{\begin{tabular}[c]{@{}c@{}}nei-\\ p5-50\end{tabular}}} & Bottle               & 0.76                 & \textbf{0.987}       & 0.965                & 0.4                  & 0.757                & nan                  & -                    & -                    & -                    & -                    \\
\multicolumn{1}{c|}{}                                                                       & Can                  & 0.474                & \textbf{0.831}       & 0.481                & 0.514                & 0.462                & nan                  & -                    & -                    & -                    & -                    \\
\multicolumn{1}{c|}{}                                                                       & Chain                & \textbf{0.689}       & \textbf{0.964}       & \textbf{0.878}       & 0.7                  & \textbf{0.533}       & \textbf{0.719}       & -                    & -                    & -                    & -                    \\
\multicolumn{1}{c|}{}                                                                       & Drink-carton         & 0.63                 & 0.977                & 0.789                & \textbf{0.662}       & 0.59                 & nan                  & -                    & -                    & -                    & -                    \\
\multicolumn{1}{c|}{}                                                                       & Hook                 & \textbf{0.682}       & \textbf{0.945}       & \textbf{0.897}       & nan                  & \textbf{0.682}       & nan                  & -                    & -                    & -                    & -                    \\
\multicolumn{1}{c|}{}                                                                       & Propeller            & \textbf{0.658}       & \textbf{0.972}       & \textbf{0.845}       & \textbf{0.6}         & \textbf{0.658}       & 0.7                  & -                    & -                    & -                    & -                    \\
\multicolumn{1}{c|}{}                                                                       & Shampoo-bottle       & 0.712                & 0.964                & \textbf{0.964}       & nan                  & 0.726                & 0.6                  & -                    & -                    & -                    & -                    \\
\multicolumn{1}{c|}{}                                                                       & Standing-bottle      & 0.767                & 1                    & 1                    & nan                  & 0.767                & nan                  & -                    & -                    & -                    & -                    \\
\multicolumn{1}{c|}{}                                                                       & Tire                 & 0.811                & 0.977                & 0.925                & 0                    & \textbf{0.826}       & 0                    & -                    & -                    & -                    & -                    \\
\multicolumn{1}{c|}{}                                                                       & Valve                & 0.682                & 0.915                & 0.836                & 0                    & 0.712                & 0                    & -                    & -                    & -                    & -                    \\
\multicolumn{1}{c|}{}                                                                       & Wall                 & \textbf{0.725}       & 0.921                & 0.844                & 0.37                 & \textbf{0.738}       & \textbf{0.795}       & -                    & -                    & -                    & -                    \\
\multicolumn{1}{c|}{}                                                                       & all                  & 0.69                 & \textbf{0.95}        & \textbf{0.857}       & 0.406                & 0.678                & 0.469                & 0.759                & 0.446                & 0.748                & 0.492                \\ \hline
\multicolumn{1}{l}{}                                                                        & \multicolumn{1}{l}{} & \multicolumn{1}{l}{} & \multicolumn{1}{l}{} & \multicolumn{1}{l}{} & \multicolumn{1}{l}{} & \multicolumn{1}{l}{} & \multicolumn{1}{l}{} & \multicolumn{1}{l}{} & \multicolumn{1}{l}{} & \multicolumn{1}{l}{} & \multicolumn{1}{l}{} \\
\multicolumn{1}{l}{}                                                                        & \multicolumn{1}{l}{} & \multicolumn{1}{l}{} & \multicolumn{1}{l}{} & \multicolumn{1}{l}{} & \multicolumn{1}{l}{} & \multicolumn{1}{l}{} & \multicolumn{1}{l}{} & \multicolumn{1}{l}{} & \multicolumn{1}{l}{} & \multicolumn{1}{l}{} & \multicolumn{1}{l}{}
\end{tabular}
}
\caption{The MarinDebris dataset is detected using the Faster R-CNN algorithm (Part 3 of 3). ``noisy" represents the original noise dataset, and dspnet, b2ub, and nei represent the target detection after denoising using the DSPNet, Blind2Unblind, and Neighbor2Neighbor single-frame denoising algorithms, respectively. Bold fonts indicate data that is higher than the corresponding ``noisy" dataset.}

\label{table16}
\end{table*}

\begin{table*}
\centering
\resizebox{\textwidth}{!}{
\begin{tabular}{c|ccccccccccc}
\hline
\multirow{2}{*}{Denoising}                                              & \multicolumn{11}{c}{SSD300}                                                                                                                                                                                                \\ \cline{2-12} 
                                                                        & class           & mAP            & mAP@0.5        & mAP@0.75        & mAP\_s         & mAP\_m         & mAP\_l         & AR             & \multicolumn{1}{l}{AR\_s} & \multicolumn{1}{l}{AR\_m} & \multicolumn{1}{l}{AR\_l} \\ \hline
\multirow{12}{*}{noisy}                                                 & Bottle          & 0.753          & 0.986          & 0.975          & 0.7            & 0.748          & nan            & -              & -                         & -                         & -                         \\
                                                                        & Can             & 0.452          & 0.779          & 0.452          & 0.538          & 0.436          & nan            & -              & -                         & -                         & -                         \\
                                                                        & Chain           & 0.683          & 0.968          & 0.844          & 0.825          & 0.506          & 0.708          & -              & -                         & -                         & -                         \\
                                                                        & Drink-carton    & 0.594          & 0.972          & 0.753          & 0.59           & 0.624          & nan            & -              & -                         & -                         & -                         \\
                                                                        & Hook            & 0.719          & 0.952          & 0.952          & nan            & 0.719          & nan            & -              & -                         & -                         & -                         \\
                                                                        & Propeller       & 0.677          & 0.993          & 0.835          & 0.4            & 0.688          & 0.6            & -              & -                         & -                         & -                         \\
                                                                        & Shampoo-bottle  & 0.709          & 0.95           & 0.902          & nan            & 0.738          & 0.3            & -              & -                         & -                         & -                         \\
                                                                        & Standing-bottle & 0.726          & 1              & 0.822          & nan            & 0.726          & nan            & -              & -                         & -                         & -                         \\
                                                                        & Tire            & 0.8            & 0.998          & 0.935          & 0.6            & 0.805          & 0.5            & -              & -                         & -                         & -                         \\
                                                                        & Valve           & 0.715          & 0.949          & 0.848          & 0.3            & 0.741          & 0              & -              & -                         & -                         & -                         \\
                                                                        & Wall            & 0.71           & 0.926          & 0.831          & 0.279          & 0.715          & 0.818          & -              & -                         & -                         & -                         \\
                                                                        & all             & 0.685          & 0.952          & 0.832          & 0.529          & 0.677          & 0.488          & 0.749          & 0.573                     & 0.744                     & 0.505                     \\ \hline
\multirow{12}{*}{dspnet}                                                & Bottle          & 0.737          & 0.982          & 0.926          & 0.233          & 0.734          & nan            & -              & -                         & -                         & -                         \\
                                                                        & Can             & 0.436          & \textbf{0.845} & \textbf{0.491} & 0.487          & 0.433          & nan            & -              & -                         & -                         & -                         \\
                                                                        & Chain           & 0.677          & \textbf{0.972} & 0.792          & 0.701          & 0.483          & \textbf{0.714} & -              & -                         & -                         & -                         \\
                                                                        & Drink-carton    & 0.562          & 0.942          & 0.634          & 0.55           & 0.585          & nan            & -              & -                         & -                         & -                         \\
                                                                        & Hook            & 0.709          & \textbf{0.959} & 0.931          & nan            & 0.709          & nan            & -              & -                         & -                         & -                         \\
                                                                        & Propeller       & 0.635          & 0.974          & 0.732          & \textbf{0.5}   & 0.641          & 0.6            & -              & -                         & -                         & -                         \\
                                                                        & Shampoo-bottle  & 0.705          & \textbf{0.951} & \textbf{0.951} & nan            & 0.706          & \textbf{0.9}   & -              & -                         & -                         & -                         \\
                                                                        & Standing-bottle & 0.702          & 1              & \textbf{0.861} & nan            & 0.702          & nan            & -              & -                         & -                         & -                         \\
                                                                        & Tire            & 0.784          & 0.998          & 0.919          & 0.5            & 0.791          & 0.3            & -              & -                         & -                         & -                         \\
                                                                        & Valve           & 0.706          & 0.932          & \textbf{0.862} & 0              & 0.736          & 0              & -              & -                         & -                         & -                         \\
                                                                        & Wall            & 0.709          & 0.925          & 0.831          & \textbf{0.283} & 0.724          & 0.796          & -              & -                         & -                         & -                         \\
                                                                        & all             & 0.669          & \textbf{0.953} & 0.812          & 0.407          & 0.659          & \textbf{0.552} & 0.736          & 0.515                     & 0.721                     & \textbf{0.57}             \\ \hline
\multirow{12}{*}{\begin{tabular}[c]{@{}c@{}}b2ub-\\ g25\end{tabular}}   & Bottle          & \textbf{0.762} & 0.986          & 0.975          & \textbf{0.9}   & \textbf{0.755} & nan            & -              & -                         & -                         & -                         \\
                                                                        & Can             & 0.444          & 0.751          & 0.426          & \textbf{0.551} & 0.429          & nan            & -              & -                         & -                         & -                         \\
                                                                        & Chain           & 0.683          & \textbf{0.972} & 0.822          & 0.735          & \textbf{0.527} & \textbf{0.71}  & -              & -                         & -                         & -                         \\
                                                                        & Drink-carton    & \textbf{0.604} & \textbf{0.992} & 0.736          & \textbf{0.606} & 0.614          & nan            & -              & -                         & -                         & -                         \\
                                                                        & Hook            & \textbf{0.724} & \textbf{0.953} & \textbf{0.953} & nan            & \textbf{0.724} & nan            & -              & -                         & -                         & -                         \\
                                                                        & Propeller       & \textbf{0.683} & 0.983          & \textbf{0.845} & 0.35           & 0.685          & \textbf{0.7}   & -              & -                         & -                         & -                         \\
                                                                        & Shampoo-bottle  & \textbf{0.729} & \textbf{0.959} & \textbf{0.903} & nan            & \textbf{0.757} & 0.3            & -              & -                         & -                         & -                         \\
                                                                        & Standing-bottle & \textbf{0.769} & 1              & \textbf{1}     & nan            & \textbf{0.769} & nan            & -              & -                         & -                         & -                         \\
                                                                        & Tire            & \textbf{0.811} & \textbf{0.999} & \textbf{0.955} & 0.6            & \textbf{0.817} & 0.5            & -              & -                         & -                         & -                         \\
                                                                        & Valve           & 0.709          & 0.918          & \textbf{0.872} & 0              & 0.74           & 0              & -              & -                         & -                         & -                         \\
                                                                        & Wall            & \textbf{0.714} & 0.92           & 0.83           & \textbf{0.347} & \textbf{0.716} & \textbf{0.823} & -              & -                         & -                         & -                         \\
                                                                        & all             & \textbf{0.694} & 0.948          & \textbf{0.847} & 0.511          & \textbf{0.685} & \textbf{0.506} & \textbf{0.76}  & \textbf{0.605}            & \textbf{0.753}            & \textbf{0.521}            \\ \hline
\multirow{12}{*}{\begin{tabular}[c]{@{}c@{}}b2ub-\\ g5-50\end{tabular}} & Bottle          & \textbf{0.755} & 0.985          & 0.962          & 0.267          & \textbf{0.749} & nan            & -              & -                         & -                         & -                         \\
                                                                        & Can             & 0.45           & 0.778          & \textbf{0.496} & 0.538          & 0.435          & nan            & -              & -                         & -                         & -                         \\
                                                                        & Chain           & \textbf{0.711} & \textbf{0.974} & \textbf{0.848} & \textbf{0.9}   & \textbf{0.546} & \textbf{0.732} & -              & -                         & -                         & -                         \\
                                                                        & Drink-carton    & \textbf{0.62}  & \textbf{0.991} & \textbf{0.77}  & \textbf{0.621} & \textbf{0.644} & nan            & -              & -                         & -                         & -                         \\
                                                                        & Hook            & \textbf{0.726} & \textbf{0.953} & 0.903          & nan            & \textbf{0.726} & nan            & -              & -                         & -                         & -                         \\
                                                                        & Propeller       & \textbf{0.684} & 0.978          & 0.784          & \textbf{0.6}   & \textbf{0.69}  & 0.6            & -              & -                         & -                         & -                         \\
                                                                        & Shampoo-bottle  & \textbf{0.766} & \textbf{0.971} & \textbf{0.971} & nan            & \textbf{0.771} & \textbf{0.7}   & -              & -                         & -                         & -                         \\
                                                                        & Standing-bottle & \textbf{0.745} & 1              & \textbf{1}     & nan            & \textbf{0.745} & nan            & -              & -                         & -                         & -                         \\
                                                                        & Tire            & \textbf{0.804} & 0.989          & \textbf{0.954} & 0.4            & \textbf{0.809} & \textbf{0.6}   & -              & -                         & -                         & -                         \\
                                                                        & Valve           & 0.698          & 0.93           & 0.837          & 0              & 0.729          & 0              & -              & -                         & -                         & -                         \\
                                                                        & Wall            & \textbf{0.725} & \textbf{0.931} & \textbf{0.832} & \textbf{0.356} & \textbf{0.73}  & 0.813          & -              & -                         & -                         & -                         \\
                                                                        & all             & \textbf{0.699} & \textbf{0.953} & \textbf{0.851} & 0.46           & \textbf{0.689} & \textbf{0.574} & \textbf{0.767} & 0.568                     & \textbf{0.762}            & \textbf{0.59}             \\ \hline
\end{tabular}
}
\caption{The MarinDebris dataset is detected using the SSD300 algorithm (Part 1 of 3). ``noisy" represents the original noise dataset, and dspnet, b2ub, and nei represent the target detection after denoising using the DSPNet, Blind2Unblind, and Neighbor2Neighbor single-frame denoising algorithms, respectively. Bold fonts indicate data that is higher than the corresponding ``noisy" dataset.}

\label{table17}
\end{table*}

\begin{table*}
\centering
\resizebox{\textwidth}{!}{
\begin{tabular}{cccccccccccc}
\hline
\multicolumn{1}{c|}{\multirow{12}{*}{\begin{tabular}[c]{@{}c@{}}b2ub-\\ p30\end{tabular}}}   & Bottle          & 0.745          & 0.984          & 0.973          & 0.233          & 0.74           & nan            & -              & -                         & -                         & -                         \\
\multicolumn{1}{c|}{}                                                                        & Can             & \textbf{0.455} & 0.751          & \textbf{0.498} & \textbf{0.585} & 0.429          & nan            & -              & -                         & -                         & -                         \\
\multicolumn{1}{c|}{}                                                                        & Chain           & 0.679          & 0.956          & \textbf{0.867} & 0.65           & \textbf{0.519} & \textbf{0.713} & -              & -                         & -                         & -                         \\
\multicolumn{1}{c|}{}                                                                        & Drink-carton    & \textbf{0.624} & \textbf{0.995} & \textbf{0.782} & \textbf{0.634} & 0.623          & nan            & -              & -                         & -                         & -                         \\
\multicolumn{1}{c|}{}                                                                        & Hook            & \textbf{0.727} & \textbf{0.956} & \textbf{0.956} & nan            & \textbf{0.727} & nan            & -              & -                         & -                         & -                         \\
\multicolumn{1}{c|}{}                                                                        & Propeller       & \textbf{0.679} & 0.988          & \textbf{0.851} & \textbf{0.5}   & 0.688          & 0.6            & -              & -                         & -                         & -                         \\
\multicolumn{1}{c|}{}                                                                        & Shampoo-bottle  & \textbf{0.74}  & \textbf{0.969} & \textbf{0.969} & nan            & \textbf{0.74}  & \textbf{0.8}   & -              & -                         & -                         & -                         \\
\multicolumn{1}{c|}{}                                                                        & Standing-bottle & \textbf{0.774} & 1              & 0.822          & nan            & \textbf{0.774} & nan            & -              & -                         & -                         & -                         \\
\multicolumn{1}{c|}{}                                                                        & Tire            & \textbf{0.811} & \textbf{0.999} & \textbf{0.954} & 0.2            & \textbf{0.819} & 0.5            & -              & -                         & -                         & -                         \\
\multicolumn{1}{c|}{}                                                                        & Valve           & 0.697          & 0.919          & \textbf{0.869} & 0              & 0.726          & 0              & -              & -                         & -                         & -                         \\
\multicolumn{1}{c|}{}                                                                        & Wall            & \textbf{0.718} & 0.922          & 0.821          & \textbf{0.351} & \textbf{0.724} & 0.81           & -              & -                         & -                         & -                         \\
\multicolumn{1}{c|}{}                                                                        & all             & \textbf{0.695} & 0.949          & \textbf{0.851} & 0.394          & \textbf{0.683} & \textbf{0.57}  & \textbf{0.759} & 0.493                     & \textbf{0.746}            & \textbf{0.588}            \\ \hline
\multicolumn{1}{c|}{\multirow{12}{*}{\begin{tabular}[c]{@{}c@{}}b2ub-\\ p5-50\end{tabular}}} & Bottle          & \textbf{0.754} & 0.985          & 0.962          & \textbf{0.8}   & \textbf{0.749} & nan            & -              & -                         & -                         & -                         \\
\multicolumn{1}{c|}{}                                                                         & Can             & \textbf{0.453} & 0.761          & \textbf{0.454} & 0.527          & \textbf{0.452} & nan            & -              & -                         & -                         & -                         \\
\multicolumn{1}{c|}{}                                                                         & Chain           & \textbf{0.684} & 0.95           & 0.815          & \textbf{0.9}   & 0.502          & \textbf{0.714} & -              & -                         & -                         & -                         \\
\multicolumn{1}{c|}{}                                                                         & Drink-carton    & \textbf{0.629} & \textbf{0.991} & \textbf{0.754} & \textbf{0.631} & \textbf{0.638} & nan            & -              & -                         & -                         & -                         \\
\multicolumn{1}{c|}{}                                                                         & Hook            & \textbf{0.722} & \textbf{0.953} & 0.934          & nan            & \textbf{0.722} & nan            & -              & -                         & -                         & -                         \\
\multicolumn{1}{c|}{}                                                                         & Propeller       & \textbf{0.697} & 0.983          & \textbf{0.837} & 0.3            & \textbf{0.705} & 0.6            & -              & -                         & -                         & -                         \\
\multicolumn{1}{c|}{}                                                                         & Shampoo-bottle  & \textbf{0.762} & 0.947          & \textbf{0.947} & nan            & \textbf{0.767} & \textbf{0.9}   & -              & -                         & -                         & -                         \\
\multicolumn{1}{c|}{}                                                                         & Standing-bottle & \textbf{0.767} & 1              & 0.822          & nan            & \textbf{0.767} & nan            & -              & -                         & -                         & -                         \\
\multicolumn{1}{c|}{}                                                                        & Tire            & \textbf{0.816} & 0.989          & \textbf{0.956} & 0.5            & \textbf{0.823} & 0.5            & -              & -                         & -                         & -                         \\
\multicolumn{1}{c|}{}                                                                         & Valve           & 0.705          & 0.93           & \textbf{0.858} & 0              & 0.735          & 0              & -              & -                         & -                         & -                         \\
\multicolumn{1}{c|}{}                                                                         & Wall            & \textbf{0.721} & 0.921          & \textbf{0.832} & \textbf{0.314} & \textbf{0.724} & \textbf{0.83}  & -              & -                         & -                         & -                         \\
\multicolumn{1}{c|}{}                                                                         & all             & \textbf{0.701} & 0.946          & \textbf{0.834} & 0.497          & \textbf{0.689} & \textbf{0.591} & \textbf{0.768} & \textbf{0.577}            & \textbf{0.759}            & \textbf{0.607}            \\ \hline

\multicolumn{1}{c|}{\multirow{12}{*}{\begin{tabular}[c]{@{}c@{}}nei-\\ g25\end{tabular}}}   & Bottle               & 0.744                & 0.986                & 0.953                & 0.7                  & 0.738                & nan                  & -                    & -                    & -                    & -                    \\
\multicolumn{1}{c|}{}                                                                       & Can                  & \textbf{0.459}       & \textbf{0.794}       & \textbf{0.498}       & \textbf{0.589}       & 0.435                & nan                  & -                    & -                    & -                    & -                    \\
\multicolumn{1}{c|}{}                                                                       & Chain                & \textbf{0.702}       & 0.955                & 0.822                & 0.75                 & \textbf{0.514}       & \textbf{0.733}       & -                    & -                    & -                    & -                    \\
\multicolumn{1}{c|}{}                                                                       & Drink-carton         & \textbf{0.614}       & \textbf{0.997}       & 0.734                & \textbf{0.613}       & 0.624                & nan                  & -                    & -                    & -                    & -                    \\
\multicolumn{1}{c|}{}                                                                       & Hook                 & \textbf{0.735}       & \textbf{0.954}       & \textbf{0.903}       & nan                  & \textbf{0.735}       & nan                  & -                    & -                    & -                    & -                    \\
\multicolumn{1}{c|}{}                                                                       & Propeller            & \textbf{0.708}       & 0.981                & 0.824                & 0.4                  & \textbf{0.717}       & \textbf{0.7}         & -                    & -                    & -                    & -                    \\
\multicolumn{1}{c|}{}                                                                       & Shampoo-bottle       & \textbf{0.745}       & \textbf{0.954}       & 0.954                & nan                  & \textbf{0.756}       & \textbf{0.7}         & -                    & -                    & -                    & -                    \\
\multicolumn{1}{c|}{}                                                                       & Standing-bottle      & \textbf{0.73}        & 1                    & \textbf{1}           & nan                  & \textbf{0.73}        & nan                  & -                    & -                    & -                    & -                    \\
\multicolumn{1}{c|}{}                                                                       & Tire                 & \textbf{0.809}       & 0.998                & \textbf{0.955}       & 0.6                  & \textbf{0.814}       & 0.4                  & -                    & -                    & -                    & -                    \\
\multicolumn{1}{c|}{}                                                                       & Valve                & \textbf{0.729}       & 0.939                & \textbf{0.88}        & 0.5                  & \textbf{0.754}       & 0                    & -                    & -                    & -                    & -                    \\
\multicolumn{1}{c|}{}                                                                       & Wall                 & \textbf{0.715}       & \textbf{0.93}        & 0.826                & \textbf{0.344}       & 0.709                & \textbf{0.836}       & -                    & -                    & -                    & -                    \\
\multicolumn{1}{c|}{}                                                                       & all                  & \textbf{0.699}       & \textbf{0.954}       & \textbf{0.85}        & \textbf{0.562}       & \textbf{0.684}       & \textbf{0.561}       & \textbf{0.762}       & \textbf{0.597}       & \textbf{0.749}       & \textbf{0.577}       \\ \hline
\multicolumn{1}{c|}{\multirow{12}{*}{\begin{tabular}[c]{@{}c@{}}nei-\\ g5-50\end{tabular}}} & Bottle               & \textbf{0.757}       & 0.986                & 0.974                & 0.35                 & \textbf{0.752}       & nan                  & -                    & -                    & -                    & -                    \\
\multicolumn{1}{c|}{}                                                                       & Can                  & 0.439                & 0.752                & 0.446                & 0.503                & 0.431                & nan                  & -                    & -                    & -                    & -                    \\
\multicolumn{1}{c|}{}                                                                       & Chain                & \textbf{0.698}       & \textbf{0.97}        & 0.839                & 0.8                  & 0.501                & \textbf{0.727}       & \textbf{-}           & -                    & -                    & -                    \\
\multicolumn{1}{c|}{}                                                                       & Drink-carton         & \textbf{0.611}       & \textbf{0.993}       & 0.739                & \textbf{0.612}       & \textbf{0.625}       & nan                  & -                    & -                    & -                    & -                    \\
\multicolumn{1}{c|}{}                                                                       & Hook                 & \textbf{0.727}       & 0.952                & 0.952                & nan                  & \textbf{0.727}       & nan                  & -                    & -                    & -                    & -                    \\
\multicolumn{1}{c|}{}                                                                       & Propeller            & \textbf{0.693}       & 0.977                & \textbf{0.861}       & \textbf{0.6}         & \textbf{0.696}       & \textbf{0.7}         & -                    & -                    & -                    & -                    \\
\multicolumn{1}{c|}{}                                                                       & Shampoo-bottle       & \textbf{0.759}       & \textbf{0.974}       & \textbf{0.912}       & nan                  & \textbf{0.778}       & \textbf{0.5}         & -                    & -                    & -                    & -                    \\
\multicolumn{1}{c|}{}                                                                       & Standing-bottle      & \textbf{0.749}       & 1                    & \textbf{1}           & nan                  & \textbf{0.749}       & nan                  & -                    & -                    & -                    & -                    \\
\multicolumn{1}{c|}{}                                                                       & Tire                 & \textbf{0.815}       & 0.99                 & \textbf{0.967}       & 0.6                  & \textbf{0.817}       & \textbf{0.6}         & -                    & -                    & -                    & -                    \\
\multicolumn{1}{c|}{}                                                                       & Valve                & 0.711                & 0.93                 & \textbf{0.868}       & 0.2                  & 0.741                & 0                    & -                    & -                    & -                    & -                    \\
\multicolumn{1}{c|}{}                                                                       & Wall                 & \textbf{0.722}       & \textbf{0.931}       & \textbf{0.844}       & \textbf{0.335}       & \textbf{0.72}        & \textbf{0.837}       & -                    & -                    & -                    & -                    \\
\multicolumn{1}{c|}{}                                                                       & all                  & \textbf{0.698}       & 0.95                 & \textbf{0.855}       & 0.5                  & \textbf{0.685}       & \textbf{0.561}       & \textbf{0.763}       & \textbf{0.592}       & \textbf{0.754}       & \textbf{0.576}       \\ \hline

\end{tabular}
}
\caption{The MarinDebris dataset is detected using the SSD300 algorithm (Part 2 of 3). ``noisy" represents the original noise dataset, and dspnet, b2ub, and nei represent the target detection after denoising using the DSPNet, Blind2Unblind, and Neighbor2Neighbor single-frame denoising algorithms, respectively. Bold fonts indicate data that is higher than the corresponding ``noisy" dataset.}

\label{table18}
\end{table*}

\begin{table*}
\centering
\resizebox{\textwidth}{!}{
\begin{tabular}{cccccccccccc}
\hline
\multicolumn{1}{c|}{\multirow{12}{*}{\begin{tabular}[c]{@{}c@{}}nei-\\ p30\end{tabular}}}   & Bottle               & 0.749                & 0.984                & 0.962                & 0.267                & 0.743                & nan                  & -                    & -                    & -                    & -                    \\
\multicolumn{1}{c|}{}                                                                       & Can                  & 0.444                & 0.734                & 0.43                 & \textbf{0.539}       & 0.431                & nan                  & -                    & -                    & -                    & -                    \\
\multicolumn{1}{c|}{}                                                                       & Chain                & 0.671                & \textbf{0.956}       & 0.843                & 0.783                & \textbf{0.534}       & 0.694                & -                    & -                    & -                    & -                    \\
\multicolumn{1}{c|}{}                                                                       & Drink-carton         & \textbf{0.645}       & \textbf{0.996}       & \textbf{0.836}       & \textbf{0.63}        & \textbf{0.682}       & nan                  & -                    & -                    & -                    & -                    \\
\multicolumn{1}{c|}{}                                                                       & Hook                 & 0.708                & \textbf{0.959}       & \textbf{0.959}       & nan                  & 0.708                & nan                  & -                    & -                    & -                    & -                    \\
\multicolumn{1}{c|}{}                                                                       & Propeller            & 0.665                & 0.988                & 0.772                & \textbf{0.5}         & 0.674                & 0.6                  & -                    & -                    & -                    & -                    \\
\multicolumn{1}{c|}{}                                                                       & Shampoo-bottle       & \textbf{0.773}       & \textbf{0.962}       & \textbf{0.962}       & nan                  & \textbf{0.776}       & \textbf{0.8}         & -                    & -                    & -                    & -                    \\
\multicolumn{1}{c|}{}                                                                       & Standing-bottle      & \textbf{0.729}       & 1                    & \textbf{1}           & nan                  & \textbf{0.729}       & nan                  & -                    & -                    & -                    & -                    \\
\multicolumn{1}{c|}{}                                                                       & Tire                 & \textbf{0.823}       & 0.999                & \textbf{0.962}       & 0.6                  & \textbf{0.829}       & 0.5                  & -                    & -                    & -                    & -                    \\
\multicolumn{1}{c|}{}                                                                       & Valve                & 0.71                 & 0.918                & \textbf{0.871}       & 0                    & 0.741                & 0                    & -                    & -                    & -                    & -                    \\
\multicolumn{1}{c|}{}                                                                       & Wall                 & \textbf{0.723}       & \textbf{0.931}       & 0.821                & \textbf{0.359}       & \textbf{0.72}        & \textbf{0.835}       & -                    & -                    & -                    & -                    \\
\multicolumn{1}{c|}{}                                                                       & all                  & \textbf{0.694}       & 0.948                & \textbf{0.856}       & 0.46                 & \textbf{0.688}       & \textbf{0.571}       & \textbf{0.763}       & \textbf{0.576}       & \textbf{0.756}       & \textbf{0.589}       \\ \hline
\multicolumn{1}{c|}{\multirow{12}{*}{\begin{tabular}[c]{@{}c@{}}nei-\\ p5-50\end{tabular}}} & Bottle               & \textbf{0.754}       & 0.984                & 0.951                & 0.4                  & 0.748                & nan                  & -                    & -                    & -                    & -                    \\
\multicolumn{1}{c|}{}                                                                       & Can                  & \textbf{0.461}       & 0.766                & 0.436                & 0.53                 & \textbf{0.448}       & nan                  & -                    & -                    & -                    & -                    \\
\multicolumn{1}{c|}{}                                                                       & Chain                & \textbf{0.698}       & 0.955                & \textbf{0.855}       & \textbf{0.834}       & \textbf{0.544}       & \textbf{0.723}       & -                    & -                    & -                    & -                    \\
\multicolumn{1}{c|}{}                                                                       & Drink-carton         & \textbf{0.624}       & \textbf{0.993}       & \textbf{0.811}       & \textbf{0.615}       & \textbf{0.64}        & nan                  & -                    & -                    & -                    & -                    \\
\multicolumn{1}{c|}{}                                                                       & Hook                 & 0.717                & \textbf{0.954}       & 0.906                & nan                  & 0.717                & nan                  & -                    & -                    & -                    & -                    \\
\multicolumn{1}{c|}{}                                                                       & Propeller            & \textbf{0.698}       & 0.986                & 0.804                & 0.4                  & \textbf{0.71}        & 0.6                  & -                    & -                    & -                    & -                    \\
\multicolumn{1}{c|}{}                                                                       & Shampoo-bottle       & \textbf{0.746}       & \textbf{0.955}       & 0.9                  & nan                  & \textbf{0.765}       & 0.5                  & -                    & -                    & -                    & -                    \\
\multicolumn{1}{c|}{}                                                                       & Standing-bottle      & \textbf{0.76}        & 1                    & \textbf{1}           & nan                  & \textbf{0.76}        & nan                  & -                    & -                    & -                    & -                    \\
\multicolumn{1}{c|}{}                                                                       & Tire                 & \textbf{0.81}        & 0.989                & \textbf{0.937}       & 0.4                  & \textbf{0.815}       & \textbf{0.6}         & -                    & -                    & -                    & -                    \\
\multicolumn{1}{c|}{}                                                                       & Valve                & 0.706                & 0.931                & 0.8                  & 0                    & 0.736                & 0                    & -                    & -                    & -                    & -                    \\
\multicolumn{1}{c|}{}                                                                       & Wall                 & \textbf{0.714}       & 0.923                & 0.816                & \textbf{0.337}       & 0.709                & \textbf{0.825}       & -                    & -                    & -                    & -                    \\
\multicolumn{1}{c|}{}                                                                       & all                  & \textbf{0.699}       & 0.949                & \textbf{0.838}       & 0.44                 & \textbf{0.69}        & \textbf{0.541}       & \textbf{0.76}        & 0.537                & \textbf{0.754}       & \textbf{0.558}       \\ \hline
\multicolumn{1}{l}{}                                                                        & \multicolumn{1}{l}{} & \multicolumn{1}{l}{} & \multicolumn{1}{l}{} & \multicolumn{1}{l}{} & \multicolumn{1}{l}{} & \multicolumn{1}{l}{} & \multicolumn{1}{l}{} & \multicolumn{1}{l}{} & \multicolumn{1}{l}{} & \multicolumn{1}{l}{} & \multicolumn{1}{l}{} \\
\multicolumn{1}{l}{}                                                                        & \multicolumn{1}{l}{} & \multicolumn{1}{l}{} & \multicolumn{1}{l}{} & \multicolumn{1}{l}{} & \multicolumn{1}{l}{} & \multicolumn{1}{l}{} & \multicolumn{1}{l}{} & \multicolumn{1}{l}{} & \multicolumn{1}{l}{} & \multicolumn{1}{l}{} & \multicolumn{1}{l}{}
\end{tabular}
}
\caption{The MarinDebris dataset is detected using the SSD300 algorithm (Part 3 of 3). ``noisy" represents the original noise dataset, and dspnet, b2ub, and nei represent the target detection after denoising using the DSPNet, Blind2Unblind, and Neighbor2Neighbor single-frame denoising algorithms, respectively. Bold fonts indicate data that is higher than the corresponding ``noisy" dataset.}

\label{table19}
\end{table*}

\begin{table*}
\centering
\resizebox{\textwidth}{!}{
\begin{tabular}{c|ccccccccccc}
\hline
\multirow{2}{*}{Denoising}                                              & \multicolumn{11}{c}{YOLOX}                                                                                                                                                                                                 \\ \cline{2-12} 
                                                                        & class           & mAP            & mAP@0.5        & mAP@0.75        & mAP\_s         & mAP\_m         & mAP\_l         & AR             & \multicolumn{1}{l}{AR\_s} & \multicolumn{1}{l}{AR\_m} & \multicolumn{1}{l}{AR\_l} \\ \hline
\multirow{12}{*}{noisy}                                                 & Bottle          & 0.74           & 0.986          & 0.964          & 0.3            & 0.735          & nan            & -              & -                         & -                         & -                         \\
                                                                        & Can             & 0.503          & 0.857          & 0.594          & 0.541          & 0.492          & nan            & -              & -                         & -                         & -                         \\
                                                                        & Chain           & 0.678          & 0.959          & 0.785          & 0.75           & 0.51           & 0.708          & -              & -                         & -                         & -                         \\
                                                                        & Drink-carton    & 0.621          & 0.999          & 0.729          & 0.619          & 0.632          & nan            & -              & -                         & -                         & -                         \\
                                                                        & Hook            & 0.725          & 0.967          & 0.933          & nan            & 0.725          & nan            & -              & -                         & -                         & -                         \\
                                                                        & Propeller       & 0.681          & 0.968          & 0.774          & 0.125          & 0.688          & 0.7            & -              & -                         & -                         & -                         \\
                                                                        & Shampoo-bottle  & 0.616          & 0.971          & 0.701          & nan            & 0.62           & 0.6            & -              & -                         & -                         & -                         \\
                                                                        & Standing-bottle & 0.712          & 1              & 0.851          & nan            & 0.712          & nan            & -              & -                         & -                         & -                         \\
                                                                        & Tire            & 0.799          & 0.99           & 0.937          & 0.8            & 0.804          & 0.4            & -              & -                         & -                         & -                         \\
                                                                        & Valve           & 0.654          & 0.927          & 0.8            & 0              & 0.683          & 0              & -              & -                         & -                         & -                         \\
                                                                        & Wall            & 0.69           & 0.939          & 0.845          & 0.384          & 0.685          & 0.756          & -              & -                         & -                         & -                         \\
                                                                        & all             & 0.675          & 0.96           & 0.81           & 0.44           & 0.662          & 0.527          & 0.736          & 0.564                     & 0.728                     & 0.544                     \\ \hline
\multirow{12}{*}{dspnet}                                                & Bottle          & 0.735          & 0.985          & \textbf{0.974} & 0.233          & 0.73           & nan            & -              & -                         & -                         & -                         \\
                                                                        & Can             & 0.432          & 0.748          & 0.489          & 0.489          & 0.424          & nan            & -              & -                         & -                         & -                         \\
                                                                        & Chain           & 0.664          & 0.944          & \textbf{0.819} & 0.651          & \textbf{0.527} & 0.699          & -              & -                         & -                         & -                         \\
                                                                        & Drink-carton    & 0.569          & 0.956          & 0.608          & 0.566          & 0.603          & nan            & -              & -                         & -                         & -                         \\
                                                                        & Hook            & 0.7            & 0.955          & 0.855          & nan            & 0.7            & nan            & -              & -                         & -                         & -                         \\
                                                                        & Propeller       & 0.658          & 0.945          & \textbf{0.838} & \textbf{0.25}  & 0.663          & 0.7            & -              & -                         & -                         & -                         \\
                                                                        & Shampoo-bottle  & 0.632          & 0.951          & \textbf{0.776} & nan            & \textbf{0.636} & 0.6            & -              & -                         & -                         & -                         \\
                                                                        & Standing-bottle & 0.657          & 1              & 0.832          & nan            & 0.657          & nan            & -              & -                         & -                         & -                         \\
                                                                        & Tire            & 0.798          & 0.979          & 0.936          & 0.7            & \textbf{0.805} & 0              & -              & -                         & -                         & -                         \\
                                                                        & Valve           & \textbf{0.668} & 0.927          & \textbf{0.822} & 0              & \textbf{0.699} & 0              & -              & -                         & -                         & -                         \\
                                                                        & Wall            & 0.666          & \textbf{0.929} & 0.83           & 0.332          & 0.666          & 0.734          & -              & -                         & -                         & -                         \\
                                                                        & all             & 0.653          & 0.938          & 0.798          & 0.403          & 0.646          & 0.456          & 0.712          & 0.526                     & 0.705                     & 0.475                     \\ \hline
\multirow{12}{*}{\begin{tabular}[c]{@{}c@{}}b2ub-\\ g25\end{tabular}}   & Bottle          & \textbf{0.746} & 0.984          & 0.964          & \textbf{0.6}   & \textbf{0.742} & nan            & -              & -                         & -                         & -                         \\
                                                                        & Can             & 0.475          & 0.838          & 0.485          & \textbf{0.552} & 0.461          & nan            & -              & -                         & -                         & -                         \\
                                                                        & Chain           & \textbf{0.687} & 0.947          & \textbf{0.814} & \textbf{0.8}   & \textbf{0.543} & \textbf{0.71}  & -              & -                         & -                         & -                         \\
                                                                        & Drink-carton    & \textbf{0.633} & 0.998          & \textbf{0.772} & \textbf{0.622} & \textbf{0.662} & nan            & -              & -                         & -                         & -                         \\
                                                                        & Hook            & \textbf{0.726} & 0.947          & 0.913          & nan            & \textbf{0.726} & nan            & -              & -                         & -                         & -                         \\
                                                                        & Propeller       & \textbf{0.686} & 0.964          & \textbf{0.833} & \textbf{0.5}   & \textbf{0.695} & 0.7            & -              & -                         & -                         & -                         \\
                                                                        & Shampoo-bottle  & \textbf{0.657} & 0.971          & \textbf{0.971} & nan            & \textbf{0.657} & \textbf{0.7}   & -              & -                         & -                         & -                         \\
                                                                        & Standing-bottle & \textbf{0.749} & 1              & 0.851          & nan            & \textbf{0.749} & nan            & -              & -                         & -                         & -                         \\
                                                                        & Tire            & \textbf{0.802} & 0.99           & \textbf{0.938} & 0.7            & \textbf{0.809} & 0.1            & -              & -                         & -                         & -                         \\
                                                                        & Valve           & \textbf{0.684} & \textbf{0.931} & \textbf{0.861} & 0              & \textbf{0.714} & 0              & -              & -                         & -                         & -                         \\
                                                                        & Wall            & \textbf{0.686} & 0.931          & 0.836          & 0.38           & 0.672          & \textbf{0.765} & -              & -                         & -                         & -                         \\
                                                                        & all             & \textbf{0.685} & 0.955          & \textbf{0.84}  & 0.519          & \textbf{0.675} & 0.496          & \textbf{0.744} & 0.562                     & \textbf{0.738}            & 0.513                     \\ \hline
\multirow{12}{*}{\begin{tabular}[c]{@{}c@{}}b2ub-\\ g5-50\end{tabular}} & Bottle          & 0.724          & 0.985          & 0.955          & 0.3            & 0.72           & nan            & -              & -                         & -                         & -                         \\
                                                                        & Can             & 0.487          & 0.825          & 0.562          & 0.539          & 0.481          & nan            & -              & -                         & -                         & -                         \\
                                                                        & Chain           & 0.676          & 0.946          & 0.775          & 0.725          & \textbf{0.538} & 0.705          & -              & -                         & -                         & -                         \\
                                                                        & Drink-carton    & 0.611          & 0.995          & 0.692          & 0.61           & 0.616          & nan            & -              & -                         & -                         & -                         \\
                                                                        & Hook            & 0.717          & 0.944          & 0.9            & nan            & 0.717          & nan            & -              & -                         & -                         & -                         \\
                                                                        & Propeller       & \textbf{0.713} & \textbf{0.985} & \textbf{0.841} & \textbf{0.15}  & \textbf{0.721} & 0.7            & -              & -                         & -                         & -                         \\
                                                                        & Shampoo-bottle  & \textbf{0.715} & 0.954          & \textbf{0.954} & nan            & \textbf{0.726} & 0.6            & -              & -                         & -                         & -                         \\
                                                                        & Standing-bottle & 0.709          & 1              & \textbf{0.861} & nan            & 0.709          & nan            & -              & -                         & -                         & -                         \\
                                                                        & Tire            & \textbf{0.82}  & 0.99           & \textbf{0.949} & \textbf{0.9}   & \textbf{0.826} & 0.2            & -              & -                         & -                         & -                         \\
                                                                        & Valve           & \textbf{0.71}  & \textbf{0.928} & \textbf{0.839} & 0              & \textbf{0.741} & 0              & -              & -                         & -                         & -                         \\
                                                                        & Wall            & \textbf{0.701} & 0.932          & 0.836          & 0.338          & \textbf{0.69}  & \textbf{0.779} & -              & -                         & -                         & -                         \\
                                                                        & all             & \textbf{0.689} & 0.953          & \textbf{0.833} & \textbf{0.445} & \textbf{0.68}  & 0.497          & \textbf{0.748} & \textbf{0.589}            & \textbf{0.739}            & 0.513                     \\ \hline

\end{tabular}
}
\caption{The MarinDebris dataset is detected using the YOLOX algorithm (Part 1 of 3). ``noisy" represents the original noise dataset, and dspnet, b2ub, and nei represent the target detection after denoising using the DSPNet, Blind2Unblind, and Neighbor2Neighbor single-frame denoising algorithms, respectively. Bold fonts indicate data that is higher than the corresponding ``noisy" dataset.}

\label{table20}
\end{table*}

\begin{table*}
\centering
\resizebox{\textwidth}{!}{
\begin{tabular}{cccccccccccc}
\hline
\multicolumn{1}{c|}{\multirow{12}{*}{\begin{tabular}[c]{@{}c@{}}b2ub-\\ p30\end{tabular}}}   & Bottle          & \textbf{0.754} & 0.986          & 0.964          & \textbf{0.7}   & \textbf{0.749} & nan            & -              & -                         & -                         & -                         \\
\multicolumn{1}{c|}{}                                                                        & Can             & 0.485          & 0.836          & 0.544          & \textbf{0.571} & 0.472          & nan            & -              & -                         & -                         & -                         \\
\multicolumn{1}{c|}{}                                                                        & Chain           & \textbf{0.688} & 0.947          & 0.809          & \textbf{0.735} & \textbf{0.524} & \textbf{0.72}  & -              & -                         & -                         & -                         \\
\multicolumn{1}{c|}{}                                                                        & Drink-carton    & 0.607          & 0.979          & 0.706          & 0.608          & 0.612          & nan            & -              & -                         & -                         & -                         \\
 \multicolumn{1}{c|}{}                                                                       & Hook            & \textbf{0.732} & 0.958          & \textbf{0.936} & nan            & \textbf{0.732} & nan            & -              & -                         & -                         & -                         \\
\multicolumn{1}{c|}{}                                                                        & Propeller       & \textbf{0.725} & \textbf{0.984} & \textbf{0.888} & \textbf{0.2}   & \textbf{0.733} & 0.7            & -              & -                         & -                         & -                         \\
\multicolumn{1}{c|}{}                                                                        & Shampoo-bottle  & \textbf{0.691} & 0.964          & \textbf{0.884} & nan            & \textbf{0.707} & 0.5            & -              & -                         & -                         & -                         \\
\multicolumn{1}{c|}{}                                                                        & Standing-bottle & \textbf{0.748} & 1              & \textbf{1}     & nan            & \textbf{0.748} & nan            & -              & -                         & -                         & -                         \\
\multicolumn{1}{c|}{}                                                                        & Tire            & \textbf{0.821} & 0.99           & \textbf{0.959} & 0.8            & \textbf{0.827} & 0.3            & -              & -                         & -                         & -                         \\
\multicolumn{1}{c|}{}                                                                        & Valve           & \textbf{0.683} & 0.923          & \textbf{0.835} & 0              & \textbf{0.713} & 0              & -              & -                         & -                         & -                         \\
\multicolumn{1}{c|}{}                                                                        & Wall            & \textbf{0.707} & 0.929          & \textbf{0.867} & \textbf{0.397} & \textbf{0.693} & \textbf{0.777} & -              & -                         & -                         & -                         \\
\multicolumn{1}{c|}{}                                                                        & all             & \textbf{0.695} & 0.954          & \textbf{0.854} & \textbf{0.501} & \textbf{0.683} & 0.5            & \textbf{0.752} & \textbf{0.6}              & \textbf{0.74}             & 0.517                     \\ \hline

\multicolumn{1}{c|}{\multirow{12}{*}{\begin{tabular}[c]{@{}c@{}}b2ub-\\ p5-50\end{tabular}}} & Bottle          & \textbf{0.752} & 0.983          & 0.961          & 0.3            & \textbf{0.747} & nan            & -              & -                         & -                         & -                         \\
\multicolumn{1}{c|}{}                                                                        & Can             & 0.478          & 0.824          & 0.522          & \textbf{0.586} & 0.446          & nan            & -              & -                         & -                         & -                         \\
\multicolumn{1}{c|}{}                                                                        & Chain           & \textbf{0.691} & 0.949          & \textbf{0.857} & \textbf{0.8}   & \textbf{0.548} & \textbf{0.721} & -              & -                         & -                         & -                         \\
\multicolumn{1}{c|}{}                                                                        & Drink-carton    & \textbf{0.637} & 0.997          & \textbf{0.792} & \textbf{0.626} & \textbf{0.671} & nan            & -              & -                         & -                         & -                         \\
\multicolumn{1}{c|}{}                                                                        & Hook            & 0.709          & 0.963          & 0.91           & nan            & 0.709          & nan            & -              & -                         & -                         & -                         \\
\multicolumn{1}{c|}{}                                                                        & Propeller       & \textbf{0.702} & \textbf{0.972} & \textbf{0.869} & 0.06           & \textbf{0.717} & 0.7            & -              & -                         & -                         & -                         \\
\multicolumn{1}{c|}{}                                                                        & Shampoo-bottle  & \textbf{0.687} & 0.951          & \textbf{0.875} & nan            & \textbf{0.694} & 0.6            & -              & -                         & -                         & -                         \\
\multicolumn{1}{c|}{}                                                                        & Standing-bottle & \textbf{0.72}  & 1              & \textbf{0.891} & nan            & \textbf{0.72}  & nan            & -              & -                         & -                         & -                         \\
\multicolumn{1}{c|}{}                                                                        & Tire            & \textbf{0.812} & \textbf{0.999} & \textbf{0.938} & 0.8            & \textbf{0.816} & 0.3            & -              & -                         & -                         & -                         \\
\multicolumn{1}{c|}{}                                                                        & Valve           & \textbf{0.686} & \textbf{0.946} & \textbf{0.836} & \textbf{0.65}  & \textbf{0.701} & 0              & -              & -                         & -                         & -                         \\
\multicolumn{1}{c|}{}                                                                        & Wall            & \textbf{0.701} & 0.93           & 0.835          & 0.369          & \textbf{0.702} & \textbf{0.769} & -              & -                         & -                         & -                         \\
\multicolumn{1}{c|}{}                                                                        & all             & \textbf{0.689} & 0.956          & \textbf{0.844} & \textbf{0.524} & \textbf{0.679} & 0.515          & \textbf{0.747} & \textbf{0.66}             & \textbf{0.739}            & 0.533                     \\ \hline
\multicolumn{1}{c|}{\multirow{12}{*}{\begin{tabular}[c]{@{}c@{}}nei-\\ g25\end{tabular}}}   & \textbf{Bottle}      & \textbf{0.759}       & 0.985                & 0.964                & \textbf{0.25}        & \textbf{0.755}       & nan                  & -                    & -                    & -                    & -                    \\
\multicolumn{1}{c|}{}                                                                       & Can                  & 0.483                & 0.793                & 0.534                & \textbf{0.58}        & 0.461                & nan                  & -                    & -                    & -                    & -                    \\
\multicolumn{1}{c|}{}                                                                       & Chain                & \textbf{0.701}       & 0.947                & \textbf{0.83}        & 0.651                & \textbf{0.627}       & \textbf{0.72}        & -                    & -                    & -                    & -                    \\
\multicolumn{1}{c|}{}                                                                       & Drink-carton         & \textbf{0.629}       & 0.997                & 0.715                & \textbf{0.622}       & \textbf{0.65}        & nan                  & -                    & -                    & -                    & -                    \\
\multicolumn{1}{c|}{}                                                                       & Hook                 & \textbf{0.747}       & 0.963                & \textbf{0.94}        & nan                  & \textbf{0.747}       & nan                  & -                    & -                    & -                    & -                    \\
\multicolumn{1}{c|}{}                                                                       & Propeller            & \textbf{0.688}       & \textbf{0.971}       & \textbf{0.834}       & 0.1                  & \textbf{0.7}         & 0.7                  & -                    & -                    & -                    & -                    \\
\multicolumn{1}{c|}{}                                                                       & Shampoo-bottle       & \textbf{0.693}       & 0.954                & \textbf{0.954}       & nan                  & \textbf{0.699}       & 0.6                  & -                    & -                    & -                    & -                    \\
\multicolumn{1}{c|}{}                                                                       & Standing-bottle      & 0.7                  & 1                    & \textbf{1}           & nan                  & 0.7                  & nan                  & -                    & -                    & -                    & -                    \\
\multicolumn{1}{c|}{}                                                                       & Tire                 & \textbf{0.829}       & \textbf{0.999}       & \textbf{0.959}       & 0.6                  & \textbf{0.835}       & 0.4                  & -                    & -                    & -                    & -                    \\
\multicolumn{1}{c|}{}                                                                       & Valve                & \textbf{0.676}       & 0.925                & \textbf{0.818}       & 0                    & \textbf{0.706}       & 0                    & -                    & -                    & -                    & -                    \\
\multicolumn{1}{c|}{}                                                                       & Wall                 & \textbf{0.717}       & 0.936                & \textbf{0.863}       & 0.384                & \textbf{0.707}       & \textbf{0.791}       & -                    & -                    & -                    & -                    \\
\multicolumn{1}{c|}{}                                                                       & all                  & \textbf{0.693}       & 0.952                & \textbf{0.856}       & 0.398                & \textbf{0.69}        & \textbf{0.535}       & \textbf{0.754}       & 0.509                & \textbf{0.752}       & \textbf{0.552}       \\ \hline
\multicolumn{1}{c|}{\multirow{12}{*}{\begin{tabular}[c]{@{}c@{}}nei-\\ g5-50\end{tabular}}} & \textbf{Bottle}      & \textbf{0.755}       & 0.985                & \textbf{0.962}       & \textbf{0.6}         & \textbf{0.75}        & nan                  & -                    & -                    & -                    & -                    \\
\multicolumn{1}{c|}{}                                                                       & Can                  & 0.49                 & 0.847                & 0.58                 & \textbf{0.572}       & 0.466                & nan                  & -                    & -                    & -                    & -                    \\
\multicolumn{1}{c|}{}                                                                       & Chain                & \textbf{0.686}       & \textbf{0.963}       & \textbf{0.824}       & \textbf{0.801}       & \textbf{0.526}       & \textbf{0.717}       & -                    & -                    & -                    & -                    \\
\multicolumn{1}{c|}{}                                                                       & Drink-carton         & \textbf{0.639}       & 0.998                & \textbf{0.81}        & \textbf{0.63}        & \textbf{0.662}       & nan                  & -                    & -                    & -                    & -                    \\
\multicolumn{1}{c|}{}                                                                       & Hook                 & \textbf{0.73}        & 0.959                & 0.924                & nan                  & \textbf{0.73}        & nan                  & -                    & -                    & -                    & -                    \\
\multicolumn{1}{c|}{}                                                                       & Propeller            & \textbf{0.69}        & \textbf{0.975}       & \textbf{0.808}       & 0.12                 & \textbf{0.7}         & 0.6                  & -                    & -                    & -                    & -                    \\
\multicolumn{1}{c|}{}                                                                       & Shampoo-bottle       & \textbf{0.672}       & 0.944                & \textbf{0.793}       & nan                  & \textbf{0.682}       & 0.5                  & -                    & -                    & -                    & -                    \\
\multicolumn{1}{c|}{}                                                                       & Standing-bottle      & \textbf{0.743}       & 1                    & \textbf{0.88}        & nan                  & \textbf{0.743}       & nan                  & -                    & -                    & -                    & -                    \\
\multicolumn{1}{c|}{}                                                                       & Tire                 & \textbf{0.832}       & \textbf{0.999}       & \textbf{0.966}       & 0.6                  & \textbf{0.84}        & 0.2                  & -                    & -                    & -                    & -                    \\
\multicolumn{1}{c|}{}                                                                       & Valve                & \textbf{0.683}       & 0.926                & \textbf{0.879}       & 0                    & \textbf{0.713}       & 0                    & -                    & -                    & -                    & -                    \\
\multicolumn{1}{c|}{}                                                                       & Wall                 & \textbf{0.669}       & 0.928                & 0.807                & \textbf{0.403}       & 0.669                & 0.729                & -                    & -                    & -                    & -                    \\
\multicolumn{1}{c|}{}                                                                       & all                  & \textbf{0.69}        & 0.957                & \textbf{0.839}       & \textbf{0.466}       & \textbf{0.68}        & 0.458                & \textbf{0.75}        & 0.561                & \textbf{0.739}       & 0.479                \\ \hline

\end{tabular}
}
\caption{The MarinDebris dataset is detected using the YOLOX algorithm (Part 2 of 3). ``noisy" represents the original noise dataset, and dspnet, b2ub, and nei represent the target detection after denoising using the DSPNet, Blind2Unblind, and Neighbor2Neighbor single-frame denoising algorithms, respectively. Bold fonts indicate data that is higher than the corresponding ``noisy" dataset.}

\label{table21}
\end{table*}

\begin{table*}
\centering
\resizebox{\textwidth}{!}{
\begin{tabular}{cccccccccccc}
\hline
\multicolumn{1}{c|}{\multirow{12}{*}{\begin{tabular}[c]{@{}c@{}}nei-\\ p30\end{tabular}}}   & Bottle               & \textbf{0.737}       & \textbf{0.987}       & \textbf{0.954}       & \textbf{0.6}         & 0.731                & nan                  & -                    & -                    & -                    & -                    \\
\multicolumn{1}{c|}{}                                                                       & Can                  & 0.481                & 0.811                & 0.555                & \textbf{0.562}       & 0.46                 & nan                  & -                    & -                    & -                    & -                    \\
\multicolumn{1}{c|}{}                                                                       & Chain                & \textbf{0.697}       & 0.946                & \textbf{0.82}        & \textbf{0.825}       & \textbf{0.55}        & \textbf{0.723}       & -                    & -                    & -                    & -                    \\
\multicolumn{1}{c|}{}                                                                       & Drink-carton         & \textbf{0.634}       & 0.999                & \textbf{0.78}        & \textbf{0.635}       & \textbf{0.643}       & nan                  & -                    & -                    & -                    & -                    \\
\multicolumn{1}{c|}{}                                                                       & Hook                 & 0.724                & 0.958                & \textbf{0.958}       & nan                  & \textbf{0.724}       & nan                  & -                    & -                    & -                    & -                    \\
\multicolumn{1}{c|}{}                                                                       & Propeller            & \textbf{0.692}       & \textbf{0.976}       & \textbf{0.848}       & \textbf{0.25}        & \textbf{0.698}       & 0.7                  & -                    & -                    & -                    & -                    \\
\multicolumn{1}{c|}{}                                                                       & Shampoo-bottle       & \textbf{0.668}       & 0.961                & \textbf{0.961}       & nan                  & \textbf{0.676}       & 0.6                  & -                    & -                    & -                    & -                    \\
\multicolumn{1}{c|}{}                                                                       & Standing-bottle      & \textbf{0.728}       & 1                    & 0.79                 & nan                  & \textbf{0.728}       & nan                  & -                    & -                    & -                    & -                    \\
\multicolumn{1}{c|}{}                                                                       & Tire                 & \textbf{0.818}       & \textbf{0.999}       & \textbf{0.957}       & 0.7                  & \textbf{0.824}       & 0.2                  & -                    & -                    & -                    & -                    \\
\multicolumn{1}{c|}{}                                                                       & Valve                & \textbf{0.655}       & 0.92                 & \textbf{0.813}       & 0                    & 0.683                & 0                    & -                    & -                    & -                    & -                    \\
\multicolumn{1}{c|}{}                                                                       & Wall                 & \textbf{0.701}       & 0.936                & \textbf{0.853}       & 0.383                & \textbf{0.693}       & \textbf{0.769}       & -                    & -                    & -                    & -                    \\
\multicolumn{1}{c|}{}                                                                       & all                  & \textbf{0.685}       & 0.954                & \textbf{0.844}       & \textbf{0.494}       & \textbf{0.674}       & 0.499                & \textbf{0.738}       & \textbf{0.569}       & 0.728                & 0.515                \\ \hline
\multicolumn{1}{c|}{\multirow{12}{*}{\begin{tabular}[c]{@{}c@{}}nei-\\ p5-50\end{tabular}}} & Bottle               & 0.739                & 0.986                & \textbf{0.952}       & \textbf{0.35}        & 0.733                & nan                  & -                    & -                    & -                    & -                    \\
\multicolumn{1}{c|}{}                                                                       & Can                  & 0.488                & 0.801                & 0.594                & \textbf{0.587}       & 0.465                & nan                  & -                    & -                    & -                    & -                    \\
\multicolumn{1}{c|}{}                                                                       & Chain                & \textbf{0.681}       & 0.944                & \textbf{0.827}       & \textbf{0.768}       & \textbf{0.523}       & 0.704                & -                    & -                    & -                    & -                    \\
\multicolumn{1}{c|}{}                                                                       & Drink-carton         & 0.615                & 0.999                & 0.691                & 0.596                & \textbf{0.655}       & nan                  & -                    & -                    & -                    & -                    \\
\multicolumn{1}{c|}{}                                                                       & Hook                 & 0.696                & 0.965                & 0.89                 & nan                  & 0.696                & nan                  & -                    & -                    & -                    & -                    \\
\multicolumn{1}{c|}{}                                                                       & Propeller            & \textbf{0.694}       & \textbf{0.984}       & \textbf{0.814}       & \textbf{0.2}         & \textbf{0.7}         & 0.7                  & -                    & -                    & -                    & -                    \\
\multicolumn{1}{c|}{}                                                                       & Shampoo-bottle       & \textbf{0.635}       & 0.951                & \textbf{0.849}       & nan                  & \textbf{0.632}       & 0.7                  & -                    & -                    & -                    & -                    \\
\multicolumn{1}{c|}{}                                                                       & Standing-bottle      & 0.701                & 1                    & \textbf{0.871}       & nan                  & 0.701                & nan                  & -                    & -                    & -                    & -                    \\
\multicolumn{1}{c|}{}                                                                       & Tire                 & 0.792                & 0.99                 & \textbf{0.938}       & 0.8                  & 0.797                & 0.2                  & -                    & -                    & -                    & -                    \\
\multicolumn{1}{c|}{}                                                                       & Valve                & 0.675                & 0.926                & \textbf{0.835}       & 0                    & \textbf{0.704}       & 0                    & -                    & -                    & -                    & -                    \\
\multicolumn{1}{c|}{}                                                                       & Wall                 & 0.672                & 0.937                & 0.823                & 0.344                & 0.678                & 0.723                & -                    & -                    & -                    & -                    \\
\multicolumn{1}{c|}{}                                                                       & all                  & 0.672                & 0.953                & \textbf{0.826}       & \textbf{0.456}       & 0.662                & 0.505                & 0.729                & \textbf{0.605}       & 0.724                & 0.522                \\ \hline
\multicolumn{1}{l}{}                                                                        & \multicolumn{1}{l}{} & \multicolumn{1}{l}{} & \multicolumn{1}{l}{} & \multicolumn{1}{l}{} & \multicolumn{1}{l}{} & \multicolumn{1}{l}{} & \multicolumn{1}{l}{} & \multicolumn{1}{l}{} & \multicolumn{1}{l}{} & \multicolumn{1}{l}{} & \multicolumn{1}{l}{} \\
\multicolumn{1}{l}{}                                                                        & \multicolumn{1}{l}{} & \multicolumn{1}{l}{} & \multicolumn{1}{l}{} & \multicolumn{1}{l}{} & \multicolumn{1}{l}{} & \multicolumn{1}{l}{} & \multicolumn{1}{l}{} & \multicolumn{1}{l}{} & \multicolumn{1}{l}{} & \multicolumn{1}{l}{} & \multicolumn{1}{l}{}
\end{tabular}
}
\caption{The MarinDebris dataset is detected using the YOLOX algorithm (Part 3 of 3). ``noisy" represents the original noise dataset, and dspnet, b2ub, and nei represent the target detection after denoising using the DSPNet, Blind2Unblind, and Neighbor2Neighbor single-frame denoising algorithms, respectively. Bold fonts indicate data that is higher than the corresponding ``noisy" dataset.}

\label{table22}
\end{table*}

\footnotesize{
\begin{table*}
\centering
\resizebox{\textwidth}{!}{
\begin{tabular}{c|cccccccccccll}
\cline{1-12}
\multirow{2}{*}{Denoising}                                             & \multicolumn{11}{c}{SSDMobileNetV2}                                                                                                                                                                       &                      &                      \\ \cline{2-12}
                                                                       & class    & mAP            & mAP@0.5        & mAP@0.75        & mAP\_s & mAP\_m         & mAP\_l         & AR             & \multicolumn{1}{l}{AR\_s} & \multicolumn{1}{l}{AR\_m} & \multicolumn{1}{l}{AR\_l} &                      &                      \\ \cline{1-12}
\multirow{4}{*}{noisy}                                                 & victim   & 0.825          & 1              & 0.96           & nan    & 0.713          & 0.828          & -              & -                         & -                         & -                         &                      &                      \\
                                                                       & plane    & 0.779          & 1              & 0.953          & nan    & 0.781          & 0.725          & -              & -                         & -                         & -                         &                      &                      \\
                                                                       & boat     & 0.733          & 0.99           & 0.938          & 0.9    & 0.732          & nan            & -              & -                         & -                         & -                         &                      &                      \\
                                                                       & all      & 0.779          & 0.996          & 0.95           & 0.9    & 0.742          & 0.776          & 0.82           & 0.9                       & 0.774                     & 0.804                     &                      &                      \\ \cline{1-12}
\multirow{4}{*}{dspnet}                                                & victim   & 0.824          & 1              & \textbf{0.962} & nan    & 0.707          & 0.827          & -              & -                         & -                         & -                         &                      &                      \\
                                                                       & plane    & 0.772          & 1              & 0.95           & nan    & 0.773          & \textbf{0.801} & -              & -                         & -                         & -                         &                      &                      \\
                                                                       & boat     & 0.717          & 0.989          & 0.91           & 0.8    & 0.718          & nan            & -              & -                         & -                         & -                         &                      &                      \\
                                                                       & all      & 0.771          & 0.996          & 0.941          & 0.8    & 0.733          & \textbf{0.814} & 0.819          & 0.8                       & 0.77                      & \textbf{0.832}            &                      &                      \\ \cline{1-12}
\multirow{4}{*}{\begin{tabular}[c]{@{}c@{}}b2ub-\\ g25\end{tabular}}   & victim   & 0.812          & 1              & 0.951          & nan    & 0.647          & 0.817          & -              & -                         & -                         & -                         &                      &                      \\
                                                                       & plane    & 0.738          & 1              & 0.929          & nan    & 0.74           & 0.7            & -              & -                         & -                         & -                         &                      &                      \\
                                                                       & boat     & 0.592          & 0.98           & 0.653          & 0.7    & 0.592          & nan            & -              & -                         & -                         & -                         &                      &                      \\
                                                                       & all      & 0.714          & 0.993          & 0.844          & 0.7    & 0.66           & 0.758          & 0.758          & 0.7                       & 0.7                       & 0.825                     &                      &                      \\ \cline{1-12}
\multirow{4}{*}{\begin{tabular}[c]{@{}c@{}}b2ub-\\ g5-50\end{tabular}} & victim   & \textbf{0.831} & 1              & \textbf{0.963} & nan    & \textbf{0.769} & \textbf{0.833} & -              & -                         & -                         & -                         &                      &                      \\
                                                                       & plane    & 0.777          & 1              & \textbf{0.965} & nan    & 0.778          & \textbf{0.75}  & -              & -                         & -                         & -                         &                      &                      \\
                                                                       & boat     & 0.716          & 0.99           & 0.913          & 0.8    & 0.716          & nan            & -              & -                         & -                         & -                         &                      &                      \\
                                                                       & all      & 0.775          & \textbf{0.997} & 0.947          & 0.8    & \textbf{0.754} & \textbf{0.792} & 0.818          & 0.8                       & \textbf{0.788}            & \textbf{0.807}            &                      &                      \\ \cline{1-12}
\multirow{4}{*}{\begin{tabular}[c]{@{}c@{}}b2ub-\\ p30\end{tabular}}   & victim   & 0.778          & 0.999          & 0.921          & nan    & 0.565          & 0.784          & -              & -                         & -                         & -                         &                      &                      \\
                                                                       & plane    & 0.7            & 0.989          & 0.944          & nan    & 0.7            & 0.767          & -              & -                         & -                         & -                         &                      &                      \\
                                                                       & boat     & 0.59           & 0.989          & 0.655          & 0.8    & 0.589          & nan            & -              & -                         & -                         & -                         &                      &                      \\
                                                                       & all      & 0.689          & 0.993          & 0.84           & 0.8    & 0.618          & 0.775          & 0.737          & 0.8                       & 0.657                     & \textbf{0.809}            &                      &                      \\ \cline{1-12}
\multirow{4}{*}{\begin{tabular}[c]{@{}c@{}}b2ub-\\ p5-50\end{tabular}} & victim   & 0.807          & 1              & \textbf{0.936} & nan    & 0.653          & 0.809          & -              & -                         & -                         & -                         &                      &                      \\
                                                                       & plane    & 0.742          & 1              & 0.942          & nan    & 0.744          & \textbf{0.759} & -              & -                         & -                         & -                         &                      &                      \\
                                                                       & boat     & 0.669          & 0.99           & 0.823          & 0.7    & 0.668          & nan            & -              & -                         & -                         & -                         &                      &                      \\
                                                                       & all      & 0.739          & \textbf{0.997} & 0.901          & 0.7    & 0.688          & \textbf{0.784} & 0.787          & 0.7                       & 0.74                      & \textbf{0.846}            & \multicolumn{1}{c}{} & \multicolumn{1}{c}{} \\ \cline{1-12}
\multirow{4}{*}{\begin{tabular}[c]{@{}c@{}}nei-\\ g25\end{tabular}}    & victim   & 0.809          & 1              & \textbf{0.967} & nan    & 0.671          & 0.813          & -              & -                         & -                         & -                         &                      &                      \\
                                                                       & plane    & 0.725          & 1              & 0.918          & nan    & 0.731          & 0.683          & -              & -                         & -                         & -                         &                      &                      \\
                                                                       & boat     & 0.58           & 0.989          & 0.688          & 0.5    & 0.58           & nan            & -              & -                         & -                         & -                         &                      &                      \\
                                                                       & all      & 0.705          & 0.996          & 0.858          & 0.5    & 0.661          & 0.748          & 0.754          & 0.5                       & 0.705                     & 0.775                     &                      &                      \\ \cline{1-12}
\multirow{4}{*}{\begin{tabular}[c]{@{}c@{}}nei-\\ g5-50\end{tabular}}  & ship     & 0.516          & 0.783          & 0.603          & 0      & 0.466          & 0.55           & -              & -                         & -                         & -                         &                      &                      \\
                                                                       & aircraft & 0.368          & 0.739          & 0.347          & nan    & 0.552          & 0.326          & -              & -                         & -                         & -                         &                      &                      \\
                                                                       & human    & 0.35           & 0.487          & 0.446          & 0.8    & 0.314          & 0.303          & -              & -                         & -                         & -                         &                      &                      \\
                                                                       & all      & 0.411          & 0.67           & 0.465          & 0.4    & 0.444          & 0.393          & 0.477          & 0.4                       & 0.452                     & 0.46                      &                      &                      \\ \cline{1-12}
\multirow{4}{*}{\begin{tabular}[c]{@{}c@{}}nei-\\ p30\end{tabular}}    & victim   & 0.751          & 0.989          & 0.899          & nan    & 0.533          & 0.756          & -              & -                         & -                         & -                         &                      &                      \\
                                                                       & plane    & 0.696          & 0.999          & 0.866          & nan    & 0.695          & 0.701          & -              & -                         & -                         & -                         &                      &                      \\
                                                                       & boat     & 0.539          & 0.988          & 0.526          & 0.2    & 0.54           & nan            & -              & -                         & -                         & -                         &                      &                      \\
                                                                       & all      & 0.662          & 0.992          & 0.764          & 0.2    & 0.59           & 0.728          & 0.712          & 0.2                       & 0.665                     & 0.745                     &                      &                      \\ \cline{1-12}
\multirow{4}{*}{\begin{tabular}[c]{@{}c@{}}nei-\\ p5-50\end{tabular}}  & victim   & \textbf{0.827} & 1              & \textbf{0.971} & nan    & 0.681          & \textbf{0.832} & -              & -                         & -                         & -                         &                      &                      \\
                                                                       & plane    & \textbf{0.796} & 1              & \textbf{0.968} & nan    & \textbf{0.798} & \textbf{0.801} & -              & -                         & -                         & -                         &                      &                      \\
                                                                       & boat     & \textbf{0.74}  & 0.99           & \textbf{0.94}  & 0.7    & 0.741          & nan            & -              & -                         & -                         & -                         &                      &                      \\
                                                                       & all      & \textbf{0.788} & \textbf{0.997} & \textbf{0.96}  & 0.7    & 0.74           & \textbf{0.816} & \textbf{0.826} & 0.7                       & \textbf{0.778}            & \textbf{0.831}            & \multicolumn{1}{c}{} & \multicolumn{1}{c}{} \\ \cline{1-12}
\end{tabular}
}
\caption{The FDD dataset is detected using the SSDMobileNetV2 algorithm. ``noisy" represents the original noise dataset, and dspnet, b2ub, and nei represent the target detection after denoising using the DSPNet, Blind2Unblind, and Neighbor2Neighbor single-frame denoising algorithms, respectively. Bold fonts indicate data that is higher than the corresponding ``noisy" dataset.}

\label{table23}
\end{table*}
}

\begin{table*}
\centering
\resizebox{\textwidth}{!}{
\begin{tabular}{c|ccccccccccc}
\hline
\multirow{2}{*}{Denoising}                                             & \multicolumn{11}{c}{Faster R-CNN}                                                                                                                                                                           \\ \cline{2-12} 
                                                                       & class    & mAP            & mAP@0.5        & mAP@0.75        & mAP\_s & mAP\_m         & mAP\_l         & AR             & \multicolumn{1}{l}{AR\_s} & \multicolumn{1}{l}{AR\_m} & \multicolumn{1}{l}{AR\_l} \\ \hline
\multirow{4}{*}{noisy}                                                 & victim   & 0.807          & 0.99           & 0.935          & nan    & 0.666          & 0.809          & -              & -                         & -                         & -                         \\
                                                                       & plane    & 0.879          & 1              & 0.987          & nan    & 0.88           & 0.725          & -              & -                         & -                         & -                         \\
                                                                       & boat     & 0.849          & 0.987          & 0.987          & 0.9    & 0.849          & nan            & -              & -                         & -                         & -                         \\
                                                                       & all      & 0.845          & 0.992          & 0.969          & 0.9    & 0.798          & 0.767          & 0.887          & 0.9                       & 0.827                     & 0.803                     \\ \hline
\multirow{4}{*}{dspnet}                                                & victim   & 0.802          & \textbf{1}     & 0.928          & nan    & 0.532          & 0.807          & -              & -                         & -                         & -                         \\
                                                                       & plane    & 0.87           & 1              & \textbf{1}     & nan    & 0.873          & \textbf{0.801} & -              & -                         & -                         & -                         \\
                                                                       & boat     & 0.833          & 0.987          & 0.987          & 0.9    & 0.833          & nan            & -              & -                         & -                         & -                         \\
                                                                       & all      & 0.835          & \textbf{0.996} & \textbf{0.972} & 0.9    & 0.746          & \textbf{0.804} & 0.876          & 0.9                       & 0.795                     & \textbf{0.825}            \\ \hline
\multirow{4}{*}{\begin{tabular}[c]{@{}c@{}}b2ub-\\ g25\end{tabular}}   & victim   & \textbf{0.81}  & \textbf{1}     & \textbf{0.947} & nan    & 0.641          & \textbf{0.815} & -              & -                         & -                         & -                         \\
                                                                       & plane    & \textbf{0.88}  & 1              & 0.986          & nan    & \textbf{0.883} & \textbf{0.75}  & -              & -                         & -                         & -                         \\
                                                                       & boat     & 0.84           & \textbf{0.988} & \textbf{0.988} & 0.8    & 0.84           & nan            & -              & -                         & -                         & -                         \\
                                                                       & all      & 0.843          & \textbf{0.996} & \textbf{0.974} & 0.8    & 0.788          & \textbf{0.783} & 0.878          & 0.8                       & 0.816                     & \textbf{0.825}            \\ \hline
\multirow{4}{*}{\begin{tabular}[c]{@{}c@{}}b2ub-\\ g5-50\end{tabular}} & victim   & \textbf{0.815} & \textbf{1}     & \textbf{0.942} & nan    & \textbf{0.714} & \textbf{0.817} & -              & -                         & -                         & -                         \\
                                                                       & plane    & \textbf{0.88}  & 1              & \textbf{1}     & nan    & \textbf{0.881} & \textbf{0.75}  & -              & -                         & -                         & -                         \\
                                                                       & boat     & \textbf{0.855} & 0.987          & 0.987          & 0.9    & \textbf{0.855} & nan            & -              & -                         & -                         & -                         \\
                                                                       & all      & \textbf{0.85}  & \textbf{0.996} & \textbf{0.977} & 0.9    & \textbf{0.817} & \textbf{0.784} & \textbf{0.89}  & 0.9                       & \textbf{0.843}            & \textbf{0.83}             \\ \hline
\multirow{4}{*}{\begin{tabular}[c]{@{}c@{}}b2ub-\\ p30\end{tabular}}   & victim   & \textbf{0.822} & \textbf{1}     & \textbf{0.954} & nan    & 0.661          & \textbf{0.825} & -              & -                         & -                         & -                         \\
                                                                       & plane    & \textbf{0.885} & 1              & 0.987          & nan    & \textbf{0.886} & \textbf{0.85}  & -              & -                         & -                         & -                         \\
                                                                       & boat     & \textbf{0.855} & 0.987          & 0.987          & 0.9    & \textbf{0.855} & nan            & -              & -                         & -                         & -                         \\
                                                                       & all      & \textbf{0.854} & \textbf{0.996} & \textbf{0.976} & 0.9    & \textbf{0.801} & \textbf{0.838} & \textbf{0.89}  & 0.9                       & \textbf{0.829}            & \textbf{0.854}            \\ \hline
\multirow{4}{*}{\begin{tabular}[c]{@{}c@{}}b2ub-\\ p5-50\end{tabular}} & victim   & 0.804          & 0.99           & \textbf{0.947} & nan    & 0.516          & \textbf{0.811} & -              & -                         & -                         & -                         \\
                                                                       & plane    & 0.86           & 1              & \textbf{1}     & nan    & 0.861          & \textbf{0.8}   & -              & -                         & -                         & -                         \\
                                                                       & boat     & 0.836          & \textbf{0.988} & 0.978          & 0.9    & 0.836          & nan            & -              & -                         & -                         & -                         \\
                                                                       & all      & 0.833          & \textbf{0.993} & \textbf{0.975} & 0.9    & 0.737          & \textbf{0.806} & 0.87           & 0.9                       & 0.768                     & \textbf{0.822}            \\ \hline
\multirow{4}{*}{\begin{tabular}[c]{@{}c@{}}nei-\\ g25\end{tabular}}    & victim   & 0.804          & \textbf{1}     & \textbf{0.939} & nan    & 0.638          & 0.809          & -              & -                         & -                         & -                         \\
                                                                       & plane    & 0.875          & 1              & 0.986          & nan    & 0.877          & \textbf{0.75}  & -              & -                         & -                         & -                         \\
                                                                       & boat     & 0.844          & \textbf{0.988} & \textbf{0.988} & 0.9    & 0.844          & nan            & -              & -                         & -                         & -                         \\
                                                                       & all      & 0.841          & \textbf{0.996} & \textbf{0.971} & 0.9    & 0.786          & \textbf{0.78}  & 0.881          & 0.9                       & 0.82                      & 0.799                     \\ \hline
\multirow{4}{*}{\begin{tabular}[c]{@{}c@{}}nei-\\ g5-50\end{tabular}}  & ship     & \textbf{0.822} & \textbf{1}     & \textbf{0.954} & nan    & 0.653          & \textbf{0.826} & -              & -                         & -                         & -                         \\
                                                                       & aircraft & \textbf{0.881} & 1              & \textbf{0.99}  & nan    & \textbf{0.882} & \textbf{0.75}  & -              & -                         & -                         & -                         \\
                                                                       & human    & 0.844          & 0.986          & 0.986          & 0.9    & 0.844          & nan            & -              & -                         & -                         & -                         \\
                                                                       & all      & \textbf{0.849} & \textbf{0.995} & \textbf{0.977} & 0.9    & 0.793          & \textbf{0.788} & \textbf{0.889} & 0.9                       & 0.822                     & \textbf{0.833}            \\ \hline
\multirow{4}{*}{\begin{tabular}[c]{@{}c@{}}nei-\\ p30\end{tabular}}    & victim   & 0.797          & 0.99           & 0.927          & nan    & 0.612          & 0.803          & -              & -                         & -                         & -                         \\
                                                                       & plane    & 0.866          & 1              & \textbf{0.988} & nan    & 0.867          & \textbf{0.75}  & -              & -                         & -                         & -                         \\
                                                                       & boat     & 0.839          & 0.986          & 0.986          & 0.9    & 0.839          & nan            & -              & -                         & -                         & -                         \\
                                                                       & all      & 0.834          & 0.992          & 0.967          & 0.9    & 0.773          & \textbf{0.777} & 0.873          & 0.9                       & 0.8                       & 0.795                     \\ \hline
\multirow{4}{*}{\begin{tabular}[c]{@{}c@{}}nei-\\ p5-50\end{tabular}}  & victim   & \textbf{0.827} & \textbf{1}     & \textbf{0.965} & nan    & 0.599          & \textbf{0.834} & -              & -                         & -                         & -                         \\
                                                                       & plane    & \textbf{0.891} & 1              & 0.987          & nan    & \textbf{0.893} & \textbf{0.75}  & -              & -                         & -                         & -                         \\
                                                                       & boat     & \textbf{0.851} & 0.986          & 0.986          & 0.9    & \textbf{0.851} & nan            & -              & -                         & -                         & -                         \\
                                                                       & all      & \textbf{0.856} & \textbf{0.995} & \textbf{0.979} & 0.9    & 0.781          & \textbf{0.792} & \textbf{0.89}  & 0.9                       & 0.804                     & \textbf{0.832}            \\ \hline
\end{tabular}
}
\caption{The FDD dataset is detected using the Faster R-CNN algorithm. ``noisy" represents the original noise dataset, and dspnet, b2ub, and nei represent the target detection after denoising using the DSPNet, Blind2Unblind, and Neighbor2Neighbor single-frame denoising algorithms, respectively. Bold fonts indicate data that is higher than the corresponding ``noisy" dataset.}
\label{table24}
\end{table*}

\begin{table*}
\centering
\resizebox{\textwidth}{!}{
\begin{tabular}{c|ccccccccccc}
\hline
\multirow{2}{*}{Denoising}                                             & \multicolumn{11}{c}{SSD300}                                                                                                                                                                                       \\ \cline{2-12} 
                                                                       & class    & mAP            & mAP@0.5        & mAP@0.75        & mAP\_s       & mAP\_m         & mAP\_l         & AR             & \multicolumn{1}{l}{AR\_s} & \multicolumn{1}{l}{AR\_m} & \multicolumn{1}{l}{AR\_l} \\ \hline
\multirow{4}{*}{noisy}                                                 & victim   & 0.805          & 1              & 0.956          & nan          & 0.634          & 0.81           & -              & -                         & -                         & -                         \\
                                                                       & plane    & 0.842          & 1              & 0.975          & nan          & 0.843          & 0.8            & -              & -                         & -                         & -                         \\
                                                                       & boat     & 0.813          & 0.989          & 0.974          & 0.85         & 0.814          & nan            & -              & -                         & -                         & -                         \\
                                                                       & all      & 0.82           & 0.996          & 0.968          & 0.85         & 0.764          & 0.805          & 0.861          & 0.9                       & 0.8                       & 0.848                     \\ \hline
\multirow{4}{*}{dspnet}                                                & victim   & 0.802          & 1              & 0.938          & nan          & \textbf{0.66}  & 0.806          & -              & -                         & -                         & -                         \\
                                                                       & plane    & 0.833          & 1              & 0.973          & nan          & 0.835          & 0.775          & -              & -                         & -                         & -                         \\
                                                                       & boat     & 0.797          & 0.989          & \textbf{0.977} & 0.8          & 0.797          & nan            & -              & -                         & -                         & -                         \\
                                                                       & all      & 0.811          & 0.996          & 0.963          & 0.8          & 0.764          & 0.79           & 0.851          & 0.8                       & 0.799                     & \textbf{0.872}            \\ \hline
\multirow{4}{*}{\begin{tabular}[c]{@{}c@{}}b2ub-\\ g25\end{tabular}}   & victim   & 0.805          & 1              & 0.956          & nan          & 0.613          & \textbf{0.811} & -              & -                         & -                         & -                         \\
                                                                       & plane    & \textbf{0.847} & 1              & \textbf{0.977} & nan          & \textbf{0.85}  & 0.767          & -              & -                         & -                         & -                         \\
                                                                       & boat     & 0.811          & \textbf{0.99}  & \textbf{0.978} & \textbf{0.9} & 0.811          & nan            & -              & -                         & -                         & -                         \\
                                                                       & all      & \textbf{0.821} & \textbf{0.997} & \textbf{0.97}  & \textbf{0.9} & 0.758          & 0.789          & \textbf{0.863} & 0.9                       & 0.795                     & \textbf{0.851}            \\ \hline
\multirow{4}{*}{\begin{tabular}[c]{@{}c@{}}b2ub-\\ g5-50\end{tabular}} & victim   & \textbf{0.814} & 1              & \textbf{0.963} & nan          & \textbf{0.678} & \textbf{0.817} & -              & -                         & -                         & -                         \\
                                                                       & plane    & \textbf{0.85}  & 1              & \textbf{0.987} & nan          & \textbf{0.851} & 0.8            & -              & -                         & -                         & -                         \\
                                                                       & boat     & \textbf{0.814} & \textbf{0.99}  & \textbf{0.975} & \textbf{0.9} & 0.814          & nan            & -              & -                         & -                         & -                         \\
                                                                       & all      & \textbf{0.826} & \textbf{0.997} & \textbf{0.975} & \textbf{0.9} & \textbf{0.781} & \textbf{0.809} & \textbf{0.864} & 0.9                       & \textbf{0.815}            & \textbf{0.851}            \\ \hline
\multirow{4}{*}{\begin{tabular}[c]{@{}c@{}}b2ub-\\ p30\end{tabular}}   & victim   & \textbf{0.807} & 1              & 0.947          & nan          & \textbf{0.678} & \textbf{0.811} & -              & -                         & -                         & -                         \\
                                                                       & plane    & \textbf{0.857} & 1              & \textbf{0.989} & nan          & \textbf{0.859} & 0.725          & -              & -                         & -                         & -                         \\
                                                                       & boat     & \textbf{0.817} & 0.989          & \textbf{0.976} & \textbf{0.9} & \textbf{0.817} & nan            & -              & -                         & -                         & -                         \\
                                                                       & all      & \textbf{0.827} & 0.996          & \textbf{0.971} & \textbf{0.9} & \textbf{0.785} & 0.768          & \textbf{0.866} & 0.9                       & \textbf{0.813}            & 0.801                     \\ \hline
\multirow{4}{*}{\begin{tabular}[c]{@{}c@{}}b2ub-\\ p5-50\end{tabular}} & victim   & 0.791          & 1              & 0.952          & nan          & 0.626          & 0.796          & -              & -                         & -                         & -                         \\
                                                                       & plane    & 0.805          & 1              & 0.974          & nan          & 0.805          & 0.725          & -              & -                         & -                         & -                         \\
                                                                       & boat     & 0.78           & 0.989          & 0.975          & 0.8          & 0.781          & nan            & -              & -                         & -                         & -                         \\
                                                                       & all      & 0.792          & 0.996          & 0.967          & 0.8          & 0.737          & 0.761          & 0.83           & 0.8                       & 0.784                     & 0.789                     \\ \hline
\multirow{4}{*}{\begin{tabular}[c]{@{}c@{}}nei-\\ g25\end{tabular}}    & victim   & \textbf{0.811} & 1              & \textbf{0.966} & nan          & \textbf{0.699} & \textbf{0.813} & -              & -                         & -                         & -                         \\
                                                                       & plane    & \textbf{0.85}  & 1              & \textbf{0.988} & nan          & \textbf{0.853} & 0.719          & -              & -                         & -                         & -                         \\
                                                                       & boat     & \textbf{0.814} & 0.989          & 0.974          & 0.7          & \textbf{0.815} & nan            & -              & -                         & -                         & -                         \\
                                                                       & all      & \textbf{0.825} & 0.996          & \textbf{0.976} & 0.7          & \textbf{0.789} & 0.766          & \textbf{0.864} & 0.7                       & \textbf{0.823}            & 0.827                     \\ \hline
\multirow{4}{*}{\begin{tabular}[c]{@{}c@{}}nei-\\ g5-50\end{tabular}}  & ship     & 0.805          & 1              & \textbf{0.975} & nan          & 0.616          & 0.81           & -              & -                         & -                         & -                         \\
                                                                       & aircraft & \textbf{0.845} & 1              & \textbf{0.989} & nan          & \textbf{0.848} & 0.717          & -              & -                         & -                         & -                         \\
                                                                       & human    & 0.809          & \textbf{0.99}  & \textbf{0.975} & \textbf{0.9} & 0.809          & nan            & -              & -                         & -                         & -                         \\
                                                                       & all      & 0.82           & \textbf{0.997} & \textbf{0.979} & \textbf{0.9} & 0.758          & 0.764          & 0.86           & 0.9                       & 0.794                     & 0.8                       \\ \hline
\multirow{4}{*}{\begin{tabular}[c]{@{}c@{}}nei-\\ p30\end{tabular}}    & victim   & 0.788          & 1              & 0.942          & nan          & 0.63           & 0.792          & -              & -                         & -                         & -                         \\
                                                                       & plane    & 0.838          & 1              & \textbf{0.988} & nan          & 0.839          & \textbf{0.846} & -              & -                         & -                         & -                         \\
                                                                       & boat     & 0.804          & \textbf{0.99}  & \textbf{0.976} & \textbf{0.9} & 0.804          & nan            & -              & -                         & -                         & -                         \\
                                                                       & all      & 0.81           & \textbf{0.997} & \textbf{0.969} & \textbf{0.9} & 0.757          & \textbf{0.819} & 0.85           & 0.9                       & 0.785                     & \textbf{0.865}            \\ \hline
\multirow{4}{*}{\begin{tabular}[c]{@{}c@{}}nei-\\ p5-50\end{tabular}}  & victim   & \textbf{0.812} & 1              & \textbf{0.961} & nan          & \textbf{0.667} & \textbf{0.815} & -              & -                         & -                         & -                         \\
                                                                       & plane    & \textbf{0.854} & 1              & \textbf{0.987} & nan          & \textbf{0.856} & 0.742          & -              & -                         & -                         & -                         \\
                                                                       & boat     & \textbf{0.815} & 0.989          & 0.963          & 0.8          & \textbf{0.815} & nan            & -              & -                         & -                         & -                         \\
                                                                       & all      & \textbf{0.827} & 0.996          & \textbf{0.97}  & 0.8          & \textbf{0.779} & \textbf{0.864} & 0.8            & 0.811                     & \textbf{0.825}            & 0.825                     \\ \hline
\end{tabular}
}
\caption{The FDD dataset is detected using the SSD300 algorithm. ``noisy" represents the original noise dataset, and dspnet, b2ub, and nei represent the target detection after denoising using the DSPNet, Blind2Unblind, and Neighbor2Neighbor single-frame denoising algorithms, respectively. Bold fonts indicate data that is higher than the corresponding ``noisy" dataset.}

\label{table25}
\end{table*}

\begin{table*}
\centering
\resizebox{\textwidth}{!}{
\begin{tabular}{c|ccccccccccc}
\hline
\multirow{2}{*}{Denoising}                                             & \multicolumn{11}{c}{YOLOX}                                                                                                                                                                                 \\ \cline{2-12} 
                                                                       & class    & mAP            & mAP@0.5 & mAP@0.75        & mAP\_s       & mAP\_m         & mAP\_l         & AR             & \multicolumn{1}{l}{AR\_s} & \multicolumn{1}{l}{AR\_m} & \multicolumn{1}{l}{AR\_l} \\ \hline
\multirow{4}{*}{noisy}                                                 & victim   & 0.83           & 1       & 0.964          & nan          & 0.627          & 0.838          & -              & -                         & -                         & -                         \\
                                                                       & plane    & 0.853          & 1       & 0.98           & nan          & 0.853          & 0.817          & -              & -                         & -                         & -                         \\
                                                                       & boat     & 0.825          & 0.99    & 0.979          & 0.8          & 0.825          & nan            & -              & -                         & -                         & -                         \\
                                                                       & all      & 0.836          & 0.997   & 0.974          & 0.8          & 0.768          & 0.827          & 0.863          & 0.8                       & 0.788                     & 0.855                     \\ \hline
\multirow{4}{*}{dspnet}                                                & victim   & 0.811          & 1       & \textbf{0.967} & nan          & \textbf{0.648} & 0.814          & -              & -                         & -                         & -                         \\
                                                                       & plane    & 0.852          & 1       & 0.979          & nan          & 0.852          & 0.783          & -              & -                         & -                         & -                         \\
                                                                       & boat     & 0.816          & 0.989   & \textbf{0.989} & \textbf{0.9} & 0.817          & nan            & -              & -                         & -                         & -                         \\
                                                                       & all      & 0.826          & 0.996   & \textbf{0.978} & \textbf{0.9} & \textbf{0.772} & 0.799          & 0.855          & \textbf{0.9}              & \textbf{0.796}            & 0.82                      \\ \hline
\multirow{4}{*}{\begin{tabular}[c]{@{}c@{}}b2ub-\\ g25\end{tabular}}   & victim   & 0.793          & 1       & 0.956          & nan          & 0.585          & 0.798          & -              & -                         & -                         & -                         \\
                                                                       & plane    & 0.832          & 1       & 0.978          & nan          & 0.832          & 0.725          & -              & -                         & -                         & -                         \\
                                                                       & boat     & 0.774          & 0.99    & 0.979          & 0.7          & 0.775          & nan            & -              & -                         & -                         & -                         \\
                                                                       & all      & 0.799          & 0.997   & 0.971          & 0.7          & 0.731          & 0.762          & 0.829          & 0.7                       & 0.752                     & 0.787                     \\ \hline
\multirow{4}{*}{\begin{tabular}[c]{@{}c@{}}b2ub-\\ g5-50\end{tabular}} & victim   & 0.818          & 1       & \textbf{0.968} & nan          & \textbf{0.715} & 0.82           & -              & -                         & -                         & -                         \\
                                                                       & plane    & \textbf{0.865} & 1       & \textbf{0.99}  & nan          & \textbf{0.866} & 0.723          & -              & -                         & -                         & -                         \\
                                                                       & boat     & 0.816          & 0.99    & \textbf{0.99}  & 0.7          & 0.816          & nan            & -              & -                         & -                         & -                         \\
                                                                       & all      & 0.833          & 0.997   & \textbf{0.983} & 0.7          & \textbf{0.799} & 0.771          & 0.861          & 0.7                       & \textbf{0.823}            & 0.824                     \\ \hline
\multirow{4}{*}{\begin{tabular}[c]{@{}c@{}}b2ub-\\ p30\end{tabular}}   & victim   & 0.816          & 1       & 0.954          & nan          & \textbf{0.628} & 0.821          & -              & -                         & -                         & -                         \\
                                                                       & plane    & \textbf{0.866} & 1       & 0.98           & nan          & \textbf{0.866} & 0.717          & -              & -                         & -                         & -                         \\
                                                                       & boat     & \textbf{0.837} & 0.99    & \textbf{0.99}  & \textbf{0.9} & \textbf{0.837} & nan            & -              & -                         & -                         & -                         \\
                                                                       & all      & \textbf{0.839} & 0.997   & \textbf{0.975} & \textbf{0.9} & \textbf{0.777} & 0.769          & \textbf{0.865} & \textbf{0.9}              & \textbf{0.799}            & 0.799                     \\ \hline
\multirow{4}{*}{\begin{tabular}[c]{@{}c@{}}b2ub-\\ p5-50\end{tabular}} & victim   & \textbf{0.843} & 1       & \textbf{0.975} & nan          & \textbf{0.699} & \textbf{0.846} & -              & -                         & -                         & -                         \\
                                                                       & plane    & \textbf{0.859} & 1       & 0.98           & nan          & \textbf{0.86}  & 0.8            & -              & -                         & -                         & -                         \\
                                                                       & boat     & \textbf{0.847} & 0.99    & \textbf{0.99}  & \textbf{0.9} & \textbf{0.847} & nan            & -              & -                         & -                         & -                         \\
                                                                       & all      & \textbf{0.85}  & 0.997   & \textbf{0.982} & \textbf{0.9} & \textbf{0.802} & 0.823          & \textbf{0.875} & \textbf{0.9}              & \textbf{0.82}             & 0.834                     \\ \hline
\multirow{4}{*}{\begin{tabular}[c]{@{}c@{}}nei-\\ g25\end{tabular}}    & victim   & 0.812          & 1       & 0.953          & nan          & \textbf{0.687} & 0.815          & -              & -                         & -                         & -                         \\
                                                                       & plane    & 0.833          & 1       & 0.98           & nan          & 0.834          & 0.717          & -              & -                         & -                         & -                         \\
                                                                       & boat     & 0.793          & 0.99    & \textbf{0.99}  & 0.8          & 0.793          & nan            & -              & -                         & -                         & -                         \\
                                                                       & all      & 0.812          & 0.997   & 0.974          & 0.8          & \textbf{0.771} & 0.766          & 0.843          & 0.8                       & 0.797                     & 0.795                     \\ \hline
\multirow{4}{*}{\begin{tabular}[c]{@{}c@{}}nei-\\ g5-50\end{tabular}}  & ship     & 0.819          & 1       & \textbf{0.974} & nan          & \textbf{0.713} & 0.822          & -              & -                         & -                         & -                         \\
                                                                       & aircraft & 0.85           & 1       & 0.979          & nan          & 0.852          & 0.723          & -              & -                         & -                         & -                         \\
                                                                       & human    & 0.824          & 0.99    & \textbf{0.98}  & 0.8          & 0.824          & nan            & -              & -                         & -                         & -                         \\
                                                                       & all      & 0.831          & 0.997   & \textbf{0.978} & 0.8          & 0.796          & 0.773          & 0.857          & 0.8                       & \textbf{0.819}            & 0.823                     \\ \hline
\multirow{4}{*}{\begin{tabular}[c]{@{}c@{}}nei-\\ p30\end{tabular}}    & victim   & 0.821          & 1       & \textbf{0.977} & nan          & \textbf{0.666} & 0.826          & -              & -                         & -                         & -                         \\
                                                                       & plane    & \textbf{0.863} & 1       & 0.98           & nan          & \textbf{0.864} & 0.7            & -              & -                         & -                         & -                         \\
                                                                       & boat     & \textbf{0.83}  & 0.989   & \textbf{0.989} & 0.8          & \textbf{0.831} & nan            & -              & -                         & -                         & -                         \\
                                                                       & all      & \textbf{0.838} & 0.996   & \textbf{0.982} & 0.8          & \textbf{0.787} & 0.763          & \textbf{0.866} & 0.8                       & \textbf{0.808}            & 0.825                     \\ \hline
\multirow{4}{*}{\begin{tabular}[c]{@{}c@{}}nei-\\ p5-50\end{tabular}}  & victim   & 0.823          & 1       & \textbf{0.965} & nan          & \textbf{0.718} & 0.825          & -              & -                         & -                         & -                         \\
                                                                       & plane    & \textbf{0.856} & 1       & 0.98           & nan          & \textbf{0.857} & 0.75           & -              & -                         & -                         & -                         \\
                                                                       & boat     & 0.821          & 0.989   & \textbf{0.989} & 0.8          & 0.822          & nan            & -              & -                         & -                         & -                         \\
                                                                       & all      & 0.833          & 0.996   & \textbf{0.978} & 0.8          & \textbf{0.799} & 0.788          & 0.859          & 0.8                       & \textbf{0.825}            & 0.824                     \\ \hline
\end{tabular}
}
\caption{The FDD dataset is detected using the YOLOX algorithm. ``noisy" represents the original noise dataset, and dspnet, b2ub, and nei represent the target detection after denoising using the DSPNet, Blind2Unblind, and Neighbor2Neighbor single-frame denoising algorithms, respectively. Bold fonts indicate data that is higher than the corresponding ``noisy" dataset.}

\label{table26}
\end{table*}

\begin{table*}
\centering
\resizebox{\textwidth}{!}{
\begin{tabular}{c|ccccccccccc}
\hline
\multirow{2}{*}{Denoising}                                              & \multicolumn{11}{c}{SSDMobileNetV2}                                                                                                                                                                                   \\ \cline{2-12} 
                                                                        & class        & mAP            & mAP@0.5        & mAP@0.75        & mAP\_s         & mAP\_m         & mAP\_l         & AR             & \multicolumn{1}{l}{AR\_s} & \multicolumn{1}{l}{AR\_m} & \multicolumn{1}{l}{AR\_l} \\ \hline
\multirow{11}{*}{noisy}                                                 & human body   & 0.475          & 0.972          & 0.345          & 0              & 0.471          & 0.552          & -              & -                         & -                         & -                         \\
                                                                        & ball         & 0.436          & 0.934          & 0.313          & nan            & 0.437          & 0.214          & -              & -                         & -                         & -                         \\
                                                                        & circle cage  & 0.415          & 0.898          & 0.354          & nan            & 0.412          & 0.474          & -              & -                         & -                         & -                         \\
                                                                        & square cage  & 0.428          & 0.922          & 0.317          & 0              & 0.431          & nan            & -              & -                         & -                         & -                         \\
                                                                        & tyre         & 0.373          & 0.863          & 0.257          & 0              & 0.348          & 0.564          & -              & -                         & -                         & -                         \\
                                                                        & cube         & 0.474          & 0.934          & 0.404          & 0.169          & 0.478          & 0.496          & -              & -                         & -                         & -                         \\
                                                                        & cylinder     & 0.313          & 0.816          & 0.12           & 0.583          & 0.309          & nan            & -              & -                         & -                         & -                         \\
                                                                        & metal bucket & 0.44           & 0.89           & 0.275          & nan            & 0.443          & 0.363          & -              & -                         & -                         & -                         \\
                                                                        & plane        & 0.562          & 0.974          & 0.586          & nan            & 0.535          & 0.605          & -              & -                         & -                         & -                         \\
                                                                        & rov          & 0.303          & 0.855          & 0.119          & 0              & 0.305          & nan            & -              & -                         & -                         & -                         \\
                                                                        & all          & 0.422          & 0.906          & 0.309          & 0.125          & 0.417          & 0.467          & 0.516          & 0.13                      & 0.512                     & 0.517                     \\ \hline
\multirow{11}{*}{dspnet}                                                & human body   & \textbf{0.484} & 0.964          & \textbf{0.411} & 0              & \textbf{0.481} & 0.564          & -              & -                         & -                         & -                         \\
                                                                        & ball         & \textbf{0.452} & \textbf{0.95}  & \textbf{0.321} & nan            & \textbf{0.451} & \textbf{0.327} & -              & -                         & -                         & -                         \\
                                                                        & circle cage  & \textbf{0.433} & 0.858          & 0.341          & nan            & \textbf{0.429} & \textbf{0.528} & -              & -                         & -                         & -                         \\
                                                                        & square cage  & \textbf{0.447} & \textbf{0.933} & \textbf{0.356} & 0              & \textbf{0.45}  & nan            & -              & -                         & -                         & -                         \\
                                                                        & tyre         & 0.372          & \textbf{0.896} & 0.234          & 0              & \textbf{0.355} & 0.507          & -              & -                         & -                         & -                         \\
                                                                        & cube         & 0.462          & 0.917          & 0.38           & \textbf{0.348} & 0.462          & \textbf{0.524} & -              & -                         & -                         & -                         \\
                                                                        & cylinder     & \textbf{0.329} & 0.795          & \textbf{0.219} & 0.288          & \textbf{0.331} & nan            & -              & -                         & -                         & -                         \\
                                                                        & metal bucket & \textbf{0.456} & \textbf{0.906} & \textbf{0.37}  & nan            & \textbf{0.455} & \textbf{0.451} & -              & -                         & -                         & -                         \\
                                                                        & plane        & 0.547          & \textbf{0.975} & 0.488          & nan            & 0.502          & \textbf{0.622} & -              & -                         & -                         & -                         \\
                                                                        & rov          & \textbf{0.32}  & 0.841          & 0.108          & \textbf{0.1}   & \textbf{0.323} & nan            & -              & -                         & -                         & -                         \\
                                                                        & all          & \textbf{0.43}  & 0.904          & \textbf{0.323} & 0.123          & \textbf{0.424} & \textbf{0.503} & \textbf{0.523} & \textbf{0.136}            & \textbf{0.518}            & \textbf{0.552}            \\ \hline
\multirow{11}{*}{\begin{tabular}[c]{@{}c@{}}b2ub-\\ g25\end{tabular}}   & human body   & 0.453          & 0.957          & \textbf{0.355} & 0              & 0.449          & 0.549          & -              & -                         & -                         & -                         \\
                                                                        & ball         & 0.425          & 0.931          & 0.243          & nan            & 0.425          & \textbf{0.277} & -              & -                         & -                         & -                         \\
                                                                        & circle cage  & 0.404          & 0.876          & 0.243          & nan            & 0.399          & \textbf{0.525} & -              & -                         & -                         & -                         \\
                                                                        & square cage  & 0.404          & 0.906          & 0.282          & 0              & 0.408          & nan            & -              & -                         & -                         & -                         \\
                                                                        & tyre         & 0.333          & 0.833          & 0.192          & 0              & 0.314          & 0.499          & -              & -                         & -                         & -                         \\
                                                                        & cube         & 0.439          & 0.931          & 0.281          & \textbf{0.283} & 0.439          & \textbf{0.543} & -              & -                         & -                         & -                         \\
                                                                        & cylinder     & 0.284          & 0.767          & \textbf{0.161} & 0.222          & 0.288          & nan            & -              & -                         & -                         & -                         \\
                                                                        & metal bucket & 0.391          & 0.832          & 0.248          & nan            & 0.393          & \textbf{0.376} & -              & -                         & -                         & -                         \\
                                                                        & plane        & 0.551          & \textbf{0.987} & 0.545          & nan            & 0.514          & \textbf{0.615} & -              & -                         & -                         & -                         \\
                                                                        & rov          & 0.296          & 0.832          & 0.098          & 0              & 0.297          & nan            & -              & -                         & -                         & -                         \\
                                                                        & all          & 0.398          & 0.885          & 0.265          & 0.084          & 0.393          & \textbf{0.483} & 0.494          & 0.091                     & 0.49                      & 0.514                     \\ \hline
\multirow{11}{*}{\begin{tabular}[c]{@{}c@{}}b2ub-\\ g5-50\end{tabular}} & human body   & 0.474          & 0.962          & \textbf{0.362} & 0              & 0.471          & 0.55           & -              & -                         & -                         & -                         \\
                                                                        & ball         & \textbf{0.446} & \textbf{0.938} & \textbf{0.325} & nan            & \textbf{0.446} & \textbf{0.252} & -              & -                         & -                         & -                         \\
                                                                        & circle cage  & 0.406          & 0.868          & 0.283          & nan            & 0.399          & \textbf{0.546} & -              & -                         & -                         & -                         \\
                                                                        & square cage  & \textbf{0.45}  & \textbf{0.959} & 0.276          & 0              & \textbf{0.453} & nan            & -              & -                         & -                         & -                         \\
                                                                        & tyre         & 0.363          & \textbf{0.893} & 0.217          & 0              & \textbf{0.352} & 0.488          & -              & -                         & -                         & -                         \\
                                                                        & cube         & 0.471          & \textbf{0.938} & 0.374          & \textbf{0.342} & 0.472          & 0.478          & -              & -                         & -                         & -                         \\
                                                                        & cylinder     & \textbf{0.317} & \textbf{0.819} & \textbf{0.182} & 0.533          & \textbf{0.315} & nan            & -              & -                         & -                         & -                         \\
                                                                        & metal bucket & \textbf{0.469} & \textbf{0.903} & \textbf{0.425} & nan            & \textbf{0.467} & \textbf{0.564} & -              & -                         & -                         & -                         \\
                                                                        & plane        & \textbf{0.57}  & \textbf{0.999} & 0.524          & nan            & \textbf{0.539} & \textbf{0.622} & -              & -                         & -                         & -                         \\
                                                                        & rov          & \textbf{0.323} & \textbf{0.918} & \textbf{0.121} & \textbf{0.2}   & \textbf{0.323} & nan            & -              & -                         & -                         & -                         \\
                                                                        & all          & \textbf{0.429} & \textbf{0.92}  & 0.309          & \textbf{0.179} & \textbf{0.424} & \textbf{0.5}   & \textbf{0.523} & \textbf{0.179}            & \textbf{0.516}            & \textbf{0.554}            \\ \hline

\end{tabular}
}
\caption{The UATD dataset is detected using the SSDMobileNetV2 algorithm (Part 1 of 3). ``noisy" represents the original noise dataset, and dspnet, b2ub, and nei represent the target detection after denoising using the DSPNet, Blind2Unblind, and Neighbor2Neighbor single-frame denoising algorithms, respectively. Bold fonts indicate data that is higher than the corresponding ``noisy" dataset.}
\label{table27}
\end{table*}

\begin{table*}
\centering
\resizebox{\textwidth}{!}{
\begin{tabular}{cccccccccccc}
\hline
\multicolumn{1}{c|}{\multirow{11}{*}{\begin{tabular}[c]{@{}c@{}}b2ub-\\ p30\end{tabular}}}   & human body   & 0.469          & \textbf{0.974} & 0.312          & 0              & 0.464          & \textbf{0.579} & -              & -                         & -                         & -                         \\
\multicolumn{1}{c|}{}                                                                         & ball         & 0.436          & \textbf{0.938} & 0.293          & nan            & 0.437          & \textbf{0.277} & -              & -                         & -                         & -                         \\
\multicolumn{1}{c|}{}                                                                         & circle cage  & \textbf{0.419} & 0.888          & 0.317          & nan            & \textbf{0.414} & \textbf{0.524} & -              & -                         & -                         & -                         \\
\multicolumn{1}{c|}{}                                                                         & square cage  & 0.426          & \textbf{0.951} & 0.248          & 0              & 0.428          & nan            & -              & -                         & -                         & -                         \\
\multicolumn{1}{c|}{}                                                                         & tyre         & 0.373          & 0.857          & 0.254          & 0              & \textbf{0.357} & 0.506          & -              & -                         & -                         & -                         \\
\multicolumn{1}{c|}{}                                                                         & cube         & 0.462          & 0.923          & 0.402          & \textbf{0.223} & 0.465          & \textbf{0.498} & -              & -                         & -                         & -                         \\
\multicolumn{1}{c|}{}                                                                         & cylinder     & 0.301          & 0.801          & \textbf{0.153} & 0.417          & 0.3            & nan            & -              & -                         & -                         & -                         \\
\multicolumn{1}{c|}{}                                                                         & metal bucket & 0.41           & \textbf{0.894} & \textbf{0.284} & nan            & 0.411          & \textbf{0.403} & -              & -                         & -                         & -                         \\
\multicolumn{1}{c|}{}                                                                         & plane        & 0.549          & \textbf{0.984} & 0.528          & nan            & 0.51           & \textbf{0.61}  & -              & -                         & -                         & -                         \\
\multicolumn{1}{c|}{}                                                                         & rov          & \textbf{0.304} & 0.832          & \textbf{0.135} & 0              & \textbf{0.306} & nan            & -              & -                         & -                         & -                         \\
\multicolumn{1}{c|}{}                                                                         & all          & 0.415          & 0.904          & 0.292          & 0.107          & 0.409          & \textbf{0.485} & 0.512          & 0.121                     & 0.506                     & \textbf{0.534}            \\ \hline
\multicolumn{1}{c|}{\multirow{11}{*}{\begin{tabular}[c]{@{}c@{}}b2ub-\\ p5-50\end{tabular}}} & human body   & 0.456          & 0.955          & 0.308          & 0              & 0.451          & \textbf{0.553} & -              & -                         & -                         & -                         \\
\multicolumn{1}{c|}{}                                                                         & ball         & 0.43           & \textbf{0.935} & 0.301          & nan            & 0.43           & \textbf{0.218} & -              & -                         & -                         & -                         \\
\multicolumn{1}{c|}{}                                                                         & circle cage  & \textbf{0.43}  & 0.882          & \textbf{0.363} & nan            & \textbf{0.429} & \textbf{0.475} & -              & -                         & -                         & -                         \\
\multicolumn{1}{c|}{}                                                                         & square cage  & 0.422          & \textbf{0.936} & 0.27           & 0              & 0.426          & nan            & -              & -                         & -                         & -                         \\
\multicolumn{1}{c|}{}                                                                         & tyre         & 0.372          & \textbf{0.889} & 0.252          & 0              & \textbf{0.35}  & 0.542          & -              & -                         & -                         & -                         \\
\multicolumn{1}{c|}{}                                                                         & cube         & 0.468          & \textbf{0.941} & 0.379          & \textbf{0.362} & 0.468          & 0.493          & -              & -                         & -                         & -                         \\
\multicolumn{1}{c|}{}                                                                         & cylinder     & 0.312          & 0.791          & \textbf{0.143} & 0.32           & \textbf{0.314} & nan            & -              & -                         & -                         & -                         \\
\multicolumn{1}{c|}{}                                                                         & metal bucket & 0.44           & \textbf{0.917} & 0.271          & nan            & \textbf{0.447} & 0.307          & -              & -                         & -                         & -                         \\
\multicolumn{1}{c|}{}                                                                         & plane        & \textbf{0.568} & \textbf{0.979} & \textbf{0.645} & nan            & \textbf{0.547} & 0.597          & -              & -                         & -                         & -                         \\
\multicolumn{1}{c|}{}                                                                         & rov          & 0.294          & 0.84           & 0.115          & 0              & 0.296          & nan            & -              & -                         & -                         & -                         \\
\multicolumn{1}{c|}{}                                                                         & all          & 0.419          & 0.906          & 0.305          & 0.114          & 0.416          & 0.455          & 0.513          & 0.121                     & 0.512                     & 0.5                       \\ \hline

\multicolumn{1}{c|}{\multirow{11}{*}{\begin{tabular}[c]{@{}c@{}}nei-\\ g25\end{tabular}}}   & human body           & 0.418                & 0.944                & 0.275                & 0                    & 0.408                & \textbf{0.592}       & -                    & -                    & -                    & -                    \\
\multicolumn{1}{c|}{}                                                                       & ball                 & 0.414                & 0.916                & 0.243                & nan                  & 0.414                & 0.202                & -                    & -                    & -                    & -                    \\
\multicolumn{1}{c|}{}                                                                       & circle cage          & 0.381                & 0.827                & 0.315                & nan                  & 0.38                 & 0.407                & -                    & -                    & -                    & -                    \\
\multicolumn{1}{c|}{}                                                                       & square cage          & 0.413                & \textbf{0.932}       & 0.247                & 0                    & 0.416                & nan                  & -                    & -                    & -                    & -                    \\
\multicolumn{1}{c|}{}                                                                       & tyre                 & 0.328                & 0.814                & 0.197                & 0                    & 0.307                & 0.474                & -                    & -                    & -                    & -                    \\
\multicolumn{1}{c|}{}                                                                       & cube                 & 0.439                & 0.919                & 0.31                 & \textbf{0.23}        & 0.44                 & \textbf{0.509}       & -                    & -                    & -                    & -                    \\
\multicolumn{1}{c|}{}                                                                       & cylinder             & \textbf{0.315}       & 0.807                & \textbf{0.13}        & 0.5                  & \textbf{0.312}       & nan                  & -                    & -                    & -                    & -                    \\
\multicolumn{1}{c|}{}                                                                       & metal bucket         & \textbf{0.444}       & \textbf{0.906}       & \textbf{0.32}        & nan                  & \textbf{0.449}       & \textbf{0.378}       & -                    & -                    & -                    & -                    \\
\multicolumn{1}{c|}{}                                                                       & plane                & \textbf{0.572}       & \textbf{0.979}       & 0.581                & nan                  & 0.534                & \textbf{0.633}       & -                    & -                    & -                    & -                    \\
\multicolumn{1}{c|}{}                                                                       & rov                  & 0.29                 & 0.853                & 0.099                & 0                    & 0.292                & nan                  & -                    & -                    & -                    & -                    \\
\multicolumn{1}{c|}{}                                                                       & all                  & 0.401                & 0.89                 & 0.272                & 0.122                & 0.395                & 0.456                & 0.496                & 0.129                & 0.491                & 0.491                \\ \hline
\multicolumn{1}{c|}{\multirow{11}{*}{\begin{tabular}[c]{@{}c@{}}nei-\\ g5-50\end{tabular}}} & human body           & 0.457                & 0.955                & 0.318                & 0                    & 0.454                & 0.521                & -                    & -                    & -                    & -                    \\
\multicolumn{1}{c|}{}                                                                       & ball                 & 0.417                & 0.926                & 0.253                & nan                  & 0.417                & \textbf{0.235}       & -                    & -                    & -                    & -                    \\
\multicolumn{1}{c|}{}                                                                       & circle cage          & \textbf{0.419}       & 0.895                & 0.295                & nan                  & \textbf{0.419}       & \textbf{0.48}        & \textbf{-}           & -                    & -                    & -                    \\
\multicolumn{1}{c|}{}                                                                       & square cage          & \textbf{0.434}       & \textbf{0.935}       & 0.267                & 0                    & \textbf{0.435}       & nan                  & -                    & -                    & -                    & -                    \\
\multicolumn{1}{c|}{}                                                                       & tyre                 & 0.318                & 0.844                & 0.122                & 0                    & 0.313                & 0.372                & -                    & -                    & -                    & -                    \\
\multicolumn{1}{c|}{}                                                                       & cube                 & 0.439                & 0.915                & 0.339                & \textbf{0.244}       & 0.441                & \textbf{0.508}       & -                    & -                    & -                    & -                    \\
\multicolumn{1}{c|}{}                                                                       & cylinder             & 0.306                & 0.776                & \textbf{0.174}       & 0.466                & 0.304                & nan                  & -                    & -                    & -                    & -                    \\
\multicolumn{1}{c|}{}                                                                       & metal bucket         & 0.418                & 0.855                & 0.265                & nan                  & 0.425                & 0.326                & -                    & -                    & -                    & -                    \\
\multicolumn{1}{c|}{}                                                                       & plane                & 0.537                & \textbf{0.997}       & 0.5                  & nan                  & 0.515                & 0.579                & -                    & -                    & -                    & -                    \\
\multicolumn{1}{c|}{}                                                                       & rov                  & 0.27                 & 0.8                  & 0.075                & 0.2                  & 0.271                & nan                  & -                    & -                    & -                    & -                    \\
\multicolumn{1}{c|}{}                                                                       & all                  & 0.401                & 0.89                 & 0.261                & \textbf{0.152}       & 0.399                & 0.432                & 0.495                & \textbf{0.16}        & 0.494                & 0.471                \\ \hline

\end{tabular}
}
\caption{The UATD dataset is detected using the SSDMobileNetV2 algorithm (Part 2 of 3). ``noisy" represents the original noise dataset, and dspnet, b2ub, and nei represent the target detection after denoising using the DSPNet, Blind2Unblind, and Neighbor2Neighbor single-frame denoising algorithms, respectively. Bold fonts indicate data that is higher than the corresponding ``noisy" dataset.}

\label{table28}
\end{table*}

\begin{table*}
\centering
\resizebox{\textwidth}{!}{
\begin{tabular}{cccccccccccc}
\hline
\multicolumn{1}{c|}{\multirow{11}{*}{\begin{tabular}[c]{@{}c@{}}nei-\\ p30\end{tabular}}}   & human body           & \textbf{0.506}       & \textbf{0.976}       & \textbf{0.42}        & 0                    & \textbf{0.507}       & 0.544                & -                    & -                    & -                    & -                    \\
\multicolumn{1}{c|}{}                                                                       & ball                 & \textbf{0.535}       & \textbf{0.968}       & \textbf{0.521}       & nan                  & \textbf{0.535}       & \textbf{0.303}       & -                    & -                    & -                    & -                    \\
\multicolumn{1}{c|}{}                                                                       & circle cage          & \textbf{0.514}       & \textbf{0.935}       & \textbf{0.498}       & nan                  & \textbf{0.518}       & \textbf{0.524}       & -                    & -                    & -                    & -                    \\
\multicolumn{1}{c|}{}                                                                       & square cage          & \textbf{0.488}       & \textbf{0.946}       & \textbf{0.385}       & 0                    & \textbf{0.491}       & nan                  & -                    & -                    & -                    & -                    \\
\multicolumn{1}{c|}{}                                                                       & tyre                 & \textbf{0.5}         & \textbf{0.92}        & \textbf{0.479}       & 0                    & \textbf{0.492}       & \textbf{0.586}       & -                    & -                    & -                    & -                    \\
\multicolumn{1}{c|}{}                                                                       & cube                 & \textbf{0.501}       & \textbf{0.942}       & \textbf{0.461}       & \textbf{0.307}       & \textbf{0.502}       & \textbf{0.549}       & -                    & -                    & -                    & -                    \\
\multicolumn{1}{c|}{}                                                                       & cylinder             & \textbf{0.352}       & 0.785                & \textbf{0.207}       & \textbf{0.655}       & \textbf{0.348}       & nan                  & -                    & -                    & -                    & -                    \\
\multicolumn{1}{c|}{}                                                                       & metal bucket         & \textbf{0.491}       & \textbf{0.915}       & \textbf{0.47}        & nan                  & \textbf{0.5}         & 0.295                & -                    & -                    & -                    & -                    \\
\multicolumn{1}{c|}{}                                                                       & plane                & \textbf{0.573}       & \textbf{0.997}       & \textbf{0.604}       & nan                  & \textbf{0.549}       & \textbf{0.614}       & -                    & -                    & -                    & -                    \\
\multicolumn{1}{c|}{}                                                                       & rov                  & \textbf{0.429}       & \textbf{0.977}       & \textbf{0.243}       & 0                    & \textbf{0.433}       & nan                  & -                    & -                    & -                    & -                    \\
\multicolumn{1}{c|}{}                                                                       & all                  & \textbf{0.489}       & \textbf{0.936}       & \textbf{0.429}       & \textbf{0.16}        & \textbf{0.488}       & \textbf{0.488}       & \textbf{0.572}       & \textbf{0.171}       & \textbf{0.571}       & \textbf{0.527}       \\ \hline
\multicolumn{1}{c|}{\multirow{11}{*}{\begin{tabular}[c]{@{}c@{}}nei-\\ p5-50\end{tabular}}} & human body           & \textbf{0.491}       & 0.961                & \textbf{0.413}       & 0                    & \textbf{0.487}       & \textbf{0.578}       & -                    & -                    & -                    & -                    \\
\multicolumn{1}{c|}{}                                                                       & ball                 & \textbf{0.461}       & \textbf{0.948}       & \textbf{0.35}        & nan                  & \textbf{0.46}        & \textbf{0.269}       & -                    & -                    & -                    & -                    \\
\multicolumn{1}{c|}{}                                                                       & circle cage          & \textbf{0.438}       & \textbf{0.904}       & \textbf{0.366}       & nan                  & \textbf{0.437}       & \textbf{0.479}       & -                    & -                    & -                    & -                    \\
\multicolumn{1}{c|}{}                                                                       & square cage          & \textbf{0.439}       & \textbf{0.96}        & 0.267                & 0                    & \textbf{0.441}       & nan                  & -                    & -                    & -                    & -                    \\
\multicolumn{1}{c|}{}                                                                       & tyre                 & \textbf{0.391}       & \textbf{0.906}       & 0.232                & 0                    & \textbf{0.378}       & 0.518                & -                    & -                    & -                    & -                    \\
\multicolumn{1}{c|}{}                                                                       & cube                 & 0.471                & \textbf{0.945}       & \textbf{0.407}       & \textbf{0.343}       & 0.47                 & \textbf{0.559}       & -                    & -                    & -                    & -                    \\
\multicolumn{1}{c|}{}                                                                       & cylinder             & 0.295                & 0.808                & 0.108                & 0.294                & 0.296                & nan                  & -                    & -                    & -                    & -                    \\
\multicolumn{1}{c|}{}                                                                       & metal bucket         & 0.438                & \textbf{0.905}       & \textbf{0.358}       & nan                  & 0.436                & \textbf{0.467}       & -                    & -                    & -                    & -                    \\
\multicolumn{1}{c|}{}                                                                       & plane                & 0.56                 & 0.974                & 0.576                & nan                  & 0.52                 & \textbf{0.62}        & -                    & -                    & -                    & -                    \\
\multicolumn{1}{c|}{}                                                                       & rov                  & \textbf{0.336}       & \textbf{0.911}       & \textbf{0.144}       & 0                    & \textbf{0.338}       & nan                  & -                    & -                    & -                    & -                    \\
\multicolumn{1}{c|}{}                                                                       & all                  & \textbf{0.432}       & \textbf{0.922}       & \textbf{0.322}       & 0.106                & \textbf{0.426}       & \textbf{0.499}       & \textbf{0.524}       & 0.123                & \textbf{0.519}       & \textbf{0.556}       \\ \hline
\multicolumn{1}{l}{}                                                                        & \multicolumn{1}{l}{} & \multicolumn{1}{l}{} & \multicolumn{1}{l}{} & \multicolumn{1}{l}{} & \multicolumn{1}{l}{} & \multicolumn{1}{l}{} & \multicolumn{1}{l}{} & \multicolumn{1}{l}{} & \multicolumn{1}{l}{} & \multicolumn{1}{l}{} & \multicolumn{1}{l}{} \\
\multicolumn{1}{l}{}                                                                        & \multicolumn{1}{l}{} & \multicolumn{1}{l}{} & \multicolumn{1}{l}{} & \multicolumn{1}{l}{} & \multicolumn{1}{l}{} & \multicolumn{1}{l}{} & \multicolumn{1}{l}{} & \multicolumn{1}{l}{} & \multicolumn{1}{l}{} & \multicolumn{1}{l}{} & \multicolumn{1}{l}{}
\end{tabular}
}
\caption{The UATD dataset is detected using the SSDMobileNetV2 algorithm (Part 3 of 3). ``noisy" represents the original noise dataset, and dspnet, b2ub, and nei represent the target detection after denoising using the DSPNet, Blind2Unblind, and Neighbor2Neighbor single-frame denoising algorithms, respectively. Bold fonts indicate data that is higher than the corresponding ``noisy" dataset.}

\label{table29}
\end{table*}

\begin{table*}
\centering

\resizebox{\textwidth}{!}{
\begin{tabular}{c|ccccccccccc}
\hline
\multirow{2}{*}{Denoising}                                              & \multicolumn{11}{c}{Faster R-CNN}                                                                                                                                                                                                                                                   \\ \cline{2-12} 
                                                                        & class        & mAP                                & mAP@0.5                            & mAP@0.75                            & mAP\_s         & mAP\_m         & mAP\_l         & AR             & \multicolumn{1}{l}{AR\_s} & \multicolumn{1}{l}{AR\_m} & \multicolumn{1}{l}{AR\_l} \\ \hline
\multirow{11}{*}{noisy}                                                 & human body   & 0.527                              & 0.977                              & 0.519                              & 0              & 0.531          & 0.547          & -              & -                         & -                         & -                         \\
                                                                        & ball         & 0.55                               & 0.959                              & 0.558                              & nan            & 0.551          & 0.252          & -              & -                         & -                         & -                         \\
                                                                        & circle cage  & 0.527                              & 0.904                              & 0.552                              & nan            & 0.52           & 0.613          & -              & -                         & -                         & -                         \\
                                                                        & square cage  & 0.528                              & 0.962                              & 0.502                              & 0              & 0.53           & nan            & -              & -                         & -                         & -                         \\
                                                                        & tyre         & 0.54                               & 0.957                              & 0.532                              & 0              & 0.53           & 0.638          & -              & -                         & -                         & -                         \\
                                                                        & cube         & 0.537                              & 0.955                              & 0.574                              & 0.343          & 0.538          & 0.578          & -              & -                         & -                         & -                         \\
                                                                        & cylinder     & 0.446                              & 0.895                              & 0.411                              & 0.734          & 0.437          & nan            & -              & -                         & -                         & -                         \\
                                                                        & metal bucket & 0.523                              & 0.932                              & 0.509                              & nan            & 0.533          & 0.37           & -              & -                         & -                         & -                         \\
                                                                        & plane        & 0.651                              & 0.99                               & 0.773                              & nan            & 0.606          & 0.702          & -              & -                         & -                         & -                         \\
                                                                        & rov          & 0.504                              & 0.968                              & 0.398                              & 0              & 0.507          & nan            & -              & -                         & -                         & -                         \\
                                                                        & all          & 0.533                              & 0.95                               & 0.533                              & 0.179          & 0.529          & 0.528          & 0.605          & 0.182                     & 0.603                     & 0.563                     \\ \hline
\multirow{11}{*}{dspnet}                                                & human body   & \multicolumn{1}{l}{\textbf{0.531}} & \multicolumn{1}{l}{\textbf{0.978}} & \multicolumn{1}{l}{0.501}          & 0              & \textbf{0.536} & \textbf{0.549} & -              & -                         & -                         & -                         \\
                                                                        & ball         & \multicolumn{1}{l}{0.549}          & \multicolumn{1}{l}{0.955}          & \multicolumn{1}{l}{0.555}          & nan            & 0.549          & \textbf{0.303} & -              & -                         & -                         & -                         \\
                                                                        & circle cage  & \multicolumn{1}{l}{\textbf{0.544}} & \multicolumn{1}{l}{\textbf{0.923}} & \multicolumn{1}{l}{\textbf{0.595}} & nan            & \textbf{0.544} & 0.574          & -              & -                         & -                         & -                         \\
                                                                        & square cage  & \multicolumn{1}{l}{0.51}           & \multicolumn{1}{l}{0.959}          & \multicolumn{1}{l}{0.446}          & 0              & 0.512          & nan            & -              & -                         & -                         & -                         \\
                                                                        & tyre         & \multicolumn{1}{l}{\textbf{0.551}} & \multicolumn{1}{l}{\textbf{0.96}}  & \multicolumn{1}{l}{\textbf{0.607}} & 0              & \textbf{0.544} & 0.632          & -              & -                         & -                         & -                         \\
                                                                        & cube         & \multicolumn{1}{l}{0.532}          & \multicolumn{1}{l}{0.954}          & \multicolumn{1}{l}{0.542}          & 0.343          & 0.533          & 0.561          & -              & -                         & -                         & -                         \\
                                                                        & cylinder     & \multicolumn{1}{l}{\textbf{0.453}} & \multicolumn{1}{l}{0.884}          & \multicolumn{1}{l}{0.387}          & 0.717          & \textbf{0.446} & nan            & -              & -                         & -                         & -                         \\
                                                                        & metal bucket & \multicolumn{1}{l}{0.515}          & \multicolumn{1}{l}{0.927}          & \multicolumn{1}{l}{0.5}            & nan            & 0.528          & 0.334          & -              & -                         & -                         & -                         \\
                                                                        & plane        & \multicolumn{1}{l}{0.638}          & \multicolumn{1}{l}{\textbf{0.999}} & \multicolumn{1}{l}{0.68}           & nan            & \textbf{0.612} & 0.674          & -              & -                         & -                         & -                         \\
                                                                        & rov          & \multicolumn{1}{l}{\textbf{0.53}}  & \multicolumn{1}{l}{\textbf{0.986}} & \multicolumn{1}{l}{\textbf{0.443}} & \textbf{0.1}   & \textbf{0.533} & nan            & -              & -                         & -                         & -                         \\
                                                                        & all          & \textbf{0.535}                     & \textbf{0.952}                     & 0.526                              & \textbf{0.193} & \textbf{0.534} & 0.518          & \textbf{0.606} & \textbf{0.196}            & \textbf{0.604}            & 0.554                     \\ \hline
\multirow{11}{*}{\begin{tabular}[c]{@{}c@{}}b2ub-\\ g25\end{tabular}}   & human body   & 0.484                              & 0.946                              & 0.421                              & 0              & 0.486          & 0.499          & -              & -                         & -                         & -                         \\
                                                                        & ball         & 0.523                              & 0.953                              & 0.503                              & nan            & 0.523          & 0.252          & -              & -                         & -                         & -                         \\
                                                                        & circle cage  & 0.505                              & 0.891                              & 0.466                              & nan            & 0.497          & 0.59           & -              & -                         & -                         & -                         \\
                                                                        & square cage  & 0.474                              & 0.929                              & 0.377                              & 0              & 0.477          & nan            & -              & -                         & -                         & -                         \\
                                                                        & tyre         & 0.466                              & 0.88                               & 0.429                              & 0              & 0.455          & 0.566          & -              & -                         & -                         & -                         \\
                                                                        & cube         & 0.512                              & 0.943                              & 0.482                              & 0.284          & 0.514          & \textbf{0.598} & -              & -                         & -                         & -                         \\
                                                                        & cylinder     & 0.42                               & 0.841                              & 0.338                              & 0.634          & 0.413          & nan            & -              & -                         & -                         & -                         \\
                                                                        & metal bucket & 0.471                              & 0.917                              & 0.432                              & nan            & \textbf{0.488} & 0.127          & -              & -                         & -                         & -                         \\
                                                                        & plane        & 0.643                              & \textbf{0.998}                     & 0.745                              & nan            & \textbf{0.606} & 0.691          & -              & -                         & -                         & -                         \\
                                                                        & rov          & \textbf{0.508}                     & 0.963                              & \textbf{0.438}                     & \textbf{0.3}   & 0.507          & nan            & -              & -                         & -                         & -                         \\
                                                                        & all          & 0.5                                & 0.926                              & 0.463                              & \textbf{0.203} & 0.497          & 0.475          & 0.577          & \textbf{0.236}            & 0.575                     & 0.506                     \\ \hline
\multirow{11}{*}{\begin{tabular}[c]{@{}c@{}}b2ub-\\ g5-50\end{tabular}} & human body   & \textbf{0.537}                     & 0.968                              & \textbf{0.525}                     & 0              & \textbf{0.535} & \textbf{0.591} & -              & -                         & -                         & -                         \\
                                                                        & ball         & \textbf{0.555}                     & \textbf{0.96}                      & \textbf{0.571}                     & nan            & \textbf{0.555} & \textbf{0.303} & -              & -                         & -                         & -                         \\
                                                                        & circle cage  & \textbf{0.535}                     & \textbf{0.923}                     & \textbf{0.583}                     & nan            & \textbf{0.537} & 0.53           & -              & -                         & -                         & -                         \\
                                                                        & square cage  & 0.517                              & 0.953                              & 0.473                              & 0              & 0.519          & nan            & -              & -                         & -                         & -                         \\
                                                                        & tyre         & \textbf{0.542}                     & 0.945                              & \textbf{0.57}                      & 0              & \textbf{0.535} & 0.628          & -              & -                         & -                         & -                         \\
                                                                        & cube         & \textbf{0.538}                     & \textbf{0.958}                     & 0.54                               & 0.322          & \textbf{0.542} & 0.547          & -              & -                         & -                         & -                         \\
                                                                        & cylinder     & \textbf{0.469}                     & 0.892                              & \textbf{0.426}                     & 0.734          & \textbf{0.463} & nan            & -              & -                         & -                         & -                         \\
                                                                        & metal bucket & \textbf{0.532}                     & \textbf{0.938}                     & \textbf{0.553}                     & nan            & \textbf{0.537} & \textbf{0.452} & -              & -                         & -                         & -                         \\
                                                                        & plane        & 0.635                              & 0.99                               & 0.741                              & nan            & 0.603          & 0.678          & -              & -                         & -                         & -                         \\
                                                                        & rov          & \textbf{0.52}                      & \textbf{0.987}                     & \textbf{0.429}                     & 0              & \textbf{0.525} & nan            & -              & -                         & -                         & -                         \\
                                                                        & all          & \textbf{0.538}                     & \textbf{0.951}                     & \textbf{0.541}                     & 0.176          & \textbf{0.535} & \textbf{0.533} & \textbf{0.607} & 0.176                     & \textbf{0.605}            & \textbf{0.569}            \\ \hline

\end{tabular}
}
\caption{The UATD dataset is detected using the Faster R-CNN algorithm (Part 1 of 3). ``noisy" represents the original noise dataset, and dspnet, b2ub, and nei represent the target detection after denoising using the DSPNet, Blind2Unblind, and Neighbor2Neighbor single-frame denoising algorithms, respectively. Bold fonts indicate data that is higher than the corresponding ``noisy" dataset.}
\label{table30}
\end{table*}

\begin{table*}
\centering
\resizebox{\textwidth}{!}{
\begin{tabular}{cccccccccccc}
\hline
\multicolumn{1}{c|}{\multirow{11}{*}{\begin{tabular}[c]{@{}c@{}}b2ub-\\ p30\end{tabular}}}  & human body   & \textbf{0.528}                     & 0.975                              & 0.512                              & 0              & 0.528          & \textbf{0.574} & -              & -                         & -                         & -                         \\
\multicolumn{1}{c|}{}                                                                         & ball         & \textbf{0.555}                     & 0.955                              & \textbf{0.593}                     & nan            & \textbf{0.555} & \textbf{0.303} & -              & -                         & -                         & -                         \\
\multicolumn{1}{c|}{}                                                                         & circle cage  & \textbf{0.53}                      & \textbf{0.914}                     & 0.537                              & nan            & \textbf{0.532} & 0.538          & -              & -                         & -                         & -                         \\
\multicolumn{1}{c|}{}                                                                         & square cage  & 0.519                              & 0.962                              & 0.501                              & 0              & 0.521          & nan            & -              & -                         & -                         & -                         \\
\multicolumn{1}{c|}{}                                                                         & tyre         & \textbf{0.541}                     & 0.928                              & \textbf{0.579}                     & 0              & \textbf{0.531} & 0.635          & -              & -                         & -                         & -                         \\
\multicolumn{1}{c|}{}                                                                         & cube         & \textbf{0.539}                     & 0.942                              & 0.566                              & 0.302          & \textbf{0.541} & \textbf{0.624} & -              & -                         & -                         & -                         \\
\multicolumn{1}{c|}{}                                                                         & cylinder     & 0.443                              & 0.885                              & 0.349                              & 0.7            & \textbf{0.44}  & nan            & -              & -                         & -                         & -                         \\
\multicolumn{1}{c|}{}                                                                         & metal bucket & 0.494                              & 0.929                              & 0.478                              & nan            & 0.498          & \textbf{0.427} & -              & -                         & -                         & -                         \\
\multicolumn{1}{c|}{}                                                                         & plane        & 0.629                              & \textbf{0.999}                     & 0.739                              & nan            & 0.605          & 0.662          & -              & -                         & -                         & -                         \\
\multicolumn{1}{c|}{}                                                                         & rov          & \textbf{0.535}                     & 0.967                              & \textbf{0.522}                     & \textbf{0.2}   & \textbf{0.538} & nan            & -              & -                         & -                         & -                         \\
\multicolumn{1}{c|}{}                                                                         & all          & 0.531                              & 0.946                              & \textbf{0.538}                     & \textbf{0.2}   & 0.529          & \textbf{0.538} & 0.605          & \textbf{0.206}            & 0.603                     & \textbf{0.574}            \\ \hline                                        
\multicolumn{1}{c|}{\multirow{11}{*}{\begin{tabular}[c]{@{}c@{}}nei-\\ g25\end{tabular}}} & human body   & \textbf{0.528}                     & 0.977                              & 0.479                              & 0              & 0.528          & \textbf{0.576} & -              & -                         & -                         & -                         \\
\multicolumn{1}{c|}{}                                                                        & ball         & \textbf{0.554}                     & \textbf{0.967}                     & \textbf{0.593}                     & nan            & \textbf{0.555} & 0.252          & -              & -                         & -                         & -                         \\
\multicolumn{1}{c|}{}                                                                        & circle cage  & \textbf{0.548}                     & \textbf{0.932}                     & \textbf{0.598}                     & nan            & \textbf{0.548} & 0.55           & -              & -                         & -                         & -                         \\
\multicolumn{1}{c|}{}                                                                        & square cage  & 0.512                              & 0.952                              & 0.49                               & 0              & 0.514          & nan            & -              & -                         & -                         & -                         \\
\multicolumn{1}{c|}{}                                                                        & tyre         & \textbf{0.543}                     & 0.949                              & \textbf{0.589}                     & 0              & \textbf{0.54}  & 0.593          & -              & -                         & -                         & -                         \\
\multicolumn{1}{c|}{}                                                                        & cube         & \textbf{0.542}                     & \textbf{0.957}                     & 0.551                              & \textbf{0.362} & \textbf{0.542} & \textbf{0.598} & -              & -                         & -                         & -                         \\
\multicolumn{1}{c|}{}                                                                        & cylinder     & \textbf{0.457}                     & \textbf{0.897}                     & \textbf{0.438}                     & 0.701          & \textbf{0.451} & nan            & -              & -                         & -                         & -                         \\
\multicolumn{1}{c|}{}                                                                        & metal bucket & 0.492                              & 0.905                              & 0.454                              & nan            & 0.5            & 0.33           & -              & -                         & -                         & -                         \\
\multicolumn{1}{c|}{}                                                                        & plane        & 0.634                              & 0.989                              & 0.741                              & nan            & 0.605          & 0.678          & -              & -                         & -                         & -                         \\
\multicolumn{1}{c|}{}                                                                        & rov          & \textbf{0.514}                     & 0.966                              & \textbf{0.408}                     & 0              & \textbf{0.516} & nan            & -              & -                         & -                         & -                         \\
\multicolumn{1}{c|}{}                                                                        & all          & 0.532                              & 0.949                              & \textbf{0.534}                     & 0.177          & \textbf{0.53}  & 0.511          & \textbf{0.606} & 0.177                     & \textbf{0.605}            & 0.553                     \\ \hline
\multicolumn{1}{c|}{\multirow{11}{*}{\begin{tabular}[c]{@{}c@{}}nei-\\ g25\end{tabular}}}   & human body           & \textbf{0.53}        & \textbf{0.982}       & \textbf{0.52}        & 0                    & 0.526                & \textbf{0.583}       & -                    & -                    & -                    & -                    \\
\multicolumn{1}{c|}{}                                                                       & ball                 & 0.541                & 0.957                & 0.558                & nan                  & 0.541                & 0.252                & -                    & -                    & -                    & -                    \\
\multicolumn{1}{c|}{}                                                                       & circle cage          & 0.517                & \textbf{0.913}       & 0.499                & nan                  & 0.519                & 0.525                & -                    & -                    & -                    & -                    \\
\multicolumn{1}{c|}{}                                                                       & square cage          & 0.505                & 0.955                & 0.468                & 0                    & 0.508                & nan                  & -                    & -                    & -                    & -                    \\
\multicolumn{1}{c|}{}                                                                       & tyre                 & 0.507                & 0.926                & 0.508                & 0                    & 0.495                & 0.61                 & -                    & -                    & -                    & -                    \\
\multicolumn{1}{c|}{}                                                                       & cube                 & 0.527                & 0.945                & 0.536                & 0.263                & 0.532                & 0.537                & -                    & -                    & -                    & -                    \\
\multicolumn{1}{c|}{}                                                                       & cylinder             & \textbf{0.45}        & 0.865                & 0.407                & 0.672                & \textbf{0.445}       & nan                  & -                    & -                    & -                    & -                    \\
\multicolumn{1}{c|}{}                                                                       & metal bucket         & 0.513                & 0.915                & \textbf{0.538}       & nan                  & 0.522                & 0.326                & -                    & -                    & -                    & -                    \\
\multicolumn{1}{c|}{}                                                                       & plane                & \textbf{0.654}       & 0.99                 & \textbf{0.777}       & nan                  & \textbf{0.617}       & 0.7                  & -                    & -                    & -                    & -                    \\
\multicolumn{1}{c|}{}                                                                       & rov                  & \textbf{0.512}       & \textbf{0.976}       & \textbf{0.442}       & \textbf{0.3}         & \textbf{0.514}       & nan                  & -                    & -                    & -                    & -                    \\
\multicolumn{1}{c|}{}                                                                       & all                  & 0.526                & 0.943                & 0.525                & \textbf{0.206}       & 0.522                & 0.505                & 0.595                & \textbf{0.21}        & 0.593                & 0.535                \\ \hline
\multicolumn{1}{c|}{\multirow{11}{*}{\begin{tabular}[c]{@{}c@{}}nei-\\ g5-50\end{tabular}}} & human body           & 0.524                & \textbf{0.979}       & 0.482                & 0                    & 0.525                & \textbf{0.56}        & -                    & -                    & -                    & -                    \\
\multicolumn{1}{c|}{}                                                                       & ball                 & \textbf{0.556}       & \textbf{0.966}       & \textbf{0.588}       & nan                  & \textbf{0.556}       & \textbf{0.303}       & -                    & -                    & -                    & -                    \\
\multicolumn{1}{c|}{}                                                                       & circle cage          & 0.526                & \textbf{0.931}       & 0.535                & nan                  & \textbf{0.526}       & 0.551                & \textbf{-}           & -                    & -                    & -                    \\
\multicolumn{1}{c|}{}                                                                       & square cage          & 0.506                & 0.954                & 0.437                & 0                    & 0.51                 & nan                  & -                    & -                    & -                    & -                    \\
\multicolumn{1}{c|}{}                                                                       & tyre                 & \textbf{0.548}       & 0.948                & \textbf{0.585}       & 0                    & \textbf{0.548}       & 0.59                 & -                    & -                    & -                    & -                    \\
\multicolumn{1}{c|}{}                                                                       & cube                 & 0.534                & 0.95                 & 0.537                & 0.343                & 0.535                & \textbf{0.581}       & -                    & -                    & -                    & -                    \\
\multicolumn{1}{c|}{}                                                                       & cylinder             & \textbf{0.454}       & 0.895                & \textbf{0.423}       & 0.723                & \textbf{0.449}       & nan                  & -                    & -                    & -                    & -                    \\
\multicolumn{1}{c|}{}                                                                       & metal bucket         & 0.52                 & 0.928                & 0.481                & nan                  & 0.524                & \textbf{0.478}       & -                    & -                    & -                    & -                    \\
\multicolumn{1}{c|}{}                                                                       & plane                & 0.649                & 0.98                 & 0.709                & nan                  & \textbf{0.608}       & 0.699                & -                    & -                    & -                    & -                    \\
\multicolumn{1}{c|}{}                                                                       & rov                  & \textbf{0.521}       & \textbf{0.983}       & \textbf{0.446}       & \textbf{0.3}         & \textbf{0.521}       & nan                  & -                    & -                    & -                    & -                    \\
\multicolumn{1}{c|}{}                                                                       & all                  & \textbf{0.534}       & \textbf{0.951}       & 0.522                & \textbf{0.228}       & \textbf{0.53}        & \textbf{0.537}       & \textbf{0.606}       & \textbf{0.246}       & 0.603                & \textbf{0.574}       \\ \hline

\end{tabular}
}
\caption{The UATD dataset is detected using the Faster R-CNN algorithm (Part 2 of 3). ``noisy" represents the original noise dataset, and dspnet, b2ub, and nei represent the target detection after denoising using the DSPNet, Blind2Unblind, and Neighbor2Neighbor single-frame denoising algorithms, respectively. Bold fonts indicate data that is higher than the corresponding ``noisy" dataset.}
\label{table31}
\end{table*}

\begin{table*}
\centering
\resizebox{\textwidth}{!}{

\begin{tabular}{cccccccccccc}
\hline
\multicolumn{1}{c|}{\multirow{11}{*}{\begin{tabular}[c]{@{}c@{}}nei-\\ p30\end{tabular}}}   & human body           & \textbf{0.569}       & 0.977                & \textbf{0.623}       & 0                    & \textbf{0.568}       & \textbf{0.618}       & -                    & -                    & -                    & -                    \\
\multicolumn{1}{c|}{}                                                                       & ball                 & \textbf{0.587}       & \textbf{0.973}       & \textbf{0.663}       & nan                  & \textbf{0.587}       & \textbf{0.353}       & -                    & -                    & -                    & -                    \\
\multicolumn{1}{c|}{}                                                                       & circle cage          & \textbf{0.532}       & \textbf{0.912}       & \textbf{0.574}       & nan                  & \textbf{0.531}       & 0.57                 & -                    & -                    & -                    & -                    \\
\multicolumn{1}{c|}{}                                                                       & square cage          & 0.513                & 0.953                & 0.496                & 0                    & 0.515                & nan                  & -                    & -                    & -                    & -                    \\
\multicolumn{1}{c|}{}                                                                       & tyre                 & \textbf{0.553}       & 0.946                & \textbf{0.601}       & 0                    & \textbf{0.548}       & 0.632                & -                    & -                    & -                    & -                    \\
\multicolumn{1}{c|}{}                                                                       & cube                 & \textbf{0.563}       & \textbf{0.96}        & \textbf{0.607}       & \textbf{0.423}       & \textbf{0.563}       & \textbf{0.642}       & -                    & -                    & -                    & -                    \\
\multicolumn{1}{c|}{}                                                                       & cylinder             & \textbf{0.476}       & \textbf{0.941}       & 0.366                & 0.634                & \textbf{0.472}       & nan                  & -                    & -                    & -                    & -                    \\
\multicolumn{1}{c|}{}                                                                       & metal bucket         & 0.509                & 0.931                & \textbf{0.516}       & nan                  & 0.522                & 0.251                & -                    & -                    & -                    & -                    \\
\multicolumn{1}{c|}{}                                                                       & plane                & \textbf{0.658}       & \textbf{1}           & 0.742                & nan                  & \textbf{0.631}       & 0.69                 & -                    & -                    & -                    & -                    \\
\multicolumn{1}{c|}{}                                                                       & rov                  & \textbf{0.525}       & \textbf{0.999}       & \textbf{0.456}       & \textbf{0.2}         & \textbf{0.527}       & nan                  & -                    & -                    & -                    & -                    \\
\multicolumn{1}{c|}{}                                                                       & all                  & \textbf{0.549}       & \textbf{0.959}       & \textbf{0.959}       & \textbf{0.209}       & \textbf{0.546}       & \textbf{0.537}       & \textbf{0.61}        & \textbf{0.209}       & \textbf{0.609}       & \textbf{0.565}       \\ \hline
\multicolumn{1}{c|}{\multirow{11}{*}{\begin{tabular}[c]{@{}c@{}}nei-\\ p5-50\end{tabular}}} & human body           & \textbf{0.537}       & \textbf{0.978}       & \textbf{0.538}       & 0                    & \textbf{0.536}       & \textbf{0.581}       & -                    & -                    & -                    & -                    \\
\multicolumn{1}{c|}{}                                                                       & ball                 & \textbf{0.554}       & \textbf{0.969}       & \textbf{0.577}       & nan                  & \textbf{0.553}       & \textbf{0.303}       & -                    & -                    & -                    & -                    \\
\multicolumn{1}{c|}{}                                                                       & circle cage          & \textbf{0.536}       & \textbf{0.93}        & 0.548                & nan                  & \textbf{0.538}       & 0.535                & -                    & -                    & -                    & -                    \\
\multicolumn{1}{c|}{}                                                                       & square cage          & 0.517                & 0.95                 & 0.467                & 0                    & 0.519                & nan                  & -                    & -                    & -                    & -                    \\
\multicolumn{1}{c|}{}                                                                       & tyre                 & \textbf{0.551}       & 0.928                & \textbf{0.615}       & 0                    & \textbf{0.543}       & 0.631                & -                    & -                    & -                    & -                    \\
\multicolumn{1}{c|}{}                                                                       & cube                 & \textbf{0.541}       & \textbf{0.957}       & 0.544                & 0.329                & \textbf{0.542}       & 0.545                & -                    & -                    & -                    & -                    \\
\multicolumn{1}{c|}{}                                                                       & cylinder             & \textbf{0.457}       & 0.873                & \textbf{0.418}       & 0.634                & \textbf{0.453}       & nan                  & -                    & -                    & -                    & -                    \\
\multicolumn{1}{c|}{}                                                                       & metal bucket         & 0.511                & \textbf{0.933}       & 0.451                & nan                  & 0.523                & 0.326                & -                    & -                    & -                    & -                    \\
\multicolumn{1}{c|}{}                                                                       & plane                & \textbf{0.653}       & \textbf{0.999}       & 0.708                & nan                  & \textbf{0.616}       & 0.701                & -                    & -                    & -                    & -                    \\
\multicolumn{1}{c|}{}                                                                       & rov                  & \textbf{0.512}       & \textbf{0.976}       & \textbf{0.44}        & \textbf{0.1}         & \textbf{0.513}       & nan                  & -                    & -                    & -                    & -                    \\
\multicolumn{1}{c|}{}                                                                       & all                  & \textbf{0.537}       & 0.949                & 0.531                & 0.177                & \textbf{0.534}       & 0.517                & \textbf{0.608}       & \textbf{0.184}       & \textbf{0.607}       & 0.549                \\ \hline
\multicolumn{1}{l}{}                                                                        & \multicolumn{1}{l}{} & \multicolumn{1}{l}{} & \multicolumn{1}{l}{} & \multicolumn{1}{l}{} & \multicolumn{1}{l}{} & \multicolumn{1}{l}{} & \multicolumn{1}{l}{} & \multicolumn{1}{l}{} & \multicolumn{1}{l}{} & \multicolumn{1}{l}{} & \multicolumn{1}{l}{} \\
\multicolumn{1}{l}{}                                                                        & \multicolumn{1}{l}{} & \multicolumn{1}{l}{} & \multicolumn{1}{l}{} & \multicolumn{1}{l}{} & \multicolumn{1}{l}{} & \multicolumn{1}{l}{} & \multicolumn{1}{l}{} & \multicolumn{1}{l}{} & \multicolumn{1}{l}{} & \multicolumn{1}{l}{} & \multicolumn{1}{l}{}
\end{tabular}
}
\caption{The UATD dataset is detected using the Faster R-CNN algorithm (Part 3 of 3). ``noisy" represents the original noise dataset, and dspnet, b2ub, and nei represent the target detection after denoising using the DSPNet, Blind2Unblind, and Neighbor2Neighbor single-frame denoising algorithms, respectively. Bold fonts indicate data that is higher than the corresponding ``noisy" dataset.}
\label{table32}
\end{table*}

\begin{table*}
\centering
\resizebox{\textwidth}{!}{

\begin{tabular}{c|ccccccccccc}
\hline
\multirow{2}{*}{Denoising}                                              & \multicolumn{11}{c}{SSD300}                                                                                                                                                                                           \\ \cline{2-12} 
                                                                        & class        & mAP            & mAP@0.5        & mAP@0.75        & mAP\_s       & mAP\_m         & mAP\_l         & AR             & \multicolumn{1}{l}{AR\_s} & \multicolumn{1}{l}{AR\_m} & \multicolumn{1}{l}{AR\_l} \\ \hline
\multirow{11}{*}{noisy}                                                 & human body   & 0.481          & 0.966          & 0.402          & 0            & 0.482          & 0.485          & -              & -                         & -                         & -                         \\
                                                                        & ball         & 0.517          & 0.963          & 0.497          & nan          & 0.516          & 0.349          & -              & -                         & -                         & -                         \\
                                                                        & circle cage  & 0.496          & 0.92           & 0.504          & nan          & 0.486          & 0.582          & -              & -                         & -                         & -                         \\
                                                                        & square cage  & 0.473          & 0.942          & 0.361          & 0            & 0.477          & nan            & -              & -                         & -                         & -                         \\
                                                                        & tyre         & 0.463          & 0.908          & 0.425          & 0            & 0.451          & 0.574          & -              & -                         & -                         & -                         \\
                                                                        & cube         & 0.499          & 0.939          & 0.448          & 0.363        & 0.499          & 0.587          & -              & -                         & -                         & -                         \\
                                                                        & cylinder     & 0.355          & 0.823          & 0.19           & 0.666        & 0.352          & nan            & -              & -                         & -                         & -                         \\
                                                                        & metal bucket & 0.484          & 0.926          & 0.43           & nan          & 0.492          & 0.337          & -              & -                         & -                         & -                         \\
                                                                        & plane        & 0.582          & 0.996          & 0.643          & nan          & 0.542          & 0.644          & -              & -                         & -                         & -                         \\
                                                                        & rov          & 0.442          & 0.976          & 0.321          & 0            & 0.447          & nan            & -              & -                         & -                         & -                         \\
                                                                        & all          & 0.479          & 0.936          & 0.422          & 0.172        & 0.474          & 0.508          & 0.561          & 0.171                     & 0.558                     & 0.586                     \\ \hline
\multirow{11}{*}{dspnet}                                                & human body   & 0.476          & 0.957          & 0.389          & 0            & 0.478          & \textbf{0.503} & -              & -                         & -                         & -                         \\
                                                                        & ball         & 0.512          & 0.962          & 0.443          & nan          & 0.511          & \textbf{0.435} & -              & -                         & -                         & -                         \\
                                                                        & circle cage  & 0.494          & \textbf{0.931} & 0.434          & nan          & 0.486          & 0.578          & -              & -                         & -                         & -                         \\
                                                                        & square cage  & 0.446          & 0.912          & 0.273          & 0            & 0.45           & nan            & -              & -                         & -                         & -                         \\
                                                                        & tyre         & 0.457          & 0.864          & 0.424          & 0            & 0.443          & \textbf{0.582} & -              & -                         & -                         & -                         \\
                                                                        & cube         & 0.487          & 0.928          & \textbf{0.458} & 0.343        & 0.486          & 0.586          & -              & -                         & -                         & -                         \\
                                                                        & cylinder     & \textbf{0.379} & \textbf{0.838} & \textbf{0.24}  & 0.623        & \textbf{0.375} & nan            & -              & -                         & -                         & -                         \\
                                                                        & metal bucket & \textbf{0.5}   & 0.924          & \textbf{0.494} & nan          & \textbf{0.51}  & 0.327          & -              & -                         & -                         & -                         \\
                                                                        & plane        & 0.578          & 0.996          & 0.557          & nan          & \textbf{0.55}  & 0.617          & -              & -                         & -                         & -                         \\
                                                                        & rov          & 0.426          & 0.946          & 0.257          & 0            & 0.437          & nan            & -              & -                         & -                         & -                         \\
                                                                        & all          & 0.476          & 0.926          & 0.397          & 0.161        & 0.473          & \textbf{0.518} & 0.558          & \textbf{0.182}            & 0.555                     & 0.569                     \\ \hline
\multirow{11}{*}{\begin{tabular}[c]{@{}c@{}}b2ub-\\ g25\end{tabular}}   & human body   & 0.454          & \textbf{0.974} & 0.312          & 0            & 0.455          & \textbf{0.498} & -              & -                         & -                         & -                         \\
                                                                        & ball         & 0.506          & 0.956          & 0.425          & nan          & 0.507          & \textbf{0.384} & -              & -                         & -                         & -                         \\
                                                                        & circle cage  & 0.477          & 0.92           & 0.468          & nan          & 0.469          & 0.546          & -              & -                         & -                         & -                         \\
                                                                        & square cage  & 0.458          & 0.929          & 0.351          & 0            & 0.461          & nan            & -              & -                         & -                         & -                         \\
                                                                        & tyre         & 0.445          & 0.89           & 0.412          & 0            & 0.43           & \textbf{0.58}  & -              & -                         & -                         & -                         \\
                                                                        & cube         & 0.486          & 0.939          & 0.412          & 0.261        & 0.484          & \textbf{0.596} & -              & -                         & -                         & -                         \\
                                                                        & cylinder     & \textbf{0.357} & \textbf{0.847} & \textbf{0.194} & 0.434        & 0.356          & nan            & -              & -                         & -                         & -                         \\
                                                                        & metal bucket & 0.458          & 0.922          & 0.316          & nan          & 0.47           & 0.201          & -              & -                         & -                         & -                         \\
                                                                        & plane        & 0.577          & 0.977          & 0.638          & nan          & \textbf{0.559} & 0.606          & -              & -                         & -                         & -                         \\
                                                                        & rov          & 0.422          & 0.954          & 0.219          & 0            & 0.426          & nan            & -              & -                         & -                         & -                         \\
                                                                        & all          & 0.464          & 0.931          & 0.375          & 0.116        & 0.462          & 0.487          & 0.546          & 0.116                     & 0.545                     & 0.538                     \\ \hline
\multirow{11}{*}{\begin{tabular}[c]{@{}c@{}}b2ub-\\ g5-50\end{tabular}} & human body   & \textbf{0.508} & \textbf{0.971} & 0.438          & 0            & \textbf{0.51}  & \textbf{0.524} & -              & -                         & -                         & -                         \\
                                                                        & ball         & \textbf{0.534} & \textbf{0.967} & \textbf{0.519} & nan          & \textbf{0.534} & \textbf{0.372} & -              & -                         & -                         & -                         \\
                                                                        & circle cage  & 0.495          & \textbf{0.934} & 0.408          & nan          & \textbf{0.491} & 0.543          & -              & -                         & -                         & -                         \\
                                                                        & square cage  & \textbf{0.49}  & 0.942          & \textbf{0.411} & 0            & \textbf{0.494} & nan            & -              & -                         & -                         & -                         \\
                                                                        & tyre         & \textbf{0.48}  & \textbf{0.913} & \textbf{0.464} & 0            & \textbf{0.469} & \textbf{0.59}  & -              & -                         & -                         & -                         \\
                                                                        & cube         & \textbf{0.504} & \textbf{0.942} & \textbf{0.481} & 0.322        & \textbf{0.505} & \textbf{0.588} & -              & -                         & -                         & -                         \\
                                                                        & cylinder     & \textbf{0.364} & \textbf{0.839} & \textbf{0.201} & 0.655        & \textbf{0.359} & nan            & -              & -                         & -                         & -                         \\
                                                                        & metal bucket & \textbf{0.492} & \textbf{0.928} & \textbf{0.487} & nan          & \textbf{0.503} & 0.271          & -              & -                         & -                         & -                         \\
                                                                        & plane        & 0.58           & 0.979          & 0.612          & nan          & \textbf{0.549} & 0.626          & -              & -                         & -                         & -                         \\
                                                                        & rov          & 0.428          & 0.958          & 0.262          & 0            & 0.435          & nan            & -              & -                         & -                         & -                         \\
                                                                        & all          & \textbf{0.488} & \textbf{0.937} & \textbf{0.428} & 0.163        & \textbf{0.485} & 0.502          & \textbf{0.57}  & 0.168                     & \textbf{0.568}            & \textbf{0.589}            \\ \hline
\multirow{11}{*}{\begin{tabular}[c]{@{}c@{}}b2ub-\\ p30\end{tabular}}   & human body   & \textbf{0.501} & \textbf{0.97}  & \textbf{0.456} & 0            & \textbf{0.504} & \textbf{0.522} & -              & -                         & -                         & -                         \\
                                                                        & ball         & \textbf{0.531} & \textbf{0.966} & \textbf{0.505} & nan          & \textbf{0.531} & \textbf{0.382} & -              & -                         & -                         & -                         \\
                                                                        & circle cage  & \textbf{0.519} & \textbf{0.927} & \textbf{0.564} & nan          & \textbf{0.514} & 0.57           & -              & -                         & -                         & -                         \\
                                                                        & square cage  & \textbf{0.495} & \textbf{0.948} & \textbf{0.452} & 0            & \textbf{0.497} & nan            & -              & -                         & -                         & -                         \\
                                                                        & tyre         & \textbf{0.476} & 0.9            & \textbf{0.46}  & 0            & \textbf{0.466} & 0.572          & -              & -                         & -                         & -                         \\
                                                                        & cube         & \textbf{0.505} & 0.936          & \textbf{0.484} & 0.236        & \textbf{0.506} & \textbf{0.599} & -              & -                         & -                         & -                         \\
                                                                        & cylinder     & \textbf{0.375} & 0.803          & \textbf{0.253} & 0.578        & \textbf{0.371} & nan            & -              & -                         & -                         & -                         \\
                                                                        & metal bucket & \textbf{0.501} & 0.916          & \textbf{0.492} & nan          & \textbf{0.508} & \textbf{0.35}  & -              & -                         & -                         & -                         \\
                                                                        & plane        & \textbf{0.585} & 0.976          & 0.62           & nan          & \textbf{0.547} & \textbf{0.647} & -              & -                         & -                         & -                         \\
                                                                        & rov          & 0.43           & 0.96           & 0.251          & 0            & 0.434          & nan            & -              & -                         & -                         & -                         \\
                                                                        & all          & \textbf{0.492} & 0.93           & \textbf{0.454} & 0.136        & \textbf{0.488} & \textbf{0.521} & \textbf{0.575} & 0.156                     & \textbf{0.573}            & \textbf{0.596}            \\ \hline
\end{tabular}
}
\caption{The UATD dataset is detected using the SSD300 algorithm (Part 1 of 3). ``noisy" represents the original noise dataset, and dspnet, b2ub, and nei represent the target detection after denoising using the DSPNet, Blind2Unblind, and Neighbor2Neighbor single-frame denoising algorithms, respectively. Bold fonts indicate data that is higher than the corresponding ``noisy" dataset.}
\label{table33}
\end{table*}

\begin{table*}
\centering
\resizebox{\textwidth}{!}{

\begin{tabular}{cccccccccccc}
\hline
\multicolumn{1}{c|}{\multirow{11}{*}{\begin{tabular}[c]{@{}c@{}}b2ub-\\ p30\end{tabular}}}   & human body   & \textbf{0.501} & \textbf{0.97}  & \textbf{0.456} & 0            & \textbf{0.504} & \textbf{0.522} & -              & -                         & -                         & -                         \\
\multicolumn{1}{c|}{}                                                                         & ball         & \textbf{0.531} & \textbf{0.966} & \textbf{0.505} & nan          & \textbf{0.531} & \textbf{0.382} & -              & -                         & -                         & -                         \\
\multicolumn{1}{c|}{}                                                                         & circle cage  & \textbf{0.519} & \textbf{0.927} & \textbf{0.564} & nan          & \textbf{0.514} & 0.57           & -              & -                         & -                         & -                         \\
\multicolumn{1}{c|}{}                                                                         & square cage  & \textbf{0.495} & \textbf{0.948} & \textbf{0.452} & 0            & \textbf{0.497} & nan            & -              & -                         & -                         & -                         \\
\multicolumn{1}{c|}{}                                                                         & tyre         & \textbf{0.476} & 0.9            & \textbf{0.46}  & 0            & \textbf{0.466} & 0.572          & -              & -                         & -                         & -                         \\
\multicolumn{1}{c|}{}                                                                         & cube         & \textbf{0.505} & 0.936          & \textbf{0.484} & 0.236        & \textbf{0.506} & \textbf{0.599} & -              & -                         & -                         & -                         \\
\multicolumn{1}{c|}{}                                                                         & cylinder     & \textbf{0.375} & 0.803          & \textbf{0.253} & 0.578        & \textbf{0.371} & nan            & -              & -                         & -                         & -                         \\
\multicolumn{1}{c|}{}                                                                         & metal bucket & \textbf{0.501} & 0.916          & \textbf{0.492} & nan          & \textbf{0.508} & \textbf{0.35}  & -              & -                         & -                         & -                         \\
\multicolumn{1}{c|}{}                                                                         & plane        & \textbf{0.585} & 0.976          & 0.62           & nan          & \textbf{0.547} & \textbf{0.647} & -              & -                         & -                         & -                         \\
\multicolumn{1}{c|}{}                                                                         & rov          & 0.43           & 0.96           & 0.251          & 0            & 0.434          & nan            & -              & -                         & -                         & -                         \\
\multicolumn{1}{c|}{}                                                                         & all          & \textbf{0.492} & 0.93           & \textbf{0.454} & 0.136        & \textbf{0.488} & \textbf{0.521} & \textbf{0.575} & 0.156                     & \textbf{0.573}            & \textbf{0.596}            \\ \hline

\multicolumn{1}{c|}{\multirow{11}{*}{\begin{tabular}[c]{@{}c@{}}b2ub-\\ p5-50\end{tabular}}} & human body   & \textbf{0.509} & \textbf{0.974} & \textbf{0.431} & 0            & \textbf{0.512} & \textbf{0.522} & -              & -                         & -                         & -                         \\
\multicolumn{1}{c|}{}                                                                         & ball         & \textbf{0.536} & \textbf{0.973} & \textbf{0.509} & nan          & \textbf{0.535} & \textbf{0.468} & -              & -                         & -                         & -                         \\
\multicolumn{1}{c|}{}                                                                        & circle cage  & \textbf{0.512} & \textbf{0.924} & \textbf{0.515} & nan          & \textbf{0.508} & 0.555          & -              & -                         & -                         & -                         \\
\multicolumn{1}{c|}{}                                                                        & square cage  & \textbf{0.487} & \textbf{0.931} & \textbf{0.4}   & 0            & \textbf{0.49}  & nan            & -              & -                         & -                         & -                         \\
\multicolumn{1}{c|}{}                                                                         & tyre         & \textbf{0.483} & \textbf{0.917} & 0.404          & 0            & \textbf{0.472} & \textbf{0.593} & -              & -                         & -                         & -                         \\
\multicolumn{1}{c|}{}                                                                         & cube         & 0.497          & \textbf{0.944} & 0.442          & 0.323        & 0.497          & 0.57           & -              & -                         & -                         & -                         \\
 \multicolumn{1}{c|}{}                                                                        & cylinder     & \textbf{0.383} & \textbf{0.837} & \textbf{0.225} & \textbf{0.7} & \textbf{0.378} & nan            & -              & -                         & -                         & -                         \\
\multicolumn{1}{c|}{}                                                                         & metal bucket & \textbf{0.501} & \textbf{0.929} & \textbf{0.518} & nan          & \textbf{0.51}  & 0.276          & -              & -                         & -                         & -                         \\
\multicolumn{1}{c|}{}                                                                         & plane        & 0.573          & 0.996          & 0.59           & nan          & \textbf{0.559} & 0.598          & -              & -                         & -                         & -                         \\
\multicolumn{1}{c|}{}                                                                         & rov          & 0.44           & 0.97           & 0.289          & 0            & 0.444          & nan            & -              & -                         & -                         & -                         \\
\multicolumn{1}{c|}{}                                                                         & all          & \textbf{0.492} & \textbf{0.94}  & \textbf{0.432} & 0.17         & \textbf{0.49}  & \textbf{0.512} & \textbf{0.572} & 0.17                      & \textbf{0.571}            & 0.568                     \\ \hline
\multicolumn{1}{c|}{\multirow{11}{*}{\begin{tabular}[c]{@{}c@{}}nei-\\ g25\end{tabular}}}   & human body           & 0.47                 & \textbf{0.971}       & 0.333                & 0                    & 0.471                & \textbf{0.513}       & -                    & -                    & -                    & -                    \\
\multicolumn{1}{c|}{}                                                                       & ball                 & 0.507                & 0.961                & 0.449                & nan                  & 0.507                & 0.252                & -                    & -                    & -                    & -                    \\
\multicolumn{1}{c|}{}                                                                       & circle cage          & 0.495                & \textbf{0.921}       & 0.494                & nan                  & \textbf{0.487}       & 0.58                 & -                    & -                    & -                    & -                    \\
\multicolumn{1}{c|}{}                                                                       & square cage          & 0.465                & 0.939                & 0.344                & 0                    & 0.468                & nan                  & -                    & -                    & -                    & -                    \\
\multicolumn{1}{c|}{}                                                                       & tyre                 & 0.446                & 0.879                & 0.387                & 0                    & 0.433                & 0.564                & -                    & -                    & -                    & -                    \\
\multicolumn{1}{c|}{}                                                                       & cube                 & 0.489                & 0.935                & \textbf{0.449}       & 0.323                & 0.487                & \textbf{0.666}       & -                    & -                    & -                    & -                    \\
\multicolumn{1}{c|}{}                                                                       & cylinder             & \textbf{0.392}       & \textbf{0.863}       & \textbf{0.322}       & 0.634                & \textbf{0.388}       & nan                  & -                    & -                    & -                    & -                    \\
\multicolumn{1}{c|}{}                                                                       & metal bucket         & 0.475                & \textbf{0.937}       & 0.391                & nan                  & 0.487                & 0.2                  & -                    & -                    & -                    & -                    \\
\multicolumn{1}{c|}{}                                                                       & plane                & \textbf{0.591}       & 0.995                & 0.621                & nan                  & \textbf{0.568}       & 0.63                 & -                    & -                    & -                    & -                    \\
\multicolumn{1}{c|}{}                                                                       & rov                  & \textbf{0.46}        & 0.965                & \textbf{0.35}        & 0                    & \textbf{0.465}       & nan                  & -                    & -                    & -                    & -                    \\
\multicolumn{1}{c|}{}                                                                       & all                  & 0.479                & \textbf{0.937}       & 0.414                & 0.159                & \textbf{0.476}       & 0.486                & 0.558                & 0.159                & 0.556                & 0.532                \\ \hline
\multicolumn{1}{c|}{\multirow{11}{*}{\begin{tabular}[c]{@{}c@{}}nei-\\ g5-50\end{tabular}}} & human body           & \textbf{0.505}       & 0.96                 & \textbf{0.474}       & 0                    & \textbf{0.508}       & \textbf{0.508}       & -                    & -                    & -                    & -                    \\
\multicolumn{1}{c|}{}                                                                       & ball                 & \textbf{0.543}       & \textbf{0.969}       & \textbf{0.558}       & nan                  & \textbf{0.543}       & 0.329                & -                    & -                    & -                    & -                    \\
\multicolumn{1}{c|}{}                                                                       & circle cage          & \textbf{0.511}       & \textbf{0.934}       & \textbf{0.526}       & nan                  & \textbf{0.503}       & 0.576                & \textbf{-}           & -                    & -                    & -                    \\
\multicolumn{1}{c|}{}                                                                       & square cage          & \textbf{0.486}       & \textbf{0.946}       & \textbf{0.394}       & 0                    & \textbf{0.489}       & nan                  & -                    & -                    & -                    & -                    \\
\multicolumn{1}{c|}{}                                                                       & tyre                 & \textbf{0.505}       & \textbf{0.926}       & \textbf{0.446}       & 0                    & \textbf{0.498}       & \textbf{0.594}       & -                    & -                    & -                    & -                    \\
\multicolumn{1}{c|}{}                                                                       & cube                 & \textbf{0.508}       & \textbf{0.946}       & \textbf{0.505}       & 0.343                & \textbf{0.507}       & \textbf{0.603}       & -                    & -                    & -                    & -                    \\
\multicolumn{1}{c|}{}                                                                       & cylinder             & \textbf{0.389}       & \textbf{0.845}       & \textbf{0.287}       & 0.606                & \textbf{0.386}       & nan                  & -                    & -                    & -                    & -                    \\
\multicolumn{1}{c|}{}                                                                       & metal bucket         & \textbf{0.496}       & 0.925                & \textbf{0.508}       & nan                  & \textbf{0.506}       & 0.276                & -                    & -                    & -                    & -                    \\
\multicolumn{1}{c|}{}                                                                       & plane                & \textbf{0.593}       & \textbf{0.998}       & \textbf{0.648}       & nan                  & \textbf{0.564}       & 0.634                & -                    & -                    & -                    & -                    \\
\multicolumn{1}{c|}{}                                                                       & rov                  & 0.438                & 0.975                & 0.289                & 0                    & 0.444                & nan                  & -                    & -                    & -                    & -                    \\
\multicolumn{1}{c|}{}                                                                       & all                  & \textbf{0.498}       & \textbf{0.942}       & \textbf{0.464}       & 0.158                & \textbf{0.495}       & 0.503                & \textbf{0.575}       & \textbf{0.173}       & \textbf{0.573}       & 0.573                \\ \hline

\end{tabular}
}
\caption{The UATD dataset is detected using the SSD300 algorithm (Part 2 of 3). ``noisy" represents the original noise dataset, and dspnet, b2ub, and nei represent the target detection after denoising using the DSPNet, Blind2Unblind, and Neighbor2Neighbor single-frame denoising algorithms, respectively. Bold fonts indicate data that is higher than the corresponding ``noisy" dataset.}
\label{table34}
\end{table*}

\begin{table*}
\centering
\resizebox{\textwidth}{!}{

\begin{tabular}{cccccccccccc}
\hline
\multicolumn{1}{c|}{\multirow{11}{*}{\begin{tabular}[c]{@{}c@{}}nei-\\ p30\end{tabular}}}   & human body           & \textbf{0.506}       & \textbf{0.976}       & \textbf{0.42}        & 0                    & \textbf{0.507}       & \textbf{0.544}       & -                    & -                    & -                    & -                    \\
\multicolumn{1}{c|}{}                                                                       & ball                 & \textbf{0.535}       & \textbf{0.968}       & \textbf{0.521}       & nan                  & \textbf{0.535}       & 0.303                & -                    & -                    & -                    & -                    \\
\multicolumn{1}{c|}{}                                                                       & circle cage          & \textbf{0.514}       & \textbf{0.935}       & 0.498                & nan                  & \textbf{0.518}       & 0.524                & -                    & -                    & -                    & -                    \\
\multicolumn{1}{c|}{}                                                                       & square cage          & \textbf{0.488}       & \textbf{0.946}       & \textbf{0.385}       & 0                    & \textbf{0.491}       & nan                  & -                    & -                    & -                    & -                    \\
\multicolumn{1}{c|}{}                                                                       & tyre                 & \textbf{0.5}         & \textbf{0.92}        & \textbf{0.479}       & 0                    & \textbf{0.492}       & \textbf{0.586}       & -                    & -                    & -                    & -                    \\
\multicolumn{1}{c|}{}                                                                       & cube                 & \textbf{0.501}       & \textbf{0.942}       & \textbf{0.461}       & 0.307                & \textbf{0.502}       & 0.549                & -                    & -                    & -                    & -                    \\
\multicolumn{1}{c|}{}                                                                       & cylinder             & 0.352                & 0.785                & \textbf{0.207}       & 0.655                & 0.348                & nan                  & -                    & -                    & -                    & -                    \\
\multicolumn{1}{c|}{}                                                                       & metal bucket         & \textbf{0.491}       & 0.915                & \textbf{0.47}        & nan                  & \textbf{0.5}         & 0.295                & -                    & -                    & -                    & -                    \\
\multicolumn{1}{c|}{}                                                                       & plane                & 0.573                & \textbf{0.997}       & 0.604                & nan                  & \textbf{0.549}       & 0.614                & -                    & -                    & -                    & -                    \\
\multicolumn{1}{c|}{}                                                                       & rov                  & 0.429                & \textbf{0.977}       & 0.243                & 0                    & 0.433                & nan                  & -                    & -                    & -                    & -                    \\
\multicolumn{1}{c|}{}                                                                       & all                  & \textbf{0.489}       & 0.936                & \textbf{0.429}       & 0.16                 & \textbf{0.488}       & 0.488                & \textbf{0.572}       & 0.171                & \textbf{0.571}       & 0.527                \\ \hline
\multicolumn{1}{c|}{\multirow{11}{*}{\begin{tabular}[c]{@{}c@{}}nei-\\ p5-50\end{tabular}}} & human body           & \textbf{0.52}        & \textbf{0.97}        & \textbf{0.499}       & 0                    & \textbf{0.519}       & \textbf{0.559}       & -                    & -                    & -                    & -                    \\
\multicolumn{1}{c|}{}                                                                       & ball                 & \textbf{0.534}       & \textbf{0.965}       & \textbf{0.512}       & nan                  & \textbf{0.534}       & \textbf{0.374}       & -                    & -                    & -                    & -                    \\
\multicolumn{1}{c|}{}                                                                       & circle cage          & \textbf{0.53}        & \textbf{0.935}       & \textbf{0.613}       & nan                  & \textbf{0.524}       & \textbf{0.593}       & -                    & -                    & -                    & -                    \\
\multicolumn{1}{c|}{}                                                                       & square cage          & \textbf{0.501}       & \textbf{0.946}       & \textbf{0.473}       & 0                    & \textbf{0.503}       & nan                  & -                    & -                    & -                    & -                    \\
\multicolumn{1}{c|}{}                                                                       & tyre                 & \textbf{0.505}       & 0.904                & \textbf{0.477}       & 0                    & \textbf{0.501}       & 0.572                & -                    & -                    & -                    & -                    \\
\multicolumn{1}{c|}{}                                                                       & cube                 & \textbf{0.511}       & 0.935                & \textbf{0.504}       & 0.362                & \textbf{0.51}        & \textbf{0.624}       & -                    & -                    & -                    & -                    \\
\multicolumn{1}{c|}{}                                                                       & cylinder             & \textbf{0.379}       & 0.821                & \textbf{0.214}       & \textbf{0.667}       & \textbf{0.375}       & nan                  & -                    & -                    & -                    & -                    \\
\multicolumn{1}{c|}{}                                                                       & metal bucket         & \textbf{0.49}        & 0.912                & \textbf{0.468}       & nan                  & \textbf{0.497}       & \textbf{0.435}       & -                    & -                    & -                    & -                    \\
\multicolumn{1}{c|}{}                                                                       & plane                & 0.581                & \textbf{0.997}       & \textbf{0.654}       & nan                  & \textbf{0.549}       & 0.629                & -                    & -                    & -                    & -                    \\
\multicolumn{1}{c|}{}                                                                       & rov                  & 0.44                 & 0.967                & 0.291                & 0                    & 0.444                & nan                  & -                    & -                    & -                    & -                    \\
\multicolumn{1}{c|}{}                                                                       & all                  & \textbf{0.499}       & 0.935                & \textbf{0.47}        & 0.172                & \textbf{0.496}       & \textbf{0.541}       & \textbf{0.578}       & 0.171                & \textbf{0.575}       & \textbf{0.607}       \\ \hline
\multicolumn{1}{l}{}                                                                        & \multicolumn{1}{l}{} & \multicolumn{1}{l}{} & \multicolumn{1}{l}{} & \multicolumn{1}{l}{} & \multicolumn{1}{l}{} & \multicolumn{1}{l}{} & \multicolumn{1}{l}{} & \multicolumn{1}{l}{} & \multicolumn{1}{l}{} & \multicolumn{1}{l}{} & \multicolumn{1}{l}{} \\
\multicolumn{1}{l}{}                                                                        & \multicolumn{1}{l}{} & \multicolumn{1}{l}{} & \multicolumn{1}{l}{} & \multicolumn{1}{l}{} & \multicolumn{1}{l}{} & \multicolumn{1}{l}{} & \multicolumn{1}{l}{} & \multicolumn{1}{l}{} & \multicolumn{1}{l}{} & \multicolumn{1}{l}{} & \multicolumn{1}{l}{}
\end{tabular}
}
\caption{The UATD dataset is detected using the SSD300 algorithm (Part 3 of 3). ``noisy" represents the original noise dataset, and dspnet, b2ub, and nei represent the target detection after denoising using the DSPNet, Blind2Unblind, and Neighbor2Neighbor single-frame denoising algorithms, respectively. Bold fonts indicate data that is higher than the corresponding ``noisy" dataset.}
\label{table35}
\end{table*}

\begin{table*}
\centering
\resizebox{\textwidth}{!}{

\begin{tabular}{c|ccccccccccc}
\hline
\multirow{2}{*}{Denoising}                                              & \multicolumn{11}{c}{YOLOX}                                                                                                                                                                                              \\ \cline{2-12} 
                                                                        & class        & mAP            & mAP@0.5        & mAP@0.75        & mAP\_s         & mAP\_m         & mAP\_l         & AR             & \multicolumn{1}{l}{AR\_s} & \multicolumn{1}{l}{AR\_m} & \multicolumn{1}{l}{AR\_l} \\ \hline
\multirow{11}{*}{noisy}                                                 & human body   & 0.563          & 0.977          & 0.562          & 0              & 0.56           & 0.639          & -              & -                         & -                         & -                         \\
                                                                        & ball         & 0.571          & 0.968          & 0.639          & nan            & 0.571          & 0.328          & -              & -                         & -                         & -                         \\
                                                                        & circle cage  & 0.537          & 0.917          & 0.598          & nan            & 0.529          & 0.63           & -              & -                         & -                         & -                         \\
                                                                        & square cage  & 0.519          & 0.956          & 0.502          & 0              & 0.522          & nan            & -              & -                         & -                         & -                         \\
                                                                        & tyre         & 0.537          & 0.935          & 0.549          & 0              & 0.526          & 0.65           & -              & -                         & -                         & -                         \\
                                                                        & cube         & 0.56           & 0.959          & 0.591          & 0.424          & 0.559          & 0.647          & -              & -                         & -                         & -                         \\
                                                                        & cylinder     & 0.475          & 0.925          & 0.355          & 0.666          & 0.471          & nan            & -              & -                         & -                         & -                         \\
                                                                        & metal bucket & 0.508          & 0.927          & 0.497          & nan            & 0.522          & 0.201          & -              & -                         & -                         & -                         \\
                                                                        & plane        & 0.663          & 1              & 0.783          & nan            & 0.637          & 0.695          & -              & -                         & -                         & -                         \\
                                                                        & rov          & 0.516          & 0.977          & 0.472          & 0.1            & 0.519          & nan            & -              & -                         & -                         & -                         \\
                                                                        & all          & 0.545          & 0.954          & 0.555          & 0.198          & 0.542          & 0.542          & 0.604          & 0.203                     & 0.602                     & 0.567                     \\ \hline
\multirow{11}{*}{dspnet}                                                & human body   & 0.559          & \textbf{0.978} & \textbf{0.591} & 0              & 0.556          & 0.622          & -              & -                         & -                         & -                         \\
                                                                        & ball         & \textbf{0.577} & \textbf{0.971} & \textbf{0.649} & nan            & \textbf{0.577} & \textbf{0.353} & -              & -                         & -                         & -                         \\
                                                                        & circle cage  & 0.528          & \textbf{0.923} & 0.567          & nan            & 0.528          & 0.555          & -              & -                         & -                         & -                         \\
                                                                        & square cage  & 0.514          & 0.953          & \textbf{0.504} & 0              & 0.515          & nan            & -              & -                         & -                         & -                         \\
                                                                        & tyre         & \textbf{0.539} & \textbf{0.942} & \textbf{0.589} & 0              & \textbf{0.531} & 0.63           & -              & -                         & -                         & -                         \\
                                                                        & cube         & 0.553          & \textbf{0.961} & 0.584          & 0.362          & 0.554          & 0.634          & -              & -                         & -                         & -                         \\
                                                                        & cylinder     & 0.473          & \textbf{0.927} & \textbf{0.415} & 0.55           & 0.471          & nan            & -              & -                         & -                         & -                         \\
                                                                        & metal bucket & 0.508          & 0.918          & 0.479          & nan            & 0.52           & \textbf{0.27}  & -              & -                         & -                         & -                         \\
                                                                        & plane        & 0.657          & 1              & 0.768          & nan            & 0.637          & 0.684          & -              & -                         & -                         & -                         \\
                                                                        & rov          & \textbf{0.551} & \textbf{0.988} & \textbf{0.541} & 0              & 0.557          & nan            & -              & -                         & -                         & -                         \\
                                                                        & all          & \textbf{0.546} & \textbf{0.956} & \textbf{0.569} & 0.152          & \textbf{0.545} & 0.536          & \textbf{0.608} & 0.154                     & \textbf{0.608}            & 0.566                     \\ \hline
\multirow{11}{*}{\begin{tabular}[c]{@{}c@{}}b2ub-\\ g25\end{tabular}}   & human body   & 0.538          & \textbf{0.978} & 0.548          & 0              & 0.538          & 0.584          & -              & -                         & -                         & -                         \\
                                                                        & ball         & \textbf{0.573} & \textbf{0.971} & 0.633          & nan            & \textbf{0.573} & 0.303          & -              & -                         & -                         & -                         \\
                                                                        & circle cage  & 0.531          & 0.905          & 0.565          & nan            & 0.523          & 0.616          & -              & -                         & -                         & -                         \\
                                                                        & square cage  & 0.502          & \textbf{0.96}  & 0.419          & 0              & 0.505          & nan            & -              & -                         & -                         & -                         \\
                                                                        & tyre         & 0.498          & 0.928          & 0.509          & 0              & \textbf{0.493} & 0.588          & -              & -                         & -                         & -                         \\
                                                                        & cube         & 0.543          & 0.957          & 0.566          & 0.363          & 0.539          & \textbf{0.695} & -              & -                         & -                         & -                         \\
                                                                        & cylinder     & \textbf{0.481} & \textbf{0.926} & \textbf{0.401} & \textbf{0.699} & \textbf{0.476} & nan            & -              & -                         & -                         & -                         \\
                                                                        & metal bucket & 0.481          & 0.917          & 0.401          & nan            & 0.493          & \textbf{0.203} & -              & -                         & -                         & -                         \\
                                                                        & plane        & 0.65           & 0.99           & 0.738          & nan            & 0.619          & 0.689          & -              & -                         & -                         & -                         \\
                                                                        & rov          & \textbf{0.542} & \textbf{0.989} & \textbf{0.539} & 0              & \textbf{0.545} & nan            & -              & -                         & -                         & -                         \\
                                                                        & all          & 0.534          & 0.952          & 0.532          & 0.177          & 0.53           & 0.525          & 0.595          & 0.177                     & 0.593                     & 0.554                     \\ \hline
\multirow{11}{*}{\begin{tabular}[c]{@{}c@{}}b2ub-\\ g5-50\end{tabular}} & human body   & 0.523          & 0.976          & 0.511          & 0              & 0.52           & 0.595          & -              & -                         & -                         & -                         \\
                                                                        & ball         & 0.55           & 0.962          & 0.556          & nan            & 0.55           & 0.303          & -              & -                         & -                         & -                         \\
                                                                        & circle cage  & 0.507          & 0.892          & 0.492          & nan            & 0.506          & 0.567          & -              & -                         & -                         & -                         \\
                                                                        & square cage  & 0.479          & 0.947          & 0.396          & 0              & 0.482          & nan            & -              & -                         & -                         & -                         \\
                                                                        & tyre         & 0.484          & 0.899          & 0.459          & 0              & 0.47           & 0.596          & -              & -                         & -                         & -                         \\
                                                                        & cube         & 0.533          & \textbf{0.963} & 0.477          & 0.363          & 0.533          & 0.625          & -              & -                         & -                         & -                         \\
                                                                        & cylinder     & 0.442          & 0.91           & 0.332          & 0.6            & 0.439          & nan            & -              & -                         & -                         & -                         \\
                                                                        & metal bucket & 0.484          & 0.924          & 0.357          & nan            & 0.497          & \textbf{0.227} & -              & -                         & -                         & -                         \\
                                                                        & plane        & 0.602          & 1              & 0.714          & nan            & 0.583          & 0.627          & -              & -                         & -                         & -                         \\
                                                                        & rov          & 0.488          & 0.969          & 0.449          & 0              & 0.491          & nan            & -              & -                         & -                         & -                         \\
                                                                        & all          & 0.509          & 0.944          & 0.474          & 0.161          & 0.507          & 0.506          & 0.575          & 0.16                      & 0.574                     & 0.544                     \\ \hline
\end{tabular}
}
\caption{The UATD dataset is detected using the YOLOX algorithm (Part 1 of 3). ``noisy" represents the original noise dataset, and dspnet, b2ub, and nei represent the target detection after denoising using the DSPNet, Blind2Unblind, and Neighbor2Neighbor single-frame denoising algorithms, respectively. Bold fonts indicate data that is higher than the corresponding ``noisy" dataset.}
\label{table36}
\end{table*}

\begin{table*}
\centering
\resizebox{\textwidth}{!}{
\begin{tabular}{cccccccccccc}
\hline
\multicolumn{1}{c|}{\multirow{11}{*}{\begin{tabular}[c]{@{}c@{}}b2ub-\\ p30\end{tabular}}}   & human body   & 0.562          & 0.973          & \textbf{0.62}  & 0              & 0.559          & 0.627          & -              & -                         & -                         & -                         \\
\multicolumn{1}{c|}{}                                                                         & ball         & \textbf{0.579} & \textbf{0.97}  & \textbf{0.679} & nan            & \textbf{0.58}  & 0.328          & -              & -                         & -                         & -                         \\
\multicolumn{1}{c|}{}                                                                         & circle cage  & \textbf{0.557} & \textbf{0.924} & \textbf{0.652} & nan            & \textbf{0.556} & 0.599          & -              & -                         & -                         & -                         \\
\multicolumn{1}{c|}{}                                                                         & square cage  & \textbf{0.529} & 0.946          & \textbf{0.534} & 0              & \textbf{0.532} & nan            & -              & -                         & -                         & -                         \\
\multicolumn{1}{c|}{}                                                                         & tyre         & \textbf{0.555} & \textbf{0.961} & \textbf{0.588} & 0              & \textbf{0.546} & 0.65           & -              & -                         & -                         & -                         \\
 \multicolumn{1}{c|}{}                                                                        & cube         & \textbf{0.563} & \textbf{0.967} & \textbf{0.604} & 0.424          & \textbf{0.563} & \textbf{0.687} & -              & -                         & -                         & -                         \\
\multicolumn{1}{c|}{}                                                                         & cylinder     & \textbf{0.481} & \textbf{0.938} & \textbf{0.392} & 0.665          & \textbf{0.478} & nan            & -              & -                         & -                         & -                         \\
\multicolumn{1}{c|}{}                                                                         & metal bucket & \textbf{0.51}  & 0.923          & 0.488          & nan            & \textbf{0.523} & \textbf{0.251} & -              & -                         & -                         & -                         \\
\multicolumn{1}{c|}{}                                                                         & plane        & 0.645          & 0.99           & 0.777          & nan            & 0.611          & 0.683          & -              & -                         & -                         & -                         \\
\multicolumn{1}{c|}{}                                                                         & rov          & \textbf{0.534} & \textbf{0.979} & \textbf{0.532} & 0              & \textbf{0.537} & nan            & -              & -                         & -                         & -                         \\
\multicolumn{1}{c|}{}                                                                         & all          & \textbf{0.552} & \textbf{0.957} & \textbf{0.587} & 0.182          & \textbf{0.548} & \textbf{0.546} & \textbf{0.614} & 0.181                     & \textbf{0.612}            & \textbf{0.581}            \\ \hline
\multicolumn{1}{c|}{\multirow{11}{*}{\begin{tabular}[c]{@{}c@{}}b2ub-\\ p5-50\end{tabular}}} & human body   & \textbf{0.569} & 0.977          & \textbf{0.623} & 0              & \textbf{0.568} & 0.618          & -              & -                         & -                         & -                         \\
\multicolumn{1}{c|}{}                                                                         & ball         & \textbf{0.587} & \textbf{0.973} & \textbf{0.663} & nan            & \textbf{0.587} & \textbf{0.353} & -              & -                         & -                         & -                         \\
\multicolumn{1}{c|}{}                                                                         & circle cage  & 0.532          & 0.912          & 0.574          & nan            & \textbf{0.531} & 0.57           & -              & -                         & -                         & -                         \\
\multicolumn{1}{c|}{}                                                                         & square cage  & 0.513          & 0.953          & 0.496          & 0              & 0.515          & nan            & -              & -                         & -                         & -                         \\
\multicolumn{1}{c|}{}                                                                         & tyre         & \textbf{0.553} & \textbf{0.946} & \textbf{0.601} & 0              & \textbf{0.548} & 0.632          & -              & -                         & -                         & -                         \\
\multicolumn{1}{c|}{}                                                                         & cube         & \textbf{0.563} & \textbf{0.96}  & \textbf{0.607} & 0.423          & \textbf{0.563} & 0.642          & -              & -                         & -                         & -                         \\
\multicolumn{1}{c|}{}                                                                         & cylinder     & \textbf{0.476} & \textbf{0.941} & \textbf{0.366} & 0.634          & \textbf{0.472} & nan            & -              & -                         & -                         & -                         \\
\multicolumn{1}{c|}{}                                                                         & metal bucket & \textbf{0.509} & \textbf{0.931} & \textbf{0.516} & nan            & 0.522          & \textbf{0.251} & -              & -                         & -                         & -                         \\
\multicolumn{1}{c|}{}                                                                         & plane        & 0.658          & 1              & 0.742          & nan            & 0.631          & 0.69           & -              & -                         & -                         & -                         \\
\multicolumn{1}{c|}{}                                                                         & rov          & \textbf{0.525} & \textbf{0.999} & 0.456          & \textbf{0.2}   & \textbf{0.527} & nan            & -              & -                         & -                         & -                         \\
\multicolumn{1}{c|}{}                                                                         & all          & \textbf{0.549} & \textbf{0.959} & \textbf{0.564} & \textbf{0.209} & \textbf{0.546} & 0.537          & \textbf{0.61}  & \textbf{0.209}            & \textbf{0.609}            & 0.565                     \\ \hline
\multicolumn{1}{c|}{\multirow{11}{*}{\begin{tabular}[c]{@{}c@{}}nei-\\ g25\end{tabular}}}   & human body           & 0.43                 & 0.917                & 0.353                & 0                    & 0.421                & 0.565                & -                    & -                    & -                    & -                    \\
\multicolumn{1}{c|}{}                                                                       & ball                 & 0.505                & 0.936                & 0.455                & nan                  & 0.505                & 0.303                & -                    & -                    & -                    & -                    \\
\multicolumn{1}{c|}{}                                                                       & circle cage          & 0.451                & 0.877                & 0.393                & nan                  & 0.442                & 0.565                & -                    & -                    & -                    & -                    \\
\multicolumn{1}{c|}{}                                                                       & square cage          & 0.438                & 0.93                 & 0.337                & 0                    & 0.44                 & nan                  & -                    & -                    & -                    & -                    \\
\multicolumn{1}{c|}{}                                                                       & tyre                 & 0.422                & 0.816                & 0.392                & 0                    & 0.412                & 0.528                & -                    & -                    & -                    & -                    \\
\multicolumn{1}{c|}{}                                                                       & cube                 & 0.491                & 0.944                & 0.422                & 0.283                & 0.493                & 0.579                & -                    & -                    & -                    & -                    \\
\multicolumn{1}{c|}{}                                                                       & cylinder             & 0.422                & 0.896                & 0.317                & 0.6                  & 0.417                & nan                  & -                    & -                    & -                    & -                    \\
\multicolumn{1}{c|}{}                                                                       & metal bucket         & 0.424                & 0.857                & 0.376                & nan                  & 0.43                 & \textbf{0.277}       & -                    & -                    & -                    & -                    \\
\multicolumn{1}{c|}{}                                                                       & plane                & 0.649                & 1                    & 0.736                & nan                  & 0.606                & \textbf{0.697}       & -                    & -                    & -                    & -                    \\
\multicolumn{1}{c|}{}                                                                       & rov                  & 0.511                & \textbf{0.978}       & 0.387                & 0                    & 0.515                & nan                  & -                    & -                    & -                    & -                    \\
\multicolumn{1}{c|}{}                                                                       & all                  & 0.474                & 0.915                & 0.417                & 0.147                & 0.468                & 0.502                & 0.541                & 0.152                & 0.537                & 0.539                \\ \hline
\multicolumn{1}{c|}{\multirow{11}{*}{\begin{tabular}[c]{@{}c@{}}nei-\\ g5-50\end{tabular}}} & human body           & 0.554                & \textbf{0.986}       & \textbf{0.577}       & 0                    & 0.552                & 0.601                & -                    & -                    & -                    & -                    \\
\multicolumn{1}{c|}{}                                                                       & ball                 & \textbf{0.581}       & \textbf{0.972}       & \textbf{0.687}       & nan                  & \textbf{0.582}       & \textbf{0.353}       & -                    & -                    & -                    & -                    \\
\multicolumn{1}{c|}{}                                                                       & circle cage          & 0.537                & \textbf{0.928}       & 0.594                & nan                  & \textbf{0.536}       & 0.578                & \textbf{-}           & -                    & -                    & -                    \\
\multicolumn{1}{c|}{}                                                                       & square cage          & \textbf{0.522}       & 0.947                & 0.48                 & 0                    & \textbf{0.524}       & nan                  & -                    & -                    & -                    & -                    \\
\multicolumn{1}{c|}{}                                                                       & tyre                 & \textbf{0.541}       & \textbf{0.952}       & 0.546                & 0                    & \textbf{0.534}       & 0.615                & -                    & -                    & -                    & -                    \\
\multicolumn{1}{c|}{}                                                                       & cube                 & 0.553                & \textbf{0.967}       & 0.578                & 0.379                & 0.553                & \textbf{0.657}       & -                    & -                    & -                    & -                    \\
\multicolumn{1}{c|}{}                                                                       & cylinder             & 0.468                & \textbf{0.939}       & \textbf{0.358}       & 0.601                & 0.465                & nan                  & -                    & -                    & -                    & -                    \\
\multicolumn{1}{c|}{}                                                                       & metal bucket         & \textbf{0.512}       & \textbf{0.935}       & 0.45                 & nan                  & \textbf{0.524}       & \textbf{0.276}       & -                    & -                    & -                    & -                    \\
\multicolumn{1}{c|}{}                                                                       & plane                & 0.655                & 1                    & 0.756                & nan                  & 0.623                & 0.695                & -                    & -                    & -                    & -                    \\
\multicolumn{1}{c|}{}                                                                       & rov                  & \textbf{0.521}       & 0.968                & 0.448                & 0                    & \textbf{0.524}       & nan                  & -                    & -                    & -                    & -                    \\
\multicolumn{1}{c|}{}                                                                       & all                  & 0.544                & \textbf{0.959}       & 0.547                & 0.163                & 0.542                & 0.539                & \textbf{0.608}       & 0.167                & \textbf{0.606}       & \textbf{0.572}       \\ \hline
\end{tabular}}
\caption{The UATD dataset is detected using the YOLOX algorithm (Part 2 of 3). ``noisy" represents the original noise dataset, and dspnet, b2ub, and nei represent the target detection after denoising using the DSPNet, Blind2Unblind, and Neighbor2Neighbor single-frame denoising algorithms, respectively. Bold fonts indicate data that is higher than the corresponding ``noisy" dataset.}
\label{table37}
\end{table*}

\begin{table*}
\centering
\resizebox{\textwidth}{!}{
\begin{tabular}{cccccccccccc}
\hline
\multicolumn{1}{c|}{\multirow{11}{*}{\begin{tabular}[c]{@{}c@{}}nei-\\ p30\end{tabular}}}   & human body           & 0.557                & 0.975                & \textbf{0.603}       & 0                    & 0.556                & 0.615                & -                    & -                    & -                    & -                    \\
\multicolumn{1}{c|}{}                                                                       & ball                 & \textbf{0.593}       & \textbf{0.971}       & \textbf{0.683}       & nan                  & \textbf{0.593}       & \textbf{0.353}       & -                    & -                    & -                    & -                    \\
\multicolumn{1}{c|}{}                                                                       & circle cage          & \textbf{0.556}       & \textbf{0.935}       & \textbf{0.642}       & nan                  & \textbf{0.555}       & 0.582                & -                    & -                    & -                    & -                    \\
\multicolumn{1}{c|}{}                                                                       & square cage          & \textbf{0.525}       & 0.956                & 0.497                & 0                    & \textbf{0.527}       & nan                  & -                    & -                    & -                    & -                    \\
\multicolumn{1}{c|}{}                                                                       & tyre                 & \textbf{0.547}       & \textbf{0.946}       & \textbf{0.587}       & 0                    & \textbf{0.545}       & 0.624                & -                    & -                    & -                    & -                    \\
\multicolumn{1}{c|}{}                                                                       & cube                 & \textbf{0.565}       & \textbf{0.972}       & 0.568                & 0.423                & \textbf{0.563}       & \textbf{0.662}       & -                    & -                    & -                    & -                    \\
\multicolumn{1}{c|}{}                                                                       & cylinder             & 0.471                & \textbf{0.933}       & \textbf{0.36}        & 0.566                & 0.469                & nan                  & -                    & -                    & -                    & -                    \\
\multicolumn{1}{c|}{}                                                                       & metal bucket         & \textbf{0.51}        & 0.926                & 0.48                 & nan                  & \textbf{0.524}       & \textbf{0.227}       & -                    & -                    & -                    & -                    \\
\multicolumn{1}{c|}{}                                                                       & plane                & 0.651                & 1                    & 0.746                & nan                  & 0.635                & 0.675                & -                    & -                    & -                    & -                    \\
\multicolumn{1}{c|}{}                                                                       & rov                  & \textbf{0.538}       & \textbf{0.99}        & \textbf{0.537}       & 0.1                  & \textbf{0.542}       & nan                  & -                    & -                    & -                    & -                    \\
\multicolumn{1}{c|}{}                                                                       & all                  & \textbf{0.551}       & \textbf{0.96}        & \textbf{0.57}        & 0.182                & \textbf{0.551}       & 0.534                & \textbf{0.612}       & 0.187                & \textbf{0.612}       & 0.567                \\ \hline
\multicolumn{1}{c|}{\multirow{11}{*}{\begin{tabular}[c]{@{}c@{}}nei-\\ p5-50\end{tabular}}} & human body           & 0.559                & \textbf{0.989}       & \textbf{0.582}       & 0                    & \textbf{0.564}       & 0.569                & -                    & -                    & -                    & -                    \\
\multicolumn{1}{c|}{}                                                                       & ball                 & \textbf{0.585}       & 0.966                & \textbf{0.687}       & nan                  & \textbf{0.586}       & \textbf{0.353}       & -                    & -                    & -                    & -                    \\
\multicolumn{1}{c|}{}                                                                       & circle cage          & 0.53                 & 0.911                & 0.578                & nan                  & 0.529                & 0.561                & -                    & -                    & -                    & -                    \\
\multicolumn{1}{c|}{}                                                                       & square cage          & 0.517                & 0.945                & 0.474                & 0                    & 0.52                 & nan                  & -                    & -                    & -                    & -                    \\
\multicolumn{1}{c|}{}                                                                       & tyre                 & \textbf{0.556}       & \textbf{0.965}       & \textbf{0.61}        & 0                    & \textbf{0.551}       & 0.636                & -                    & -                    & -                    & -                    \\
\multicolumn{1}{c|}{}                                                                       & cube                 & 0.56                 & \textbf{0.968}       & \textbf{0.607}       & 0.412                & 0.559                & 0.645                & -                    & -                    & -                    & -                    \\
\multicolumn{1}{c|}{}                                                                       & cylinder             & \textbf{0.487}       & \textbf{0.933}       & \textbf{0.432}       & 0.633                & \textbf{0.485}       & nan                  & -                    & -                    & -                    & -                    \\
\multicolumn{1}{c|}{}                                                                       & metal bucket         & \textbf{0.51}        & 0.923                & 0.43                 & nan                  & 0.517                & \textbf{0.351}       & -                    & -                    & -                    & -                    \\
\multicolumn{1}{c|}{}                                                                       & plane                & 0.658                & 0.989                & \textbf{0.785}       & nan                  & 0.627                & 0.693                & -                    & -                    & -                    & -                    \\
\multicolumn{1}{c|}{}                                                                       & rov                  & \textbf{0.546}       & \textbf{0.978}       & \textbf{0.571}       & 0                    & \textbf{0.55}        & nan                  & -                    & -                    & -                    & -                    \\
\multicolumn{1}{c|}{}                                                                       & all                  & \textbf{0.551}       & \textbf{0.957}       & \textbf{0.576}       & 0.174                & \textbf{0.549}       & \textbf{0.544}       & \textbf{0.614}       & 0.181                & \textbf{0.613}       & \textbf{0.575}       \\ \hline
\multicolumn{1}{l}{}                                                                        & \multicolumn{1}{l}{} & \multicolumn{1}{l}{} & \multicolumn{1}{l}{} & \multicolumn{1}{l}{} & \multicolumn{1}{l}{} & \multicolumn{1}{l}{} & \multicolumn{1}{l}{} & \multicolumn{1}{l}{} & \multicolumn{1}{l}{} & \multicolumn{1}{l}{} & \multicolumn{1}{l}{} \\
\multicolumn{1}{l}{}                                                                        & \multicolumn{1}{l}{} & \multicolumn{1}{l}{} & \multicolumn{1}{l}{} & \multicolumn{1}{l}{} & \multicolumn{1}{l}{} & \multicolumn{1}{l}{} & \multicolumn{1}{l}{} & \multicolumn{1}{l}{} & \multicolumn{1}{l}{} & \multicolumn{1}{l}{} & \multicolumn{1}{l}{}
\end{tabular}
}
\caption{The UATD dataset is detected using the YOLOX algorithm (Part 3 of 3). ``noisy" represents the original noise dataset, and dspnet, b2ub, and nei represent the target detection after denoising using the DSPNet, Blind2Unblind, and Neighbor2Neighbor single-frame denoising algorithms, respectively. Bold fonts indicate data that is higher than the corresponding ``noisy" dataset.}
\label{table38}
\end{table*}

\end{document}